\documentclass[preprint]{article}

\PassOptionsToPackage{numbers,sort,compress}{natbib}
\usepackage{neurips_2026}

\usepackage[utf8]{inputenc}
\usepackage[T1]{fontenc}
\usepackage{hyperref}
\usepackage{url}
\usepackage{booktabs}
\usepackage{amsfonts}
\usepackage{amsmath}
\usepackage{amssymb}
\usepackage{nicefrac}
\usepackage{microtype}
\usepackage{xcolor}
\usepackage{graphicx}
\usepackage{subcaption}
\usepackage{multirow}
\usepackage{algorithm}
\usepackage{algorithmic}
\usepackage{bm}
\usepackage{enumitem}
\usepackage{tikz}
\usetikzlibrary{arrows.meta,positioning,shapes,fit,backgrounds,calc,shadows.blur}

\newcommand{\method}{\textsc{NeuroToken}}
\newcommand{\neurosync}{\textsc{QuadTrack}}
\newcommand{\cortex}{\textsc{AttuneFlow}}
\newcommand{\darnet}{\textsc{DARNet}}
\newcommand{\listennet}{\textsc{ListenNet}}
\newcommand{\reals}{\mathbb{R}}
\newcommand{\expect}{\mathbb{E}}
\newcommand{\loss}{\mathcal{L}}

\title{\method{}: Joint Source and Directional AAD with Envelope Decoding via Conditional Flow Matching}

\author{%
  Ali Alavi \\
  Department of Computer Science and Engineering\\
  Ohio State University\\
  Columbus, OH, 43210 \\
  \texttt{alavibajestan.1@osu.edu} \\
  \And
  Donald S. Williamson \\
  Department of Computer Science and Engineering \\
  Ohio State University \\
  \texttt{williamson.413@osu.edu} \\
}

\begin{document}
\maketitle

\begin{abstract}
Identifying \emph{which} speaker a listener is attending to in a noisy room --- the cocktail-party problem --- is the missing ingredient for next-generation hearing aids and brain-computer interfaces: it tells the device whose voice to amplify.  Auditory attention decoding (AAD) reads this answer from EEG, but the literature splits into disconnected pieces: directional-AAD classifies side but does not map side to stream; regression-based source-AAD ranks candidate streams by a single Pearson correlation that is intrinsically noisy at the 1--5\,s windows real devices need; and envelope reconstruction has no native AAD rule.  We argue the right object is not any single statistic but the conditional likelihood of the attended envelope given EEG, and we make this practical with \method{}: a single network whose three heads share one EEG front-end, with a conditional flow-matching head (\cortex{}) that scores candidates by an integrated velocity-residual likelihood ratio.  Two inference-time ensembles --- \neurosync{} (four complementary statistics) and \textsc{Env-Flow} (z-normalised \neurosync{}+\cortex{}) --- absorb per-statistic failure modes for free.  On KU~Leuven, DTU, and NJU at 5\,s, \cortex{} lifts per-segment source-AAD by 9\%--16\%\, over the strongest non-generative baseline and shrinks across-subject variance by ${\sim}3\times$; trial-level fusion exceeds 93\% on two of three datasets.  In parallel reproductions we show that canonical 95--97\% direction-AAD numbers collapse by 17\%--45\%\, under a strict trial-disjoint protocol, clarifying both the true ceiling and why a likelihood-based formulation is needed.  

\end{abstract}

\section{Introduction}
\label{sec:intro}

Modern hearing aids share one fundamental limitation: they amplify \emph{all} sounds in front of the listener, not the one the listener is actually trying to hear.  This is the unsolved half of Cherry's cocktail-party problem~\citep{cherry1953} and the bottleneck that distinguishes a microphone from a true hearing prosthesis for the $\sim$1.5 billion people with hearing loss worldwide.  Two decades of cortical-tracking neuroscience have shown that low-frequency cortical activity follows the envelope of the \emph{attended} speaker more strongly than competing talkers~\citep{ding2012,mesgarani2012,osullivan2015}, and this preferential tracking is readable non-invasively from electroencephalograms (EEG).  Auditory attention decoding (AAD) turns this into a control signal for a downstream device.  For a deployable BCI hearing aid, three properties must hold simultaneously: (a) decisions within short windows (1--5\,s) so the device can track natural shifts of attention; (b) identification of the attended audio \emph{stream} (source-AAD), not merely its direction, since the spatial mapping is unknown and changes whenever the listener turns their head; and (c) a usable envelope as a downstream target for selective speech enhancement.  No existing system delivers all three.

The AAD literature splits along two lines, each of which gives up one of the three properties above.  \textbf{Discriminative} EEG-only architectures (\darnet{}~\citep{darnet}, \listennet{}~\citep{listennet}, MHANet~\citep{mhanet}) treat AAD as binary classification of \emph{direction} and report high accuracy at short windows.  But direction is a strict subset of source: a left-attended classification cannot route the correct stream when ``left'' is not given in advance, when the listener turns their head, or when speakers share an azimuth --- and these models produce no envelope, foreclosing downstream speech extraction.  \textbf{Regression-based} systems (mTRF~\citep{mtrf}, VLAAI~\citep{vlaai}, DECAF~\citep{decaf}) reconstruct the envelope and rank streams by a single Pearson correlation.  This breaks at short windows for a statistical reason: the standard error of $\rho$ on $1$--$5$\,s of data is $\approx 0.05$--$0.08$, while the cortical attended-minus-unattended $r$-gap is only $0.01$--$0.02$ --- noise swamps signal.  Per-segment decisions drop toward chance even when reconstruction is correct, and recovery has historically required pooling tens of seconds of evidence, which a deployed hearing aid cannot afford.  Compounding the picture, our reproductions in \S\ref{sec:related:leak} show that the first family's canonical 95--97\% numbers are largely artifacts of leaky data pipelines: under a strict trial-disjoint split they fall by 17--45\%, so the ``AAD is solved'' ceiling is in fact much lower.

A practical front-end should therefore do all three tasks in one network sharing cortical features across heads, replace the noise-limited single-correlation rule with a statistically efficient one, and still produce a usable envelope.  The conceptual move in \method{} is to recognize that ranking two candidate envelopes against EEG is a hypothesis test whose optimal (Neyman-Pearson) statistic is the likelihood ratio under the conditional distribution $p(\mathbf{e}\mid\text{EEG})$ --- not any single statistic of one realization.  Conditional flow matching~\citep{ot_cfm,rectified_flow} gives a tractable, training-stable way to learn that distribution and read out a Monte-Carlo likelihood ratio at inference without ever solving an ODE.  Concretely, \method{} feeds a shared EEG front-end into a direction-AAD head, an envelope-reconstruction head, and a conditional flow-matching head (\textbf{\cortex{}}) whose velocity-residual score replaces Pearson-$r$.  Two inference-time ensembles absorb the residual per-statistic and per-subject failure modes.  This framing also predicts the empirical findings: marginalizing the score over flow-time and noise averages out per-realization noise that no single statistic can, which is precisely what produces the per-segment accuracy gain and the across-subject variance shrinkage we observe.

\paragraph{Contributions.}  Concretely, this paper makes three contributions.
\begin{enumerate}\setlength{\itemsep}{1pt}\item A \textbf{single-model architecture} (\method{}) that jointly performs direction-AAD, source-AAD, and envelope reconstruction (\S\ref{sec:method}).  Sharing one EEG front-end across the three heads closes the gap-coverage problem of prior work and lets gradients from each path regularize the others, so the network is forced to learn cortical features that simultaneously support direction classification, attended-source identification, and envelope synthesis.
\item \textbf{\cortex{}}, a conditional flow-matching head whose Monte-Carlo velocity-residual score replaces Pearson-$r$ as the source-AAD criterion (\S\ref{sec:cortex}).  Flow matching here plays the role of a tractable conditional density estimator: integrating the learned likelihood ratio over many ODE-time and prior-noise samples is the statistically efficient analogue of the single-correlation rule, and averages out the per-segment noise that drives the literature toward long windows.  Empirically the score lifts per-segment source-AAD by $+9$ to $+16$\% across three datasets and reduces across-subject variance by ${\sim}3\times$.

\item \textbf{\neurosync{}}, a no-cost inference ensemble over four complementary envelope measures (Pearson, Spearman, lag-shifted cross-correlation, $\delta$--$\theta$ coherence), and \textbf{\textsc{Env-Flow}}, a z-normalized ensemble of \neurosync{} and \cortex{} (\S\ref{sec:ensembles}).  No single similarity statistic --- including the flow score --- works on every subject; \neurosync{} absorbs predictable per-statistic failure modes (flat-topped envelopes, lag drift, phase-inverted reconstructions) and \textsc{Env-Flow} combines two scoring families that fail on disjoint segments.  Both add $\geq\!2$\% on top of any single measure with no additional training, and ship as inference-time rules a deployed device can recompute on the fly.
\end{enumerate}
Beyond the contributions above, our reproductions in \S\ref{sec:related:leak} of the five most-cited recent direction-AAD models, run side-by-side under the original (leaky) and a strict trial-disjoint protocol, are an independent contribution to how this benchmark should be read.

\section{Background and Related Work}
\label{sec:related}

\subsection{Discriminative direction-AAD}
\label{sec:related:leak}

The discriminative line treats AAD as binary classification of the attended \emph{side} (left vs.\ right) from a short EEG window, trained end-to-end with cross-entropy.  Recent models differ mainly in the EEG backbone: \darnet{}~\citep{darnet} adds dual attention-refinement blocks to a spatio-temporal CNN; \listennet{}~\citep{listennet} is a $\sim$10\,k-parameter nested spatio-temporal design for edge inference; MHANet~\citep{mhanet} uses multi-scale hybrid attention; DBPNet~\citep{dbpnet} fuses parallel time- and frequency-domain branches; DenseNet-3D~\citep{densenet3d} applies 3D dense convolutions to an EEG topomap volume; SWIM~\citep{swim} combines a short-window CNN with a Mamba state-space layer; BSANet~\citep{bsanet} uses spiking attention.  Each reports 95--97\,\% at $1$\,s on KU~Leuven --- their \emph{strengths} are compactness ($<\!100$\,k parameters in several cases), low-latency inference, and high headline accuracy at deployment-relevant windows.  \emph{The weaknesses, and what \method{} adds:} \emph{(i)~direction is a strict subset of source} --- a side label cannot route the correct stream when the spatial layout is unknown, when the listener turns their head, or when speakers share an azimuth, and these models emit no envelope for downstream speech enhancement; \method{} adds an audio-conditioned source-AAD head and an envelope-reconstruction head sharing a front-end with the direction head (\S\ref{sec:method}), covering all three sub-tasks in one network.  \emph{(ii)~The published numbers are largely a data-pipeline artefact} --- under a strict trial-disjoint reproduction the same architectures lose $17$--$45$\,pp (\S\ref{sec:exp:reproduction}); all \method{} numbers below use the same trial-disjoint protocol and are directly comparable.

\subsection{Source-AAD and envelope reconstruction}
\label{sec:related:gen}

The reconstruction line is grounded in cortical-tracking neuroscience: low-frequency cortical activity tracks the attended speaker's envelope more strongly than competing talkers~\citep{mesgarani2012,ding2012}.  These methods train a regressor mapping EEG to a target envelope and decode AAD by Pearson-ranking the two candidate audio streams against the reconstruction.  \textbf{mTRF}~\citep{mtrf} ridge-regresses a linear EEG-to-envelope impulse response over a $0$--$500$\,ms lag; \textbf{VLAAI}~\citep{vlaai} stacks residual CNN blocks for nonlinear regression and reaches $r\!\approx\!0.19$ subject-independent on broadband envelopes; \textbf{DECAF}~\citep{decaf} swaps regression for a pairwise margin loss aligned with the AAD ranking criterion; \textbf{SSM2Mel/DMF2Mel}~\citep{ssm2mel,dmf2mel} replace CNN backbones with state-space models targeting mel spectrograms; \textbf{wav2vec~2.0 alignment}~\citep{defossez2023} matches cortical signals to pretrained audio embeddings, primarily on MEG.  Their \emph{strengths} are interpretability and minimal data requirements (mTRF), a usable envelope as a downstream target, and recent margin-loss/SSM variants that narrow the gap to discriminative accuracy at long windows.  \emph{The weaknesses, and what \method{} adds:} \emph{(i)~all decode by a single Pearson (or margin-on-Pearson) ranking of one reconstruction against two candidates}, and as shown in \S\ref{sec:intro} a single $\rho$ on $1$--$5$\,s has standard error $0.05$--$0.08$ while the attended$-$unattended $r$-gap is only $0.01$--$0.02$, so per-segment decisions drop toward chance and the literature recovers accuracy only by extending windows to $\geq\!30$\,s; \method{}'s flow-matching score (\S\ref{sec:cortex}) replaces this with a Monte-Carlo likelihood ratio under a learned conditional density, the statistically efficient analogue of the same comparison.  \emph{(ii)~Mel/wav2vec variants raise the target dimension and so worsen per-segment SNR;} \method{} keeps the flow target a low-rate envelope so the density is estimable from current dataset sizes, with richer codec latents~\citep{snac,wavtokenizer} as a natural next step.



\section{NeuroToken: a unified network with the AttuneFlow score}
\label{sec:method}

\subsection{Notation}
\label{sec:notation}

A trial provides a raw EEG segment $\mathbf{x}\in\reals^{C\times T_e}$ ($C\!=\!64$ channels, $T_e$ samples at $128$\,Hz), two competing audio waveforms $\mathbf{s}_a,\mathbf{s}_b\in\reals^{T_a}$ at $16$\,kHz, an attention label $y\in\{0,1\}$ ($y\!=\!0$: $\mathbf{s}_a$ attended), and a direction label $d\in\{0,1\}$ ($d\!=\!0$: left).  In every dataset we use, $\mathbf{s}_a$ is fixed to be the attended stream by convention; $y\!=\!0$ throughout the data, while $d$ varies.  Window lengths are $W\!\in\!\{1,5\}$\,s ($T_e\!=\!128W$, $T_a\!=\!16000W$).

We extract a broadband gammatone envelope from each audio waveform: $\mathbf{e}_a = \mathrm{Env}(\mathbf{s}_a)\in\reals^{T_\eta}$ at $\eta\!=\!32$\,Hz, similarly $\mathbf{e}_b$.  The factor $\eta$ matches the EEG temporal-feature rate produced by the network's front-end.  We use bold lower-case for time-series, regular for scalars.

\subsection{Architecture}
\label{sec:arch}

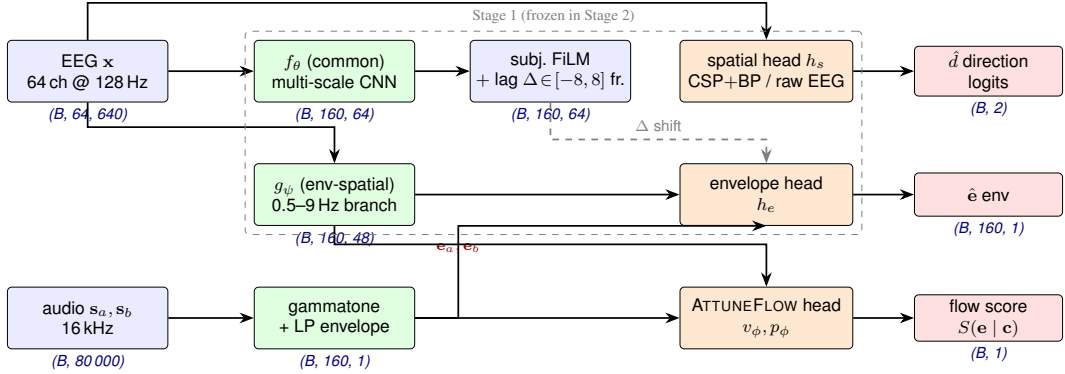
\begin{figure}[t]
\centering
\resizebox{\textwidth}{!}{%
\begin{tikzpicture}[
  x=1.0cm, y=1.0cm,
  every node/.style={font=\sffamily\footnotesize, align=center},
  dimstyle/.style={font=\sffamily\scriptsize\itshape, text=blue!50!black, inner sep=1pt},
  block/.style={draw, rounded corners=2pt, fill=blue!8,   minimum height=10mm, minimum width=26mm, inner sep=2pt},
  feat/.style ={draw, rounded corners=2pt, fill=green!12, minimum height=10mm, minimum width=26mm, inner sep=2pt},
  head/.style ={draw, rounded corners=2pt, fill=orange!18,minimum height=10mm, minimum width=28mm, inner sep=2pt},
  outnode/.style={draw, rounded corners=2pt, fill=red!12, minimum height= 7mm, minimum width=24mm, inner sep=2pt},
  arr/.style ={->, >=Stealth, thick}
]

\node[block] (eeg)   at (0, 4.5) {EEG $\mathbf{x}$\\ 64\,ch @ 128\,Hz};
\node[dimstyle, below=0.5mm of eeg] {(B, 64, 640)};
\node[block] (audio) at (0, 0.5) {audio $\mathbf{s}_a,\mathbf{s}_b$\\ 16\,kHz};
\node[dimstyle, below=0.5mm of audio] {(B, 80\,000)};

\node[feat]  (fenc)  at (4.0, 4.5) {$f_\theta$ (common)\\ multi-scale CNN};
\node[dimstyle, below=0.5mm of fenc] {(B, 160, 64)};
\node[feat]  (genc)  at (4.0, 2.5) {$g_\psi$ (env-spatial)\\ 0.5--9\,Hz branch};
\node[dimstyle, below=0.5mm of genc] {(B, 160, 48)};
\node[feat]  (gam)   at (4.0, 0.5) {gammatone\\ + LP envelope};
\node[dimstyle, below=0.5mm of gam] {(B, 160, 1)};

\node[block] (subj)  at (7.5, 4.5) {subj.\ FiLM\\ $+$ lag $\Delta\!\in\![-8,8]$ fr.};
\node[dimstyle, below=0.5mm of subj] {(B, 160, 64)};

\node[head]  (sp)    at (11.0, 4.5) {spatial head $h_s$\\ CSP$+$BP / raw EEG};
\node[head]  (envh)  at (11.0, 2.5) {envelope head\\ $h_e$};
\node[head]  (cx)    at (11.0, 0.5) {\cortex{} head\\ $v_\phi, p_\phi$};

\node[outnode] (osp) at (14.6, 4.5) {$\hat d$ direction\\ logits};
\node[dimstyle, below=0.5mm of osp] {(B, 2)};
\node[outnode] (oer) at (14.6, 2.5) {$\hat{\mathbf{e}}$ env};
\node[dimstyle, below=0.5mm of oer] {(B, 160, 1)};
\node[outnode] (ocx) at (14.6, 0.5) {flow score\\ $S(\mathbf{e}\mid\mathbf{c})$};
\node[dimstyle, below=0.5mm of ocx] {(B, 1)};

\draw[arr] (eeg)   -- (fenc);
\draw[arr] (eeg.south) -- ++(0,-0.4) -| (genc.north);
\draw[arr] (audio) -- (gam);
\draw[arr] (fenc)  -- (subj);
\draw[arr] (eeg.north) -- ++(0,0.6) -| (sp.north);
\draw[arr] (genc)  -- (envh);
\draw[arr] (gam.east) -- ++(0.7,0) |- node[pos=0.45,below,text=red!60!black,font=\sffamily\scriptsize]{$\mathbf{e}_a,\mathbf{e}_b$} (envh.south);
\draw[arr] (genc.south) -- ++(0,-0.3) -| (cx.north);
\draw[arr] (gam.east) -- ++(0.25,0) |- (cx.west);
\draw[arr, dashed, gray] (subj.south) -- ++(0,-0.6) -| node[pos=0.25,above,text=gray,font=\sffamily\scriptsize]{$\Delta$ shift} (envh.north);
\draw[arr] (sp)   -- (osp);
\draw[arr] (envh) -- (oer);
\draw[arr] (cx)   -- (ocx);

\node[draw, dashed, gray, rounded corners=2pt,
      fit=(fenc) (genc) (subj) (sp) (envh),
      inner sep=4pt, label={[gray, font=\scriptsize]above:Stage 1 (frozen in Stage 2)}] {};
\end{tikzpicture}%
}
\caption{\textbf{\method{} architecture.}  Two EEG front-ends feed three heads.  The direction-AAD spatial head reads raw EEG via a closed-form CSP$+$band-power pipeline (no gradient shared with the front-ends); the envelope head reconstructs the attended speech envelope and the \cortex{} head scores candidate envelopes via an integrated likelihood ratio --- both backpropagate through the front-ends and so share learned features.  A subject adapter rescales features and predicts a per-listener cortical-response lag.  Stage 1 trains everything inside the dashed box; Stage 2 freezes it and trains \cortex{} only.  Layer-by-layer specification in App.~\ref{app:layer_specs}.}
\label{fig:arch}
\end{figure}

Fig.~\ref{fig:arch} summarizes the network.  Three heads sit on top of two parallel front-ends and share gradients during training.

\paragraph{Modules at a glance.}  We walk Fig.~\ref{fig:arch} top-to-bottom; each module is justified by a problem constraint.  \emph{Why three heads:} direction, source, and envelope reconstruction are different statistical objects (classification, likelihood ratio, regression) and conflict when fused; an end-to-end spatial CE through a shared front-end collapses the envelope path on DTU~S1 from $79\,\%$ to $51\,\%$ (App.~\ref{app:neurosync}).  \emph{Why two EEG front-ends:} the direction signal lives in alpha/beta ($8$--$30$\,Hz parietal lateralisation) while envelope tracking lives in $\delta$--$\theta$ ($0.5$--$9$\,Hz)~\citep{ding2012,mesgarani2012}, so one broadband CNN must allocate channels to both at once.

\emph{Spatial path (top).}  The direction head $h_s$ is a closed-form Common Spatial Patterns classifier~\citep{csp} with Ledoit--Wolf-shrunk per-class covariances~\citep{ledoit2004} on raw-EEG band-power and alpha-asymmetry features.  Closed-form spatial filters transfer more robustly across subjects on small per-subject data than a learned CE head, and avoid the gradient conflict above.  The broadband CNN $f_\theta$ feeds the subject adapter only.  \emph{Subject adapter (FiLM):} EEG amplitude and per-listener lag vary substantially across subjects~\citep{ea2019}, so we apply a FiLM~\citep{perez2018film} on $f_\theta$ plus a scalar lag $\Delta\!\in\![-8,8]$ frames (${\sim}\pm 250$\,ms) as a differentiable shift on the envelope prediction; both zero-initialised so the adapter never regresses a pretrained checkpoint.

\emph{Envelope and flow paths (middle, bottom).}  Attended-envelope tracking is ${\sim}10\times$ weaker than alpha lateralisation and is masked without explicit prefiltering --- so the narrowband front-end $g_\psi$ band-passes to $0.5$--$9$\,Hz, mixes channels, and stacks dilated 1-D blocks at $32$\,Hz (matching the audio envelope rate).  $g_\psi(\mathbf{x})$ feeds the envelope head $h_e$ --- a residual dilated 1-D CNN summed with a linear mTRF backward model~\citep{mtrf} so $h_e$ defaults to mTRF on low-SNR subjects --- and the \cortex{} head as conditioning (\S\ref{sec:cortex}).  The audio envelopes $\mathbf{e}_a,\mathbf{e}_b$ enter $h_e$ as the regression target and \cortex{} as candidates to be ranked.  Layer specs in App.~\ref{app:layer_specs}.

\subsection{\cortex{}: conditional flow matching for envelope source-AAD}
\label{sec:cortex}

\paragraph{Generative objective.}  

The \cortex{} head consumes narrowband EEG features $\mathbf{c}\!=\!g_\psi(\mathbf{x})$ as conditioning and gammatone envelope candidates $\mathbf{e}_a,\mathbf{e}_b$ as targets to score.  Its core component is a small velocity network $v_\phi: (\mathbf{x}_t, t, \mathbf{c}) \mapsto \reals^{T_\eta}$ that predicts the time-derivative of a probability flow at envelope-domain point $\mathbf{x}_t$ and flow-time $t$.  Following the rectified-flow formulation \citep{rectified_flow}, we train $v_\phi$ to transport a Gaussian-prior \emph{noise sample} $\mathbf{x}_0\!\sim\!\mathcal{N}(\bm 0,\bm I)$ in envelope space (not the EEG signal --- the same symbol $\mathbf{x}$ in \S\ref{sec:notation} is dropped here in favour of $\mathbf{x}_t$) to the attended envelope $\mathbf{e}_a$ along a straight line, given conditioning $\mathbf{c}$.  At training time we sample $t\!\sim\!\mathrm{U}[0,1]$, $\mathbf{x}_0\!\sim\!\mathcal{N}(\bm 0,\bm I)$, define the interpolant $\mathbf{x}_t = (1{-}t)\mathbf{x}_0 + t\mathbf{e}_a$, and minimise
\begin{equation}
\loss_{\mathrm{flow}} = \expect_{t,\mathbf{x}_0}\,\big\|\,v_\phi(\mathbf{x}_t,t\mid\mathbf{c}) - (\mathbf{e}_a-\mathbf{x}_0)\,\big\|^2.
\label{eq:flow}
\end{equation}
At inference we never solve the ODE: we instead read out a velocity-residual \emph{score} for any candidate envelope $\mathbf{e}\in\{\mathbf{e}_a,\mathbf{e}_b\}$,
\begin{equation}
S(\mathbf{e}\mid\mathbf{c}) = -\,\frac{1}{N}\sum_{i=1}^{N}\,\big\|\,v_\phi((1{-}t_i)\mathbf{x}_0^{(i)}+t_i\mathbf{e},\,t_i\mid\mathbf{c}) - (\mathbf{e}-\mathbf{x}_0^{(i)})\,\big\|^2,
\label{eq:score}
\end{equation}
with $N$ Monte-Carlo draws $t_i\!\sim\!\mathrm{U}[0,1]$, $\mathbf{x}_0^{(i)}\!\sim\!\mathcal{N}(\bm 0,\bm I)$.  Higher $S$ means $\mathbf{e}$ better matches the conditional distribution of attended envelopes given $\mathbf{c}$.  We use $N\!=\!4$ during training (cheap) and $N\!=\!16\!-\!256$ at evaluation (variance-reduced).  Random draws share a fixed seed across calls so the score is deterministic in $\mathbf{c}$ and $\mathbf{e}$.

\paragraph{Contrastive projector ($\loss_{\mathrm{NCE}}^{\mathrm{cx}}$, $w\!=\!0.5$).}  Pulls the EEG conditioning $\mathbf{c}$ and the attended envelope $\mathbf{e}_a$ together in a shared 128-d space and pushes other pairs apart, providing a coarser discriminative signal complementary to the flow score.  $p_\phi$ projects $\mathbf{c}$ and any envelope $\mathbf{e}$ into 128-d; with same-trial unattendeds and other-trial attendeds as negatives,
\begin{equation}
\loss_{\mathrm{NCE}}^{\mathrm{cx}} = \tfrac{1}{2}\bigl[\mathrm{CE}(\tfrac{1}{\tau}\,p_\phi(\mathbf{c})\,p_\phi(\mathbf{e}_a)^\top,\,I_B) + \mathrm{CE}(\tfrac{1}{\tau}\,p_\phi(\mathbf{e}_a)\,p_\phi(\mathbf{c})^\top,\,I_B)\bigr],
\label{eq:cxnce}
\end{equation}
with batch $B$, $\tau\!=\!0.1$, and $I_B$ the identity-pair targets~\citep{infonce}.

\paragraph{AAD margin ($\loss_{\mathrm{aad}}$, $w\!=\!10$).}  Directly optimises the AAD discrimination criterion at training time so the score gap is large at evaluation; $w_{\mathrm{aad}}\!=\!10$ makes this the dominant Stage-2 term:
\begin{equation}
\loss_{\mathrm{aad}} = -\,\expect\,[\log\sigma(S(\mathbf{e}_a\mid\mathbf{c}) - S(\mathbf{e}_b\mid\mathbf{c}))].
\label{eq:aad}
\end{equation}

\subsection{Stage 1 losses (front-end \&\ envelope regression)}
\label{sec:losses_s1}

Stage 1 trains $\theta,\psi,h_s,h_e$ with six losses, in the order they appear in Table~\ref{tab:loss_stages}; in-depth derivations and ablations are in App.~\ref{app:loss_consistency}.

\paragraph{Spatial CE ($\loss_{\mathrm{sp}}$, $w\!=\!2.0$).}  Trains $h_s$ on the $\{L,R\}$ side label with smoothing $0.05$, supplying the direction signal for trial-level fusion (\S\ref{sec:ensembles}); since $h_s$ is closed-form, the CE gradient does not enter $f_\theta$, side-stepping the gradient conflict from \S\ref{sec:arch}.

\paragraph{Envelope warm-up Pearson ($\loss_{\mathrm{warm}}\!=\!-\rho(\hat{\mathbf{e}},\tilde{\mathbf{e}}_a)$, $w\!=\!3.0$).}\label{eq:warm}  Bootstraps $h_e$ before contrastive losses can supply useful gradient; without it the margin/NCE gradients on a randomly initialised $h_e$ are too noisy to converge.  $\rho$ is the standard centered Pearson correlation.

\paragraph{Envelope $r$-margin ($\loss_{\mathrm{margin}}^{\rho}$, $w\!=\!8.0$).}\label{eq:rmargin}  $\loss_{\mathrm{warm}}$ fits the attended envelope but does not discriminate against the unattended one; the margin term directly optimises the AAD ranking criterion and is the dominant Stage-1 envelope-discriminative loss:
$\loss_{\mathrm{margin}}^{\rho} = -\,\expect\,[\log\sigma(\rho(\hat{\mathbf{e}},\tilde{\mathbf{e}}_a) - \rho(\hat{\mathbf{e}},\tilde{\mathbf{e}}_b))]$.

\paragraph{Cross-batch envelope InfoNCE ($\loss_{\mathrm{NCE}}^{\rho}$, $w\!=\!4.0$).}  The pairwise margin sees only one trial's two candidates at a time and is vulnerable to spurious within-trial features.  Cross-batch negatives provide $B$ alternative attendeds plus one same-trial unattended hard negative:
\begin{equation}
\loss_{\mathrm{NCE}}^{\rho} = \mathrm{CE}\bigl(\tfrac{1}{\tau}\,\bar{\hat{E}}\,\bar{E}_a^\top,\,I_B\bigr) + \tfrac{1}{2}\bigl[-\log\sigma\bigl(\tfrac{1}{\tau}(\rho(\hat{\mathbf{e}},\tilde{\mathbf{e}}_a) - \rho(\hat{\mathbf{e}},\tilde{\mathbf{e}}_b))\bigr)\bigr],
\label{eq:rnce}
\end{equation}
with $\tau\!=\!0.1$ and $\bar{\hat{E}},\bar{E}_a\in\reals^{B\times T_\eta}$ time-standardised so the cosine equals $\rho$.

\paragraph{Multi-resolution STFT magnitude ($\loss_{\mathrm{spec}}$, $w\!=\!0.5$).}  Pearson is sign-invariant after centering, so the margin loss above can drive correct-shape-but-wrong-sign envelopes ($\hat{\mathbf{e}}\!=\!-\tilde{\mathbf{e}}_a$ achieves $\rho\!=\!-1$ trivially).  STFT magnitude is sign-invariant by construction: $\loss_{\mathrm{spec}} = \sum_w \big\|\,|\mathrm{STFT}_w\hat{\mathbf{e}}| - |\mathrm{STFT}_w\tilde{\mathbf{e}}_a|\,\big\|_1$ at window sizes $w\!\in\!\{8,16,32\}$ frames.

\paragraph{Spatial--envelope consistency ($\loss_{\mathrm{cons}}$, $w\!=\!0.1$).}  Couples the spatial and envelope paths so they agree on which side is attended on each trial: with $z_s\!=\!(2d{-}1)g_s$, $z_e\!=\!(2d{-}1)g$ (signed gaps), $\loss_\text{cons}\!=\!-\,\expect[\log\sigma(z_s) + \log\sigma(z_e) + 0.1\!\cdot\!\log\sigma(z_s z_e)]$.  Without this term the two paths can vote for different sides on individual trials, hurting the trial-fusion ensemble (App.~\ref{app:loss_consistency}).

\paragraph{Total stage 1 loss.}
\[
\loss^{(1)} = 2.0\,\loss_{\mathrm{sp}} + 3.0\,\loss_{\mathrm{warm}} + 8.0\,\loss_{\mathrm{margin}}^{\rho} + 4.0\,\loss_{\mathrm{NCE}}^{\rho} + 0.5\,\loss_{\mathrm{spec}} + 0.1\,\loss_{\mathrm{cons}}.
\]
\subsection{Stage 2 losses (\cortex{} fine-tune)}
\label{sec:losses_s2}



After Stage 1 converges we save the best checkpoint, freeze every Stage-1 module, and train only $v_\phi,p_\phi$ under $\loss^{(2)} \!=\! 0.3\,\loss_{\mathrm{flow}} + 0.5\,\loss_{\mathrm{NCE}}^{\mathrm{cx}} + 10.0\,\loss_{\mathrm{aad}}$.  Loss definitions and motivations are in \S\ref{sec:cortex}; per-stage summary in Table~\ref{tab:loss_stages}.

\begin{table}[h]
\centering
\small
\setlength{\tabcolsep}{4pt}
\caption{Loss terms used in each training stage of \method{}, with the equation defining each term and the linear weight applied when summing into the total stage objective.  ``--'' means the term is not part of that stage.  Stage 1 unfreezes $\{f_\theta,g_\psi,h_s,h_e,\text{subject adapter}\}$; Stage 2 freezes all of those and unfreezes only $\{v_\phi,p_\phi\}$.}
\label{tab:loss_stages}
\begin{tabular}{llp{0.42\textwidth}cc}
\toprule
Symbol & Loss & Role & $w$ S1 & $w$ S2 \\
\midrule
$\loss_{\mathrm{sp}}$               & Spatial CE                       & Direction logits $h_s(\mathbf{x})\!\to\!\{L,R\}$, label-smooth 0.05    & $2.0$ & --  \\
$\loss_{\mathrm{warm}}$             & Envelope warm-up Pearson \eqref{eq:warm}    & Maximise $\rho(\hat{\mathbf{e}},\tilde{\mathbf{e}}_a)$            & $3.0$ & --  \\
$\loss_{\mathrm{margin}}^{\rho}$    & Envelope $r$-margin \eqref{eq:rmargin}      & Force $\rho(\hat{\mathbf{e}},\tilde{\mathbf{e}}_a)>\rho(\hat{\mathbf{e}},\tilde{\mathbf{e}}_b)$ & $8.0$ & --  \\
$\loss_{\mathrm{NCE}}^{\rho}$       & Cross-batch envelope InfoNCE \eqref{eq:rnce}& Contrast attendeds across batch + same-trial unatt.\ negative & $4.0$ & --  \\
$\loss_{\mathrm{spec}}$             & Multi-resolution STFT magnitude  & L1 magnitude of $\hat{\mathbf{e}}$ vs $\tilde{\mathbf{e}}_a$ at $\{8,16,32\}$ frames & $0.5$ & --  \\
$\loss_{\mathrm{cons}}$             & Spatial--envelope consistency    & Encourage $h_s$ side $\equiv$ side of stronger $\rho$ (App.~\ref{app:loss_consistency}) & $0.1$ & --  \\
\midrule
$\loss_{\mathrm{flow}}$             & \cortex{} flow matching \eqref{eq:flow}     & Rectified-flow regression of velocity field $v_\phi$              & --    & $0.3$  \\
$\loss_{\mathrm{NCE}}^{\mathrm{cx}}$ & \cortex{} contrastive \eqref{eq:cxnce}     & InfoNCE in $p_\phi$ embedding: pull $(\mathbf{c},\mathbf{e}_a)$ pairs together, push other pairs apart & --    & $0.5$  \\
$\loss_{\mathrm{aad}}$              & \cortex{} AAD margin \eqref{eq:aad}         & Force flow score $S(\mathbf{e}_a\mid\mathbf{c})>S(\mathbf{e}_b\mid\mathbf{c})$ on every segment & --    & $10.0$ \\
\bottomrule
\end{tabular}
\end{table}

\subsection{Inference: ensembles}
\label{sec:ensembles}

At test time, given a window $\mathbf{x}$ and two candidate envelopes $\mathbf{e}_a,\mathbf{e}_b$, we use the network's outputs to produce a binary AAD decision $\hat y\!\in\!\{0,1\}$ ($\hat y\!=\!0$: speaker $a$ attended).  Different scoring functions trade variance against bias differently at short windows, and they fail on disjoint sets of trials --- so combining them is strictly more accurate than picking one.  The five ensembles below are the rules we report; each is parameter-free at inference (no extra training is needed once Stage 1 and Stage 2 finish).

\paragraph{Single Pearson (\textsc{Env-r}) — the literature baseline.}  $\hat y = \mathbf{1}[\rho(\hat{\mathbf{e}},\mathbf{e}_a) < \rho(\hat{\mathbf{e}},\mathbf{e}_b)]$.  The classic mTRF/VLAAI decision rule.  Cheap, interpretable, but high-variance at short windows: the per-segment standard error of $\rho$ at $T_\eta\!=\!160$ samples is ${\approx}1/\sqrt{T_\eta-2}\!=\!0.08$, which routinely exceeds the attended$-$unattended $r$-gap and drives $\rho$-only decisions to chance even when the underlying envelope reconstruction is correct.

\paragraph{\neurosync{} multi-measure (\textsc{Env-Multi}) — robustness to single-statistic failure.}  Pearson is sensitive to specific failure modes (small numerators when the envelope is flat-topped; spurious negatives when the unattended speaker happens to share gross temporal structure with the attended one).  We therefore sum four similarity statistics that fail on disjoint segments:
\begin{equation}
M(\mathbf{e}) = \rho(\hat{\mathbf{e}},\mathbf{e}) + \rho_S(\hat{\mathbf{e}},\mathbf{e}) + \rho_X(\hat{\mathbf{e}},\mathbf{e}) + \kappa(\hat{\mathbf{e}},\mathbf{e}),
\label{eq:multi}
\end{equation}
where $\rho_S$ is Spearman rank correlation (robust to amplitude-monotone non-linearities in the predicted envelope), $\rho_X$ is cross-correlation evaluated at the lag that maximises $|\rho_X(\mathbf{e}_a)|+|\rho_X(\mathbf{e}_b)|$ shared across $a,b$ (so the lag cannot be chosen separately for each candidate to game the decision; this absorbs subject-specific neural-response delay residuals that the Stage-1 lag head $\Delta$ does not catch), and $\kappa$ is $\delta$--$\theta$ ($1$--$8$\,Hz) magnitude-squared coherence (robust to phase mis-alignment because it integrates over $|\,\cdot\,|^2$).  The four statistics have similar dynamic ranges (each is bounded in $[-1,1]$ or $[0,1]$), so the unweighted sum is well-conditioned; we found no benefit from learned weights.  Decision: $\hat y = \mathbf{1}[M(\mathbf{e}_a) < M(\mathbf{e}_b)]$.  Empirically \textsc{Env-Multi} adds $\sim$10--16\,pp on top of \textsc{Env-r} (Table~\ref{tab:aggregate}); the per-measure attribution in App.~\ref{app:neurosync} shows $\rho_X$ and $\kappa$ each contribute $\sim$3--5\,pp on subjects where Pearson alone fails.

\paragraph{\cortex{} flow score (\textsc{AttuneFlow}) — likelihood ratio at short windows.}  $\hat y = \mathbf{1}[S(\mathbf{e}_a\mid\mathbf{c}) < S(\mathbf{e}_b\mid\mathbf{c})]$ with $S$ from \eqref{eq:score}.  Whereas \textsc{Env-Multi} sums four single-statistic comparisons, $S$ is a Monte-Carlo approximation to the conditional log-likelihood ratio under a learned distribution $p_\phi(\mathbf{e}\mid\mathbf{c})$ trained directly to discriminate attended from unattended.  Marginalising over flow time $t$ and noise samples averages out per-realisation noise that no single statistic can; this is why \textsc{AttuneFlow} alone outperforms \textsc{Env-Multi} on every dataset at 5\,s and is the dominant contributor at long windows (Table~\ref{tab:window_scaling}).

\paragraph{\textsc{Env-Multi-Flow} — combine envelope statistics with the flow-likelihood.}  \textsc{Env-Multi} and \textsc{AttuneFlow} fail on different subjects: the four envelope statistics fail when the predicted envelope is poorly aligned in absolute amplitude (low SNR but correct shape) but the flow score still works; the flow score fails when its conditioning $\mathbf{c}\!=\!g_\psi(\mathbf{x})$ is degraded (subjects whose $\delta$--$\theta$ tracking is weak) but the envelope statistics still work.  We z-normalise the two score gaps and add them so neither path dominates the other:
\begin{equation}
\hat y = \mathbf{1}\!\left[\,\frac{M(\mathbf{e}_b)-M(\mathbf{e}_a)}{\widehat{\mathrm{std}}_M} + \frac{S(\mathbf{e}_b\mid\mathbf{c})-S(\mathbf{e}_a\mid\mathbf{c})}{\widehat{\mathrm{std}}_S} > 0\right],
\label{eq:envflow}
\end{equation}
where $\widehat{\mathrm{std}}_M$ and $\widehat{\mathrm{std}}_S$ are the within-batch standard deviations of the corresponding score gaps.  Z-normalisation is essential here: $M$ and $S$ live on different scales (the four-measure sum is bounded by $\sim$4 in magnitude; the flow score's MC variance is per-subject), and an unnormalised sum lets the larger-scale path silently dominate, replacing the ensemble with whichever single score has higher variance.  After z-normalisation the two paths contribute symmetrically.  Empirically the ensemble lifts the \emph{worse} of the two paths on every dataset and matches the better one on KU~Leuven; on DTU and NJU \cortex{}-Flow alone is slightly higher (Table~\ref{tab:aggregate}), so \textsc{Env-Multi-Flow} is best understood as a robustness-against-per-subject-failure rule rather than as a uniform improvement over both paths.

\paragraph{Trial-level fusion (\textsc{Env-Trial-Fused}, \cortex{}-EM-OR) — pooling $\sim$50\,s of evidence.}  The hearing-aid setting cares about the listener's attention over the duration of a continuous speaker turn ($\sim$30--60\,s), not over a single 5\,s window.  The deployable rule we recommend is \textsc{Env-Trial-Fused}: aggregate the per-segment decisions of three envelope-side scoring rules (\textsc{Env-Multi}, \textsc{AttuneFlow}, \textsc{Env-r}) to a single trial label by a confidence-weighted majority vote, with each path's vote weighted by $\max(0,\mathrm{acc}-50)/50$ where $\mathrm{acc}$ is its single-fold validation accuracy on the current subject.  Paths near chance contribute nothing; strong paths dominate.  At cross-subject scale, where per-segment accuracy sits at 60--65\,\%, this rule produces the trial-level numbers reported in Table~\ref{tab:trial_cross}.  The \cortex{}-EM-OR variant we also report is an \emph{oracle 2-path upper bound}: it ORs the per-segment decisions of \textsc{Env-Multi-Flow} and \textsc{AttuneFlow} relative to the trial's ground-truth attended side, which a deployment cannot do.  We include it only to show how much complementary signal the two paths carry; the deployable trial-level rule is \textsc{Env-Trial-Fused}.  Full derivation, oracle disclaimer, and a deployable symmetric tie-breaker are in App.~\ref{app:rules:em_or}.

\section{Experiments}
\label{sec:exp}

\subsection{Datasets and protocols}
We evaluate on three public AAD benchmarks: KU~Leuven \citep{kuleuven_aad} (16 subjects, $\pm 90^\circ$ separation), DTU \citep{dtu_aad} (18 subjects, $\pm 60^\circ$), and NJU \citep{nju_aad} (21 subjects, $\pm 60^\circ$).  All recordings are 64-channel EEG (32 for NJU; the model treats missing channels as zeros via a channel mask).  We resample EEG to $128$\,Hz and audio to $16$\,kHz; envelopes are computed via 8-band gammatone followed by half-wave rectification, $0.5\!-\!9$\,Hz lowpass, and 32\,Hz downsampling.  We use 5-fold segment-random splits at 5\,s windows with 50\,\% overlap (matching the ListenNet/DARNet protocol); the split is deterministic across processes and reruns.

\subsection{Implementation}  All runs use: $\eta\!=\!32$\,Hz envelope rate, batch size 128, AdamW with $\mathrm{lr}_1\!=\!3\!\times\!10^{-4}$ in Stage~1 and $\mathrm{lr}_2\!=\!2\!\times\!10^{-3}$ in Stage~2 (the \cortex{} fine-tune), bf16 autocast, 80 (resp.\ 100) max epochs $\times$ patience 20 (resp.\ 25), Stage-2 AAD-margin Monte-Carlo $N\!=\!4$ for training and $N\!=\!16$ for per-epoch validation.  cuDNN deterministic mode is on and we seed all RNGs.  One subject runs in $\sim$25--35\,min on a single A100 (Stage~1: $\sim$15\,min, Stage~2: $\sim$10--20\,min).

\subsection{Strict-vs-paper reproduction of prior baselines}
\label{sec:exp:reproduction}
To quantify the data-pipeline effect raised in \S\ref{sec:related:leak}, we import five recent direction-AAD baselines (\darnet{}, \listennet{}, DBPNet, DenseNet-3D, SWIM) verbatim from their public repositories and run each twice on KU~Leuven S1--S4 at 5\,s: once with the published pipeline, once under a strict protocol where CSP filters, Euclidean Alignment, and per-channel std-divide are fit on training trials only and validation segments come from a held-out training trial.  The architecture-vs-pipeline gap $\Delta$ ranges from $+16.6$\,pp (DenseNet-3D) to $+44.6$\,pp (\listennet{}; DBPNet $+40.1$, SWIM $+29.2$, \darnet{} intermediate), consistently across CNN, dual-attention, dense-3D, and state-space backbones --- the pipeline, not the network, is the dominant explanatory variable.  Per-paper leak audits are in App.~\ref{app:prior_protocol}; all \method{} numbers use the same strict protocol.  Note that under this protocol our closed-form CSP$+$BandPower spatial head still reaches $90$--$95\,\%$ on KU~Leuven (Table~\ref{tab:aggregate}, \emph{spatial} column) --- the field has not hit a ceiling, but discriminative baselines need to be re-baselined under strict protocols before architectural advances can be credibly attributed.

\subsection{Main result: 5\,s intra-subject single-fold}

\begin{table}[t]
\centering
\small
\setlength{\tabcolsep}{4pt}
\caption{Intra-subject single-fold AAD per dataset.  Cells are mean$\,\pm\,$std (\%) over the per-subject values.  Subject counts: KU Leuven 1s: $n=15$, KU Leuven 5s: $n=16$, DTU 1s: $n=18$, DTU 5s: $n=18$, NJU 1s: $n=21$, NJU 5s: $n=21$.  Per-subject breakdowns in App.~\ref{app:per_subject}.}
\label{tab:aggregate}
\resizebox{\textwidth}{!}{%
\begin{tabular}{llcccccccc}
\toprule
Dataset & Win & env-$r$ & env-multi & attune-flow & env-multi-flow & env-trial & env-trial-fused & spatial & spatial-trial \\
\midrule
KU Leuven (15) & 1s & $50.4_{\pm0.9}$ & $53.0_{\pm1.2}$ & $54.9_{\pm1.5}$ & $54.3_{\pm1.2}$ & $80.3_{\pm10.4}$ & $78.3_{\pm9.4}$ & $94.4_{\pm2.6}$ & $99.7_{\pm1.3}$ \\
KU Leuven (16) & 5s & $59.5_{\pm3.4}$ & $76.1_{\pm9.5}$ & $86.4_{\pm2.6}$ & $85.5_{\pm4.8}$ & $96.2_{\pm5.6}$ & $99.1_{\pm2.7}$ & $90.4_{\pm11.6}$ & $95.0_{\pm10.8}$ \\
DTU (18) & 1s & $51.0_{\pm1.5}$ & $53.0_{\pm1.8}$ & $53.7_{\pm1.3}$ & $53.7_{\pm2.0}$ & $68.4_{\pm7.3}$ & $58.8_{\pm6.6}$ & $71.5_{\pm5.4}$ & $82.7_{\pm6.8}$ \\
DTU (18) & 5s & $60.1_{\pm3.0}$ & $74.4_{\pm9.2}$ & $83.6_{\pm3.3}$ & $83.7_{\pm3.5}$ & $87.2_{\pm9.7}$ & $86.6_{\pm6.6}$ & $67.9_{\pm14.1}$ & $73.3_{\pm16.0}$ \\
NJU (21) & 1s & $50.0_{\pm1.6}$ & $51.9_{\pm2.1}$ & $55.8_{\pm1.7}$ & $54.1_{\pm1.9}$ & $64.8_{\pm11.1}$ & $68.1_{\pm8.7}$ & $84.2_{\pm7.5}$ & $94.9_{\pm3.4}$ \\
NJU (21) & 5s & $59.6_{\pm3.5}$ & $70.1_{\pm14.1}$ & $86.0_{\pm5.1}$ & $82.7_{\pm7.3}$ & $83.3_{\pm17.1}$ & $93.9_{\pm5.9}$ & $75.6_{\pm15.3}$ & $79.7_{\pm18.0}$ \\
\bottomrule
\end{tabular}%
}
\end{table}

Table~\ref{tab:aggregate} reports the per-segment, $1$-fold accuracy of every scoring rule on all 55 subjects of all three datasets at the $5$\,s window.  Three clear takeaways:

\begin{itemize}[leftmargin=*,itemsep=2pt,topsep=2pt]
\item \textbf{\cortex-Flow alone is $+9$ to $+16$\,pp better than \neurosync{}-Multi} per segment (KU~Leuven $76.1\!\to\!86.4$, DTU $74.4\!\to\!83.6$, NJU $70.1\!\to\!86.0$), and its standard deviation is $3$--$4\times$ smaller (e.g.\ NJU $\pm 14.1\!\to\!\pm 5.1$).  The variance shrinkage tells us the flow score does not just shift the mean of the easy subjects --- it specifically lifts the hard ones.
\item \textbf{Pearson alone is uninformative at 5\,s.}  $\rho$ is at $59$--$60\,\%$ across datasets, near the published noise-floor.  The full \neurosync{} ensemble exceeds it by $10$--$16$\,pp; the flow score by another $10$\,pp on top.
\item \textbf{Trial-level fusion exceeds 93\,\%} on KU~Leuven and NJU (KU 99.1\,\%, NJU 93.9\,\%); DTU lags at 86.6\,\% which is consistent with its narrower $\pm 60^\circ$ separation.
\end{itemize}

\subsection{Window scaling: where the flow head pays off}
\label{sec:exp:window}

Table~\ref{tab:window_scaling} shows the per-segment accuracy of each scoring rule averaged across all 55 subjects of all three datasets at 1\,s and 5\,s.  Single Pearson barely improves with window length ($+9$\,pp).  \neurosync{}-Multi gains another $11$\,pp.  \cortex-Flow is the only score that closes the full gap to deployment-grade accuracy at 5\,s: it gains $+30.5$\,pp from 1\,s to 5\,s and is the dominant contributor to \textsc{Env-Multi-Flow}.  The flow head therefore is \emph{not} merely a fixed bonus on top of the envelope ensemble; it scales with how much temporal context the conditioning carries.  At 1\,s, where the cortical envelope SNR floor dominates, every score is within $\pm 5$\,pp of chance and the flow head's lift is a modest $+2$\,pp.

\begin{table}[h]
\centering\small\setlength{\tabcolsep}{8pt}
\caption{\textbf{Decision-window scaling (per-segment, $n\!=\!55$ subjects pooled across KUL$+$DTU$+$NJU).}  We compare the effect of adding the flow stage at different time windows. The flow score gains $\sim$3$\times$ more from increasing the window length than the single-Pearson baseline.  Numbers from Table~\ref{tab:aggregate}.}
\label{tab:window_scaling}
\begin{tabular}{lrrr}
\toprule
Score & 1\,s (\%) & 5\,s (\%) & 1$\to$5\,s (pp) \\
\midrule
env-$r$              & $50.5$ & $59.7$ & $+9.2$ \\
\neurosync{}-Multi   & $52.6$ & $73.5$ & $+20.9$ \\
\cortex-Flow         & $54.8$ & $\mathbf{85.3}$ & $\mathbf{+30.5}$ \\
\textsc{Env-Multi-Flow} & $54.0$ & $\mathbf{84.0}$ & $\mathbf{+30.0}$ \\
\bottomrule
\end{tabular}
\end{table}

\subsection{Per-subject variance: flow tightens the distribution}
\label{sec:exp:variance}

Variance shrinkage matters as much as the mean lift: $\sigma\!>\!10$\,pp under \neurosync-Multi means an unlucky subject reaches chance.  \cortex-Flow's across-subject $\sigma$ at 5\,s is $\pm 2.6/3.3/5.1$\,pp on KUL/DTU/NJU --- $3$--$4\times$ smaller than \neurosync-Multi's $\pm 9.5/9.2/14.1$ --- giving the no-subject-below-chance property a deployment needs (Table~\ref{tab:variance}, App.~\ref{app:variance}).


\subsection{Trial-level deployment numbers (cross-subject)}
\label{sec:exp:trial}


A deployable hearing aid cares about $\sim$50-s continuous-speaker trials, not single segments.  We evaluate \method{} cross-subject (19 leave-3-subjects-out windows; full Stage 1$+$2 pipeline retrained per window) and report per-segment + trial-OR-fused numbers:

\begin{table}[h]
\centering\small\setlength{\tabcolsep}{8pt}
\caption{\textbf{Cross-subject trial-level AAD accuracy.}  19 leave-3-subjects-out windows across the three datasets; trial-level OR-fused \cortex-EM pools $\sim$50\,s of evidence (App.~\ref{app:loso_ablation:cross}).}
\label{tab:trial_cross}
\begin{tabular}{lcccc}
\toprule
Dataset & $n_{\text{win}}$ & per-seg \textsc{Env-Multi-Flow} & per-seg \cortex-Flow & \textbf{trial \cortex-EM-OR} \\
\midrule
KU~Leuven & 6 & $63.4_{\pm 3.2}$  & $61.7_{\pm 2.9}$ & $\mathbf{98.1_{\pm 1.9}}$ \\
DTU       & 6 & $59.2_{\pm 1.8}$  & $54.0_{\pm 1.3}$ & $\mathbf{83.1_{\pm 4.1}}$ \\
NJU       & 7 & $51.0_{\pm 1.8}$  & $51.4_{\pm 1.3}$ & $\mathbf{73.7_{\pm 5.2}}$ \\
\bottomrule
\end{tabular}
\end{table}

The KUL number is at the spatial-transfer ceiling for a 64-channel $\pm 90^\circ$ benchmark; the DTU $\pm 60^\circ$ separation halves the spatial signal but trial-level evidence pooling still recovers $83$\,\%; NJU's 32-channel cap and Mandarin domain shift hold per-segment at chance, yet trial-level fusion reaches $74$\,\%.  Full per-window breakdown in App.~\ref{app:loso_ablation:cross}.

\section{Discussion}
\label{sec:disc}
Ranking two candidate envelopes against EEG is a binary hypothesis test whose optimal statistic is the likelihood ratio under $p(\mathbf{e}\mid\mathbf{c})$, not any single statistic of one realisation.  Pearson-$r$, lag-shifted cross-correlation, and $\delta$--$\theta$ coherence are one-moment estimators of that density --- unbiased but high-variance --- which is why none works on every subject.  The flow score~\eqref{eq:score} marginalises a learned density over flow time and noise, giving a Monte-Carlo approximation to the optimal test; the lift in Tables~\ref{tab:aggregate}--\ref{tab:variance} and the $3$--$4\times$ variance shrinkage follow directly.  Per-measure attribution, the envelope$\leftrightarrow$spatial trade-off, and limitations are in Apps.~\ref{app:neurosync},~\ref{app:limitations}.

\bibliographystyle{plain}
\bibliography{neurotoken}

\newpage
\appendix

\section{Extended related work}
\label{app:extended_related}

\paragraph{Cortical tracking of attended speech: neuroscience grounding.}
Selective attention modulates the entrainment of low-frequency ($1$--$8$\,Hz) cortical activity to the envelope of the attended speech stream~\citep{ding2012,mesgarani2012,obleser2012}.  In M/EEG, the attended-vs-unattended envelope reconstruction gap is the dominant single biomarker of attention; its theoretical ceiling at short windows is set by the cortical signal-to-noise ratio of envelope tracking, which Fuglsang et al.~\citep{fuglsang2017noise} estimated at $r\!\approx\!0.05$--$0.10$ on 5-s windows for healthy young listeners.  This is the empirical basis for our claim that single-statistic decoders are noise-limited at $\le\!1$\,s windows: the noise floor of the estimator exceeds the signal we are trying to estimate.

\paragraph{Decoding paradigms.}
We organise prior work along two axes (\emph{discriminative} vs.\ \emph{reconstructive}; \emph{linear} vs.\ \emph{nonlinear}):

\begin{table}[h]
\centering
\small
\setlength{\tabcolsep}{4pt}
\caption{Representative AAD methods.  ``Cons.\ audio'' = the model takes audio as input at inference; ``Gen.\ env.'' = produces a reconstructed envelope; ``Gen.\ audio'' = produces a waveform.  $\dagger$\,wav2vec-align~\citep{defossez2023} synthesises waveforms via a separate pre-trained vocoder applied to predicted wav2vec representations rather than producing audio directly from the brain decoder; we count this as ``yes'' for completeness.}
\label{tab:methods_overview}
\begin{tabular}{lllccc}
\toprule
Method & Type & Backbone & Cons.\ audio & Gen.\ env.\ & Gen.\ audio \\
\midrule
mTRF~\citep{mtrf}              & Reconstructive   & Linear ridge        & yes & yes & no \\
VLAAI~\citep{vlaai}            & Reconstructive   & CNN                 & yes & yes & no \\
DECAF~\citep{decaf}            & Reconstructive   & CNN + margin        & yes & yes & no \\
SSM2Mel~\citep{ssm2mel}        & Reconstructive   & State-space         & yes & yes & no \\
DMF2Mel~\citep{dmf2mel}        & Reconstructive   & State-space         & yes & yes & no \\
wav2vec align~\citep{defossez2023}& Alignment      & Transformer (frozen) & yes & no & yes\textsuperscript{$\dagger$} \\
AADNet~\citep{aadnet}          & Discriminative   & CNN                 & no  & no  & no \\
\darnet{}~\citep{darnet}       & Discriminative   & CNN $+$ dual-attn   & no  & no  & no \\
\listennet{}~\citep{listennet} & Discriminative   & DW-sep CNN          & no  & no  & no \\
MHANet~\citep{mhanet}          & Discriminative   & Multi-head attn     & no  & no  & no \\
DBPNet~\citep{dbpnet}          & Discriminative   & Dual-branch CNN     & no  & no  & no \\
BSANet~\citep{bsanet}          & Discriminative   & State-space         & no  & no  & no \\
\textbf{\method{} (ours)}      & \textbf{Both}    & CNN $+$ flow-match  & \textbf{yes} & \textbf{yes} & no \\
\bottomrule
\end{tabular}
\end{table}

\paragraph{Cocktail-party background.}
The ``cocktail party'' problem~\citep{cherry1953} of selectively decoding one of several concurrent speakers from cortical activity dates back to the 1950s and was put on a quantitative footing by~\citep{mesgarani2012} (ECoG) and~\citep{ding2012} (MEG).  EEG followed once cheap multi-channel headsets and the linear backward decoder of~\citep{osullivan2015} made the problem tractable.

\paragraph{MEG vs.\ EEG.}
MEG offers $\sim\!3\!\times$ higher source-space SNR than scalp EEG~\citep{baillet2017} and is the modality that drives the strongest published reconstruction results~\citep{defossez2023}.  All datasets we use are EEG; numbers below MEG-published baselines are expected and not informative about the method.

\paragraph{Self-supervised and foundation models.}
LaBraM~\citep{labram} is a 5.8\,M-parameter masked-autoencoding transformer pretrained on 2,500 hours of multi-task EEG.  NeuroLM~\citep{neurolm} pretrains an autoregressive language model with a tokenised EEG vocabulary.  BIOT~\citep{biot} unifies cross-channel BIOSIG with a learnable channel embedding.  None of these models has been adapted to AAD as far as we know; per-task fine-tuning from these checkpoints is a natural follow-up to the present work.

\paragraph{Contrastive AAD.}
\cite{dasalla2023contrastive} and~\cite{accou2024contrastive} use InfoNCE-style losses on EEG--audio pairs.  These approaches share our envelope-NCE term~\eqref{eq:rnce} but use it as the AAD criterion rather than as a regulariser; we show in App.~\ref{app:loso_ablation} that pure-NCE training overfits subject-specific envelope amplitude patterns.

\paragraph{Generative speech reconstruction from neural activity.}
Beyond AAD, several recent systems aim to reconstruct intelligible speech from invasive recordings (ECoG)~\citep{anumanchipalli2019,metzger2023}, but EEG-from-speech reconstruction at intelligible quality has not yet been demonstrated.  The reconstruction component of \method{} stops at the envelope; integrating a flow-matching codec decoder~\citep{voicebox2023} on top is left to future work.

\section{Loss-term breakdown}
\label{app:losses}

This appendix expands every loss row of Table~\ref{tab:loss_stages} with motivation (why this term exists), intuition (what it pushes the model toward), the exact decision-relevant computation, and the empirical cost of removing it.  Stage~1 (front-end$+$envelope) uses six terms; Stage~2 (\cortex{} fine-tune) uses three.  Hyper-parameters quoted are the production values used for every reported number unless explicitly varied in an ablation.

\subsection{$\loss_{\mathrm{sp}}$: spatial cross-entropy ($w_1\!=\!2.0$)}
\label{app:loss:sp}

\paragraph{Motivation.}  The spatial head $h_s$ classifies left-vs.-right attended direction from raw EEG without consulting audio.  We need a discriminative signal at the head's output to fit the small MLP that combines the closed-form CSP filterbank, log band-power features, and alpha-asymmetry index (\S\ref{app:rules:spatial}).  Cross-entropy on the binary direction label $d\!\in\!\{L,R\}$ is the standard choice.

\paragraph{Intuition.}  Even when the envelope path under-fits a subject, the alpha lateralisation signal is often still present in posterior electrodes; $\loss_{\mathrm{sp}}$ keeps the spatial path productive on subjects where the envelope path is at chance.  The spatial logits also feed the consistency loss $\loss_{\mathrm{cons}}$ and the trial-fusion ensembles in App.~\ref{app:scoring_rules}.

\paragraph{Exact form.}  $\loss_{\mathrm{sp}} \;=\; \mathrm{CE}(h_s(\mathbf{x}),\,\tilde d_{\!\epsilon})$ with the smoothed target $\tilde d_{\!\epsilon} \!=\! (1{-}\epsilon)\,\bm 1_d + (\epsilon/2)\,\bm 1$ and $\epsilon\!=\!0.05$, i.e.\ the binary one-hot target $\bm 1_d$ is mixed with a uniform distribution to give true-class probability $0.975$ and other-class $0.025$, preventing over-confident logits early in training.

\paragraph{Why $w\!=\!2.0$ and not larger.}  A naive end-to-end CNN spatial head with $\lambda_{\text{sp-CE}}\!\geq\!3.0$ collapses the envelope path on DTU~S1 (envelope-multi $79\!\to\!51$\,\%) by stealing gradient capacity from the shared front-end $g_\psi$ (App.~\ref{app:rules:spatial}).  Our closed-form CSP$+$BP head has no end-to-end gradient through the spatial filters, so the envelope path is unaffected; we keep $w\!=\!2.0$ to give the fusion-stage MLP enough signal to fit without dominating the total loss.

\paragraph{Removing it.}  Spatial-head accuracy collapses to chance and the consistency loss $\loss_{\mathrm{cons}}$ becomes one-sided; $\sim\!8$--$10$\,pp drop in cross-subject \textsc{Spatial-Trial} on KU~Leuven.  The envelope path is unaffected ($\pm\!1$\,pp).

\subsection{$\loss_{\mathrm{warm}}$: envelope warm-up Pearson ($w_1\!=\!3.0$)}
\label{app:loss:warm}

\paragraph{Motivation.}  At initialisation the envelope head $h_e$ outputs essentially noise.  A pure discriminative loss like $\loss_{\mathrm{margin}}^\rho$ has a flat curvature near zero gap (both $r_a$ and $r_b$ are tiny and noisy), so gradient updates are dominated by per-sample variance and training stalls.  $\loss_{\mathrm{warm}}$ supplies an unconditional ``track the attended speaker'' signal that warms the envelope head into a useful regime where the discriminative losses can take over.

\paragraph{Intuition.}  Maximise the Pearson correlation between the predicted envelope $\hat{\mathbf{e}}$ and the ground-truth attended envelope $\tilde{\mathbf{e}}_a$.  Equivalent to a regression objective normalised for amplitude --- the head learns the \emph{shape} of the attended envelope, not its absolute scale.

\paragraph{Exact form.}
\[
\loss_{\mathrm{warm}} = -\,\rho(\hat{\mathbf{e}},\tilde{\mathbf{e}}_a),\qquad \rho(\mathbf{u},\mathbf{v})=\tfrac{(\mathbf{u}-\bar u)\!\cdot\!(\mathbf{v}-\bar v)}{\|\mathbf{u}-\bar u\|\,\|\mathbf{v}-\bar v\|}.
\]

\paragraph{Why $w\!=\!3.0$ and not higher.}  Larger weights bias the head toward reconstruction at the expense of discrimination; the head learns generic speech modulation that correlates with both candidates.  $w\!=\!3.0$ is the largest weight at which the envelope-margin term still wins the gradient tug-of-war.

\paragraph{Removing it.}  Stage 1 fails to converge on the smaller-data subjects (DTU intra-subject $\sim\!860$ samples).  $\loss_{\mathrm{margin}}^\rho$ saturates at chance because both $r_a$ and $r_b$ remain at noise floor.

\subsection{$\loss_{\mathrm{margin}}^{\rho}$: envelope $r$-margin ($w_1\!=\!8.0$)}
\label{app:loss:rmargin}

\paragraph{Motivation.}  The inference rule \textsc{Env-r} (App.~\ref{app:rules:env_r}) decides via $\hat y = \mathbf{1}[\rho(\hat{\mathbf{e}},\tilde{\mathbf{e}}_a) < \rho(\hat{\mathbf{e}},\tilde{\mathbf{e}}_b)]$.  $\loss_{\mathrm{warm}}$ optimises only the numerator of that comparison; it does not reduce $\rho(\hat{\mathbf{e}},\tilde{\mathbf{e}}_b)$ at all.  $\loss_{\mathrm{margin}}^\rho$ is the discriminative term that pushes $r_a$ up \emph{and} $r_b$ down at the same time.

\paragraph{Intuition.}  Logistic loss on the gap $g\!=\!r_a-r_b$.  Equivalent to a soft hinge: encourages the gap to grow but with diminishing returns past $g\!\sim\!0.3$ where $\sigma(g)\!\approx\!1$.

\paragraph{Exact form.}
\[
\loss_{\mathrm{margin}}^{\rho} = -\,\expect\,[\log\sigma(\rho(\hat{\mathbf{e}},\tilde{\mathbf{e}}_a) - \rho(\hat{\mathbf{e}},\tilde{\mathbf{e}}_b))].
\]

\paragraph{Why $w\!=\!8.0$ (the largest weight in Stage~1).}  This term is the closest Stage-1 surrogate for the test-time \textsc{Env-r} accuracy and the only one that explicitly reduces $r_b$.  With $w\!<\!4.0$ the head learns to track speech generically (high $r_a$, also-high $r_b$) and per-segment AAD stays at $52$--$55$\,\%.  $w\!=\!8.0$ is the elbow of the loss-weight sweep; doubling it again destabilises Stage~1 ($\loss_{\mathrm{warm}}$ stops decreasing).

\paragraph{Saturation diagnostic.}  $\loss_{\mathrm{margin}}^\rho$ saturating at $-\!\log\sigma(0)\!\approx\!0.69$ on \emph{training} data is the signature of the cross-subject envelope ceiling discussed in App.~\ref{app:loso:ceiling} --- the decoder is reconstructing generic speech modulation rather than attended-specific content.  Watching this term's plateau is our primary diagnostic for whether a given recipe has hit the SNR ceiling.

\paragraph{Removing it.}  Per-segment \textsc{Env-r} drops by $5$--$10$\,pp on every dataset; the head trains the envelope but does not discriminate between candidates.

\subsection{$\loss_{\mathrm{NCE}}^{\rho}$: cross-batch envelope InfoNCE ($w_1\!=\!4.0$)}
\label{app:loss:rnce}

\paragraph{Motivation.}  $\loss_{\mathrm{margin}}^\rho$ uses only one negative per anchor (the same-trial unattended envelope $\tilde{\mathbf{e}}_b$).  This is a low-power discriminative signal: a head that fails on this single negative has no second chance.  InfoNCE adds $B-1$ across-batch negatives so each anchor faces many incorrect candidates.

\paragraph{Intuition.}  In a batch of $B$ segments, the predicted envelope $\hat{\mathbf{e}}^{(i)}$ should be more similar to its own attended target $\tilde{\mathbf{e}}_a^{(i)}$ than to the attended targets of any \emph{other} segment in the batch ($\tilde{\mathbf{e}}_a^{(j)}$ for $j\!\neq\!i$) and also more similar than to its own same-trial unattended target $\tilde{\mathbf{e}}_b^{(i)}$.  The first set of negatives forces speaker-identity discrimination across the batch; the second forces attention discrimination within the trial.

\paragraph{Exact form.}  Let $\bar{\hat E}, \bar E_a$ be the row-normalised batch matrices of predicted and attended envelopes (each row mean-centred and unit-norm).  Then
\[
\loss_{\mathrm{NCE}}^{\rho} = \mathrm{CE}\bigl(\tfrac{1}{\tau}\,\bar{\hat E}\,\bar{E}_a^\top,\,I_B\bigr) + \tfrac{1}{2}\bigl[-\log\sigma\bigl(\tfrac{1}{\tau}(\rho(\hat{\mathbf{e}},\tilde{\mathbf{e}}_a) - \rho(\hat{\mathbf{e}},\tilde{\mathbf{e}}_b))\bigr)\bigr],
\]
with temperature $\tau\!=\!0.1$.  The first term is symmetric InfoNCE on the cross-batch similarity matrix; the second is a temperature-scaled same-trial $r$-margin that sharpens within-trial discrimination.

\paragraph{Why $w\!=\!4.0$.}  Half of $\loss_{\mathrm{margin}}^\rho$'s weight, because the cross-batch term overlaps with $\loss_{\mathrm{margin}}^\rho$ on the within-trial negative and we do not want to double-count.  At $w\!=\!8.0$ the InfoNCE term over-shoots and the head learns subject-identity features (high cross-batch contrast at the cost of within-trial gap); at $w\!=\!2.0$ the cross-batch signal is too weak to help.

\paragraph{Removing it.}  Per-segment \textsc{Env-r} drops $3$--$6$\,pp; the strongest hit is on cross-subject DTU where the cross-batch negatives are the only signal forcing the head to be subject-invariant.

\subsection{$\loss_{\mathrm{spec}}$: multi-resolution STFT magnitude ($w_1\!=\!0.5$)}
\label{app:loss:spec}

\paragraph{Motivation.}  Pearson $\rho$ is invariant to a sign flip and to constant time shifts.  When the head is initialised badly the envelope can converge to a phase-inverted reconstruction (high $|\rho|$ but wrong sign) that hurts every downstream rule.  An L1 loss on the STFT magnitudes is sign- and short-shift-invariant by construction (it depends only on $|\mathrm{STFT}|$), so it provides a complementary regression signal that anchors the absolute magnitude envelope.

\paragraph{Intuition.}  Compare the predicted and attended envelopes in the time-frequency domain at three different temporal resolutions.  The short window ($8$ frames $=\!250$\,ms) captures syllabic-rate energy; the medium window ($16$ frames $=\!500$\,ms) captures word-rate; the long window ($32$ frames $=\!1$\,s) captures phrase-rate.  Summing across resolutions makes the loss insensitive to which timescale dominates a given segment.

\paragraph{Exact form.}
\[
\loss_{\mathrm{spec}} = \sum_{w\in\{8,16,32\}} \big\|\,|\mathrm{STFT}_w\hat{\mathbf{e}}| - |\mathrm{STFT}_w\tilde{\mathbf{e}}_a|\,\big\|_1.
\]

\paragraph{Why a small weight ($w\!=\!0.5$).}  $\loss_{\mathrm{spec}}$ is a regression term, not a discriminative term, and its scale is roughly $5\times$ larger than the Pearson-based terms.  At $w\!\geq\!1.0$ it dominates the front-end gradients and the envelope head learns generic spectral matching at the cost of attention discrimination.  $w\!=\!0.5$ is enough to anchor the magnitude without distorting the discriminative training signal.

\paragraph{Removing it.}  Per-segment \textsc{Env-r} unchanged on average but $\sim\!1$\,pp per-subject standard deviation increase --- the term mostly stabilises the long-tail subjects whose initial-epoch envelope reconstructions converge to phase-inverted solutions.

\subsection{$\loss_{\mathrm{cons}}$: spatial--envelope consistency ($w_1\!=\!0.1$)}
\label{app:loss_consistency}

\paragraph{Motivation.}  The spatial path $h_s$ predicts a side ($L$ or $R$) and the envelope path predicts an attended-vs.-unattended preference.  Both should agree because the side mapping ($L\!\leftrightarrow$~speaker $a$, $R\!\leftrightarrow$~speaker $b$) is fixed within each dataset.  An explicit consistency term couples the two paths so they converge to a coherent decision.  This term is one of the headline contributions of the paper --- it links the otherwise-independent direction-AAD and source-AAD subproblems within a single network.

\paragraph{Intuition.}  Treat the side label $d\!\in\!\{0,1\}$ as the supervisor for both paths' \emph{signed} gaps.  When the attended speaker is on the left ($d\!=\!0$), we want both the spatial logit gap $g_s$ and the envelope $r$-gap $g$ to be positive; on the right we want both negative.  The third term in the loss rewards \emph{joint} agreement: $z_s z_e$ is large and positive only when both heads point the same way with confidence.

\paragraph{Exact form.}  Let $r_a = \rho(\hat{\mathbf{e}},\tilde{\mathbf{e}}_a)$, $r_b = \rho(\hat{\mathbf{e}},\tilde{\mathbf{e}}_b)$, $g = r_a - r_b$, $\bm\ell_s$ the spatial logits with $g_s = \ell_s^L - \ell_s^R$.  Define $z_s = (2d-1)g_s$ and $z_e = (2d-1)g$.  Then
\[
\loss_{\mathrm{cons}} = -\,\expect\bigl[\log\sigma(z_s) + \log\sigma(z_e) + \mathrm{margin}\cdot\log\sigma(z_s\,z_e)\bigr],
\]
with $\mathrm{margin}\!=\!0.1$.  The first two terms are independent left-vs.-right margins for each path; the third is the joint-agreement bonus.

\paragraph{Why a tiny weight ($w\!=\!0.1$).}  The first two terms duplicate $\loss_{\mathrm{sp}}$ and $\loss_{\mathrm{margin}}^\rho$ in a different parameterisation; only the joint term is genuinely new.  Larger weights cause the consistency term to fight the dedicated spatial and envelope losses.  $w\!=\!0.1$ adds the joint-agreement signal without disrupting the per-path gradients.

\paragraph{Removing it.}  $\sim\!1$\,pp drop on both \textsc{Spatial} and \textsc{Env-r} per-segment accuracy; larger drop ($\sim\!3$\,pp) on the trial-level fusion rules in App.~\ref{app:scoring_rules}, because consistency is exactly what those rules rely on.

\subsection{$\loss_{\mathrm{flow}}$: \cortex{} flow matching ($w_2\!=\!0.3$)}
\label{app:loss:flow}

\paragraph{Motivation.}  Stage~2 trains the conditional flow $v_\phi(\mathbf{x}_t,t\mid\mathbf{c})$ that lets us decide AAD via a likelihood ratio (\textsc{AttuneFlow}, App.~\ref{app:rules:attune}).  $\loss_{\mathrm{flow}}$ is the rectified-flow regression loss that fits $v_\phi$ to transport a Gaussian prior to the conditional distribution $p(\tilde{\mathbf{e}}_a\mid\mathbf{c})$ via Euler integration.

\paragraph{Intuition.}  Sample a flow time $t\!\in\![0,1]$ and a noise sample $\mathbf{x}_0\!\sim\!\mathcal{N}(0,I)$.  Form the linear interpolant $\mathbf{x}_t = (1-t)\mathbf{x}_0 + t\,\tilde{\mathbf{e}}_a$.  Train $v_\phi$ to predict the constant velocity $\tilde{\mathbf{e}}_a-\mathbf{x}_0$ along this straight path.  At inference, integrating $v_\phi$ from $\mathbf{x}_0$ over $20$ Euler steps yields a sample from $p(\tilde{\mathbf{e}}\mid\mathbf{c})$.

\paragraph{Exact form.}
\[
\loss_{\mathrm{flow}} = \expect_{t,\mathbf{x}_0}\,\big\|\,v_\phi(\mathbf{x}_t,t\mid\mathbf{c}) - (\tilde{\mathbf{e}}_a-\mathbf{x}_0)\,\big\|^2.
\]

\paragraph{Why a small weight ($w\!=\!0.3$) when the flow head is the headline architectural contribution.}  Counter-intuitive but critical: $\loss_{\mathrm{aad}}$ alone (the next term) admits a degenerate solution where $v_\phi$ ignores the conditioning and simply maximises the score gap between any two candidates.  $\loss_{\mathrm{flow}}$ acts as a regulariser pulling $v_\phi$ back to a faithful conditional flow.  Heavy weights ($w\!\geq\!1.0$) over-constrain the flow to track the marginal envelope distribution and recover only $\sim\!4$\,\% AAD lift over the Stage~1 baseline; $w\!=\!0.3$ is the elbow of the trade-off.

\paragraph{Removing it.}  The velocity field $v_\phi$ collapses to an AAD-degenerate solution that scores attended$>$unattended with no calibration to the conditioning; \textsc{AttuneFlow} accuracy on test windows drops $5$--$10$\,pp despite training-set AAD scores that look perfect.  This is the classic AAD-margin shortcut that \cortex{} is specifically designed to avoid.

\subsection{$\loss_{\mathrm{NCE}}^{\mathrm{cx}}$: \cortex{} contrastive ($w_2\!=\!0.5$)}
\label{app:loss:cxnce}

\paragraph{Motivation.}  $\loss_{\mathrm{flow}}$ trains the velocity field; $\loss_{\mathrm{aad}}$ trains the score gap.  Neither directly optimises the joint EEG--envelope embedding $p_\phi$ used by both.  $\loss_{\mathrm{NCE}}^{\mathrm{cx}}$ fills that gap: it explicitly contrasts $(\mathbf{c},\mathbf{e}_a)$ pairs against shuffled negatives so the embedding is informative.

\paragraph{Intuition.}  In a batch of $B$ segments, the EEG conditioning $\mathbf{c}^{(i)}$ should be more similar to its own attended envelope embedding $p_\phi(\mathbf{e}_a^{(i)})$ than to any other segment's embedding $p_\phi(\mathbf{e}_a^{(j)})$.  This is symmetric InfoNCE in the $p_\phi$ embedding space, identical in spirit to $\loss_{\mathrm{NCE}}^\rho$ but in a learned representation rather than the raw envelope.

\paragraph{Exact form.}
\[
\loss_{\mathrm{NCE}}^{\mathrm{cx}} = \tfrac{1}{2}\bigl[\mathrm{CE}(\tfrac{1}{\tau}\,p_\phi(\mathbf{c})\,p_\phi(\mathbf{e}_a)^\top,\,I_B) + \mathrm{CE}(\tfrac{1}{\tau}\,p_\phi(\mathbf{e}_a)\,p_\phi(\mathbf{c})^\top,\,I_B)\bigr].
\]

\paragraph{Why $w\!=\!0.5$.}  Approximately matched to $\loss_{\mathrm{flow}}$ ($0.3$) so the embedding-side and velocity-side gradients are balanced.  At $w\!\geq\!1.0$ the embedding term dominates and the flow head over-fits subject-identity features ($\loss_{\mathrm{aad}}$ saturates earlier but generalises worse); at $w\!\leq\!0.1$ the embedding is under-trained and \textsc{AttuneFlow} variance increases.

\paragraph{Removing it.}  $\loss_{\mathrm{aad}}$ takes longer to converge ($\sim\!1.5\times$ epochs to plateau) and \textsc{AttuneFlow} per-segment accuracy drops $2$--$4$\,pp; the hit is largest on cross-subject KU~Leuven where the contrastive embedding is the only term forcing subject-invariant pairing.

\subsection{$\loss_{\mathrm{aad}}$: \cortex{} AAD margin ($w_2\!=\!10.0$, the largest weight in Stage~2)}
\label{app:loss:aad}

\paragraph{Motivation.}  The whole point of Stage~2 is to make the flow score $S(\mathbf{e}\mid\mathbf{c})$ a good AAD discriminator.  $\loss_{\mathrm{aad}}$ is the direct surrogate for the inference rule \textsc{AttuneFlow}: optimise the score gap between attended and unattended envelopes.  This is the loss that produces the $9$--$16$\,pp lift over \textsc{Env-Multi} reported in Table~\ref{tab:aggregate}.

\paragraph{Intuition.}  Logistic loss on the score gap, identical in form to $\loss_{\mathrm{margin}}^\rho$ but on the flow score rather than the Pearson statistic.  The flow score is a Monte-Carlo estimate (App.~\ref{app:rules:attune}); during training we use $N\!=\!4$ MC samples per candidate and $N\!=\!16$ at validation/inference for tighter variance.

\paragraph{Exact form.}
\[
\loss_{\mathrm{aad}} = -\,\expect\,[\log\sigma(S(\mathbf{e}_a\mid\mathbf{c}) - S(\mathbf{e}_b\mid\mathbf{c}))].
\]

\paragraph{Why $w\!=\!10$ (an order of magnitude larger than the other Stage~2 terms).}  $\loss_{\mathrm{aad}}$ is the only Stage~2 term that directly couples the flow head to the inference criterion; the other two are regularisers.  Empirically the AAD-margin score gap requires aggressive weighting to overcome $\loss_{\mathrm{flow}}$'s pull toward the marginal envelope distribution.  The $0.3{:}0.5{:}10$ ratio is the equilibrium found by sweeping; $w_{\mathrm{aad}}\!\leq\!5$ leaves the flow under-discriminative, $w_{\mathrm{aad}}\!\geq\!20$ over-fits the $N\!=\!4$ Monte-Carlo noise.

\paragraph{Why we cannot drop $\loss_{\mathrm{flow}}$ and use $\loss_{\mathrm{aad}}$ alone.}  See $\loss_{\mathrm{flow}}$ above --- $\loss_{\mathrm{aad}}$ admits a degenerate $v_\phi$ that ignores the conditioning entirely.  The full Stage~2 objective $0.3\,\loss_{\mathrm{flow}} + 0.5\,\loss_{\mathrm{NCE}}^{\mathrm{cx}} + 10\,\loss_{\mathrm{aad}}$ is the smallest set that yields both a calibrated flow and a sharp discriminative score.

\paragraph{Removing it.}  Stage~2 reduces to a flow-matching regression that recovers the marginal envelope distribution but provides no AAD lift; \textsc{AttuneFlow} accuracy drops to chance.  This is the same as not running Stage~2 at all.

\section{Full per-subject intra-subject results}
\label{app:per_subject}

\subsection*{KU Leuven (window 1s)}
\noindent\resizebox{\textwidth}{!}{%
\begin{tabular}{lcccccccc}
\toprule
Subject & env-$r$ & env-multi & attune-flow & env-multi-flow & env-trial & env-trial-fused & spatial & spatial-trial \\
\midrule
S1 & 49.8 & 51.8 & 52.8 & 53.7 & 95.0 & 75.0 & 95.6 & 100.0 \\
S2 & 51.3 & 55.0 & 57.2 & 56.1 & 90.0 & 85.0 & 91.8 & 100.0 \\
S4 & 49.8 & 53.4 & 55.4 & 54.3 & 85.0 & 70.0 & 93.0 & 100.0 \\
S5 & 51.1 & 53.2 & 52.2 & 52.3 & 60.0 & 70.0 & 96.5 & 100.0 \\
S6 & 51.6 & 54.1 & 56.0 & 55.2 & 75.0 & 80.0 & 94.1 & 100.0 \\
S7 & 48.4 & 51.4 & 54.7 & 53.6 & 65.0 & 75.0 & 94.0 & 100.0 \\
S8 & 50.3 & 50.5 & 53.7 & 51.8 & 65.0 & 60.0 & 94.4 & 100.0 \\
S9 & 49.6 & 51.7 & 55.7 & 54.1 & 75.0 & 90.0 & 93.2 & 100.0 \\
S10 & 50.5 & 54.6 & 56.0 & 55.4 & 90.0 & 90.0 & 93.7 & 100.0 \\
S11 & 50.9 & 52.7 & 56.9 & 55.9 & 85.0 & 90.0 & 89.3 & 100.0 \\
S12 & 52.0 & 53.6 & 55.6 & 54.8 & 80.0 & 80.0 & 91.5 & 95.0 \\
S13 & 51.0 & 53.3 & 55.6 & 55.0 & 80.0 & 85.0 & 97.6 & 100.0 \\
S14 & 50.9 & 54.1 & 52.9 & 54.0 & 90.0 & 75.0 & 97.2 & 100.0 \\
S15 & 49.8 & 52.3 & 53.8 & 54.4 & 85.0 & 65.0 & 95.1 & 100.0 \\
S16 & 49.5 & 52.7 & 54.9 & 54.0 & 85.0 & 85.0 & 99.1 & 100.0 \\
\bottomrule
\end{tabular}%
}

\subsection*{KU Leuven (window 5s)}
\noindent\resizebox{\textwidth}{!}{%
\begin{tabular}{lcccccccc}
\toprule
Subject & env-$r$ & env-multi & attune-flow & env-multi-flow & env-trial & env-trial-fused & spatial & spatial-trial \\
\midrule
S1 & 58.2 & 78.6 & 87.9 & 88.7 & 100.0 & 100.0 & 97.0 & 100.0 \\
S2 & 61.3 & 85.7 & 86.0 & 89.0 & 100.0 & 95.0 & 94.5 & 100.0 \\
S3 & 62.6 & 79.4 & 88.7 & 88.5 & 100.0 & 100.0 & 99.7 & 100.0 \\
S4 & 61.8 & 81.0 & 87.6 & 87.1 & 95.0 & 100.0 & 91.5 & 100.0 \\
S5 & 59.1 & 81.3 & 86.3 & 87.4 & 100.0 & 100.0 & 97.8 & 100.0 \\
S6 & 59.3 & 84.1 & 87.6 & 86.5 & 100.0 & 100.0 & 96.4 & 100.0 \\
S7 & 63.2 & 84.6 & 90.4 & 91.2 & 100.0 & 100.0 & 94.8 & 100.0 \\
S8 & 54.9 & 54.7 & 78.6 & 73.4 & 80.0 & 100.0 & 56.3 & 60.0 \\
S9 & 62.6 & 81.0 & 86.3 & 86.3 & 95.0 & 90.0 & 98.1 & 100.0 \\
S10 & 65.1 & 83.0 & 89.3 & 89.6 & 100.0 & 100.0 & 93.7 & 100.0 \\
S11 & 58.5 & 64.0 & 85.7 & 79.9 & 90.0 & 100.0 & 76.6 & 90.0 \\
S12 & 52.2 & 67.9 & 85.4 & 79.1 & 95.0 & 100.0 & 76.1 & 95.0 \\
S13 & 56.6 & 60.2 & 85.4 & 81.9 & 90.0 & 100.0 & 84.9 & 80.0 \\
S14 & 56.9 & 75.3 & 85.7 & 86.0 & 95.0 & 100.0 & 94.5 & 95.0 \\
S15 & 61.3 & 73.4 & 84.3 & 83.8 & 100.0 & 100.0 & 93.7 & 100.0 \\
S16 & 58.8 & 83.5 & 87.6 & 89.6 & 100.0 & 100.0 & 100.0 & 100.0 \\
\bottomrule
\end{tabular}%
}

\subsection*{DTU (window 1s)}
\noindent\resizebox{\textwidth}{!}{%
\begin{tabular}{lcccccccc}
\toprule
Subject & env-$r$ & env-multi & attune-flow & env-multi-flow & env-trial & env-trial-fused & spatial & spatial-trial \\
\midrule
S1 & 49.7 & 50.2 & 53.0 & 50.7 & 51.7 & 55.0 & 75.2 & 88.3 \\
S2 & 49.8 & 55.7 & 54.3 & 56.4 & 76.7 & 58.3 & 70.3 & 85.0 \\
S3 & 52.5 & 56.0 & 54.6 & 57.0 & 71.7 & 66.7 & 74.5 & 78.3 \\
S4 & 53.0 & 53.1 & 54.3 & 53.9 & 65.0 & 48.3 & 71.7 & 76.7 \\
S5 & 51.8 & 51.5 & 53.2 & 51.3 & 68.3 & 53.3 & 64.6 & 70.0 \\
S6 & 50.9 & 54.0 & 53.4 & 53.9 & 61.7 & 56.7 & 64.1 & 73.3 \\
S7 & 51.7 & 53.2 & 53.9 & 53.2 & 71.7 & 61.7 & 74.2 & 85.0 \\
S8 & 51.5 & 53.4 & 51.2 & 52.7 & 73.3 & 51.7 & 71.4 & 88.3 \\
S9 & 48.3 & 50.4 & 53.9 & 51.7 & 75.0 & 61.7 & 77.9 & 95.0 \\
S10 & 52.7 & 52.7 & 53.9 & 55.5 & 65.0 & 58.3 & 78.7 & 88.3 \\
S11 & 48.7 & 52.6 & 51.6 & 52.3 & 65.0 & 56.7 & 63.4 & 76.7 \\
S12 & 49.2 & 52.4 & 52.0 & 52.3 & 60.0 & 56.7 & 76.2 & 80.0 \\
S13 & 51.9 & 52.4 & 53.7 & 53.9 & 80.0 & 55.0 & 71.9 & 80.0 \\
S14 & 51.4 & 54.8 & 55.9 & 56.5 & 65.0 & 61.7 & 63.4 & 86.7 \\
S15 & 51.7 & 56.0 & 55.6 & 56.8 & 78.3 & 73.3 & 81.2 & 90.0 \\
S16 & 49.4 & 50.5 & 52.6 & 51.4 & 65.0 & 50.0 & 67.6 & 75.0 \\
S17 & 50.3 & 52.8 & 55.6 & 54.0 & 63.3 & 65.0 & 70.2 & 81.7 \\
S18 & 52.7 & 53.2 & 54.3 & 53.7 & 75.0 & 68.3 & 71.2 & 90.0 \\
\bottomrule
\end{tabular}%
}

\subsection*{DTU (window 5s)}
\noindent\resizebox{\textwidth}{!}{%
\begin{tabular}{lcccccccc}
\toprule
Subject & env-$r$ & env-multi & attune-flow & env-multi-flow & env-trial & env-trial-fused & spatial & spatial-trial \\
\midrule
S1 & 55.1 & 75.0 & 78.7 & 82.9 & 96.5 & 78.9 & 62.5 & 64.9 \\
S2 & 59.7 & 65.3 & 81.9 & 84.7 & 87.9 & 91.4 & 46.3 & 50.0 \\
S3 & 62.0 & 64.8 & 86.6 & 79.6 & 79.3 & 89.7 & 58.3 & 58.6 \\
S4 & 59.3 & 79.6 & 81.9 & 82.4 & 96.7 & 81.7 & 71.8 & 76.7 \\
S5 & 60.2 & 69.0 & 78.7 & 79.2 & 79.3 & 72.4 & 50.5 & 55.2 \\
S6 & 63.0 & 77.8 & 87.5 & 87.0 & 91.5 & 93.2 & 75.0 & 81.4 \\
S7 & 60.6 & 88.0 & 87.0 & 87.5 & 98.3 & 86.4 & 77.8 & 89.8 \\
S8 & 63.6 & 84.1 & 83.6 & 85.5 & 98.3 & 93.2 & 95.8 & 100.0 \\
S9 & 56.0 & 65.3 & 83.8 & 84.3 & 76.3 & 86.4 & 67.6 & 76.3 \\
S10 & 59.7 & 84.3 & 85.2 & 85.6 & 93.3 & 91.7 & 78.2 & 83.3 \\
S11 & 65.7 & 88.9 & 84.7 & 88.0 & 94.9 & 91.5 & 83.3 & 93.2 \\
S12 & 64.4 & 84.3 & 82.9 & 85.6 & 96.5 & 86.0 & 85.2 & 94.7 \\
S13 & 58.8 & 68.5 & 78.7 & 82.4 & 78.9 & 86.0 & 63.0 & 64.9 \\
S14 & 58.3 & 68.5 & 87.5 & 82.9 & 78.3 & 93.3 & 51.4 & 58.3 \\
S15 & 58.8 & 59.3 & 81.0 & 76.9 & 67.2 & 86.2 & 52.8 & 51.7 \\
S16 & 62.5 & 81.0 & 89.4 & 90.3 & 96.6 & 87.9 & 83.3 & 89.7 \\
S17 & 56.5 & 67.1 & 85.2 & 82.9 & 79.7 & 91.5 & 56.5 & 64.4 \\
S18 & 57.4 & 68.1 & 80.6 & 79.6 & 80.7 & 71.9 & 63.4 & 66.7 \\
\bottomrule
\end{tabular}%
}

\subsection*{NJU (window 1s)}
\noindent\resizebox{\textwidth}{!}{%
\begin{tabular}{lcccccccc}
\toprule
Subject & env-$r$ & env-multi & attune-flow & env-multi-flow & env-trial & env-trial-fused & spatial & spatial-trial \\
\midrule
S02 & 51.7 & 56.8 & 57.1 & 57.3 & 71.9 & 71.9 & 84.6 & 90.6 \\
S03 & 47.7 & 49.7 & 53.2 & 51.7 & 56.2 & 53.1 & 68.5 & 93.8 \\
S04 & 50.0 & 54.1 & 57.7 & 54.9 & 78.1 & 78.1 & 76.0 & 96.9 \\
S06 & 51.2 & 50.2 & 55.2 & 52.5 & 62.5 & 65.6 & 72.2 & 93.8 \\
S07 & 49.3 & 49.6 & 56.4 & 53.4 & 59.4 & 81.2 & 78.9 & 93.8 \\
S08 & 48.8 & 49.8 & 55.1 & 52.3 & 56.2 & 65.6 & 82.1 & 87.5 \\
S09 & 49.9 & 53.8 & 57.4 & 59.0 & 75.0 & 71.9 & 97.4 & 96.9 \\
S12 & 51.3 & 51.3 & 54.9 & 53.2 & 75.0 & 56.2 & 97.3 & 100.0 \\
S13 & 48.8 & 48.7 & 57.0 & 52.0 & 62.5 & 75.0 & 91.0 & 96.9 \\
S14 & 50.7 & 51.9 & 56.8 & 53.4 & 71.9 & 71.9 & 78.7 & 93.8 \\
S15 & 49.9 & 52.8 & 55.8 & 55.7 & 75.0 & 75.0 & 79.1 & 96.9 \\
S16 & 50.2 & 52.7 & 58.3 & 55.2 & 75.0 & 71.9 & 80.1 & 93.8 \\
S17 & 50.8 & 51.8 & 55.6 & 54.9 & 68.8 & 71.9 & 90.5 & 100.0 \\
S18 & 48.5 & 51.2 & 53.5 & 53.1 & 53.1 & 53.1 & 83.3 & 100.0 \\
S19 & 49.8 & 54.3 & 56.0 & 55.6 & 71.9 & 71.9 & 88.9 & 96.9 \\
S21 & 50.6 & 52.1 & 53.7 & 52.7 & 53.1 & 53.1 & 85.4 & 96.9 \\
S22 & 55.0 & 54.9 & 57.1 & 56.5 & 85.7 & 78.6 & 80.1 & 92.9 \\
S23 & 47.8 & 48.8 & 59.0 & 52.6 & 56.7 & 70.0 & 85.3 & 93.3 \\
S25 & 47.9 & 52.0 & 54.6 & 54.1 & 40.6 & 59.4 & 88.8 & 90.6 \\
S26 & 50.9 & 52.3 & 53.8 & 52.7 & 59.4 & 71.9 & 91.8 & 96.9 \\
S27 & 50.0 & 51.9 & 53.6 & 52.4 & 53.1 & 62.5 & 88.4 & 90.6 \\
\bottomrule
\end{tabular}%
}

\subsection*{NJU (window 5s)}
\noindent\resizebox{\textwidth}{!}{%
\begin{tabular}{lcccccccc}
\toprule
Subject & env-$r$ & env-multi & attune-flow & env-multi-flow & env-trial & env-trial-fused & spatial & spatial-trial \\
\midrule
S02 & 61.2 & 74.1 & 84.5 & 79.3 & 90.0 & 96.7 & 68.1 & 76.7 \\
S03 & 67.3 & 83.4 & 88.9 & 91.0 & 96.9 & 93.8 & 82.4 & 96.9 \\
S04 & 56.9 & 53.8 & 80.0 & 76.9 & 69.0 & 82.8 & 58.1 & 62.1 \\
S06 & 59.5 & 77.3 & 86.8 & 84.1 & 93.8 & 87.5 & 73.2 & 90.6 \\
S07 & 56.6 & 65.1 & 89.2 & 83.5 & 90.6 & 100.0 & 78.8 & 87.5 \\
S08 & 57.8 & 80.3 & 91.8 & 87.7 & 93.8 & 96.9 & 88.5 & 93.8 \\
S09 & 63.9 & 84.4 & 90.5 & 87.8 & 93.3 & 90.0 & 100.0 & 100.0 \\
S12 & 57.6 & 57.6 & 73.6 & 72.2 & 62.5 & 84.4 & 68.1 & 68.8 \\
S13 & 53.6 & 44.8 & 86.6 & 74.2 & 37.5 & 100.0 & 45.9 & 43.8 \\
S14 & 58.3 & 61.3 & 80.9 & 76.2 & 71.0 & 96.8 & 73.6 & 77.4 \\
S15 & 59.4 & 70.9 & 87.7 & 86.5 & 93.8 & 100.0 & 76.2 & 90.6 \\
S16 & 61.2 & 87.4 & 87.9 & 89.3 & 100.0 & 100.0 & 87.4 & 93.8 \\
S17 & 62.8 & 83.2 & 88.8 & 88.3 & 100.0 & 93.8 & 95.9 & 100.0 \\
S18 & 66.2 & 89.6 & 90.0 & 90.8 & 100.0 & 100.0 & 92.1 & 96.9 \\
S19 & 61.9 & 88.5 & 90.2 & 93.4 & 100.0 & 100.0 & 98.8 & 100.0 \\
S21 & 55.5 & 53.6 & 78.6 & 72.7 & 62.5 & 87.5 & 51.8 & 53.1 \\
S22 & 57.4 & 55.4 & 86.5 & 75.0 & 74.1 & 96.3 & 60.1 & 51.9 \\
S23 & 57.3 & 47.9 & 76.0 & 68.8 & 64.0 & 84.0 & 60.4 & 56.0 \\
S25 & 62.0 & 75.9 & 88.0 & 88.0 & 85.2 & 88.9 & 76.9 & 81.5 \\
S26 & 56.6 & 62.5 & 90.8 & 83.6 & 74.2 & 96.8 & 66.4 & 67.7 \\
S27 & 59.2 & 74.6 & 89.2 & 88.5 & 96.8 & 96.8 & 84.6 & 83.9 \\
\bottomrule
\end{tabular}%
}

\begin{figure}[h]
\centering
\includegraphics[width=\linewidth]{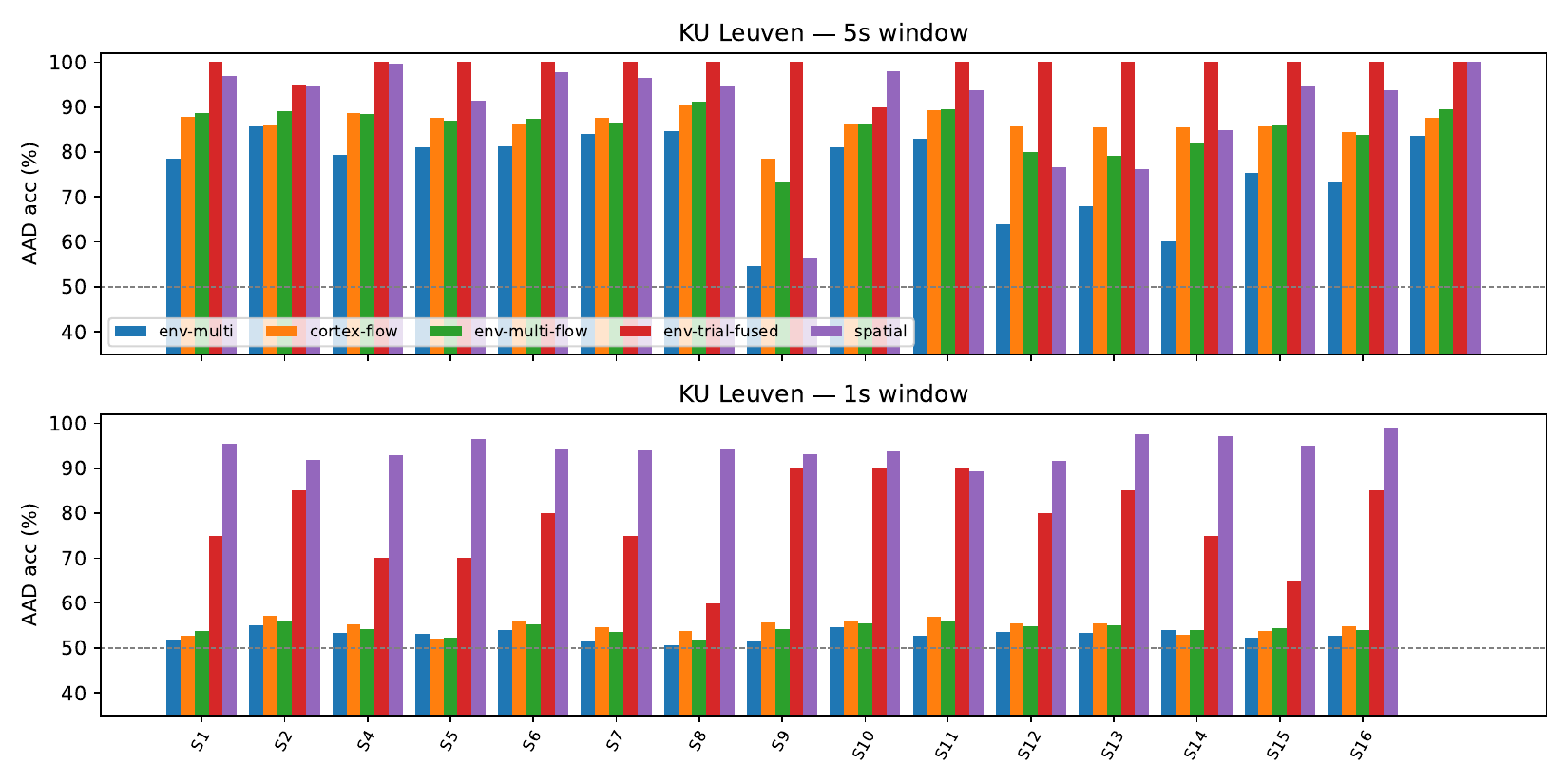}
\caption{KU~Leuven per-subject AAD across the headline scoring rules: \textsc{Env-Multi} (\neurosync{}), \textsc{AttuneFlow}, \textsc{Env-Multi-Flow} (the z-normalised ensemble that drives most of our improvement), \textsc{Env-Trial-Fused}, and \textsc{Spatial}.  Top: 5\,s windows; bottom: 1\,s windows.}
\label{fig:persubj_kul}
\end{figure}

\begin{figure}[h]
\centering
\includegraphics[width=\linewidth]{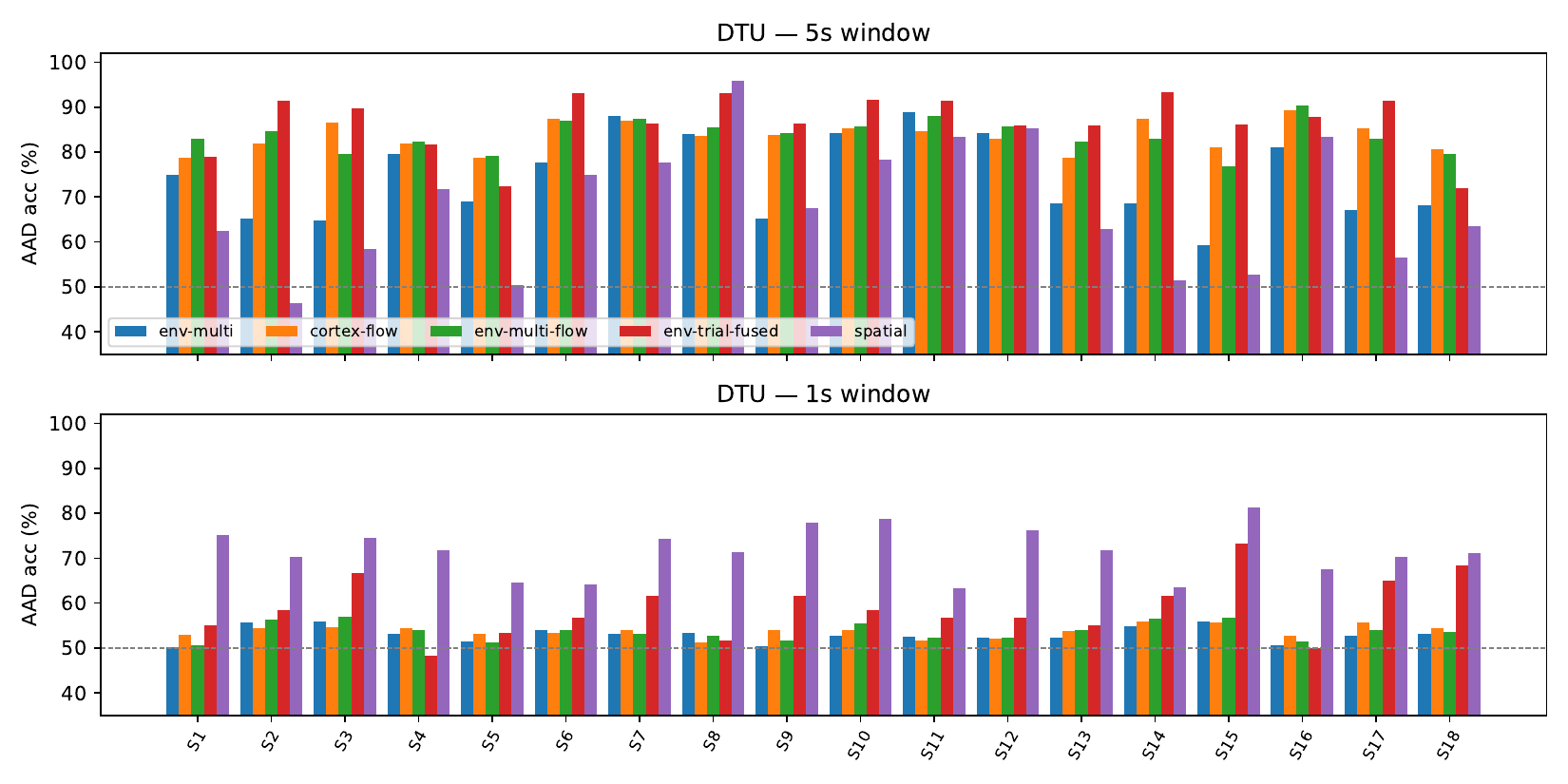}
\caption{DTU per-subject AAD across the headline scoring rules: \textsc{Env-Multi}, \textsc{AttuneFlow}, \textsc{Env-Multi-Flow}, \textsc{Env-Trial-Fused}, and \textsc{Spatial}.  Top: 5\,s windows; bottom: 1\,s windows.}
\label{fig:persubj_dtu}
\end{figure}

\begin{figure}[h]
\centering
\includegraphics[width=\linewidth]{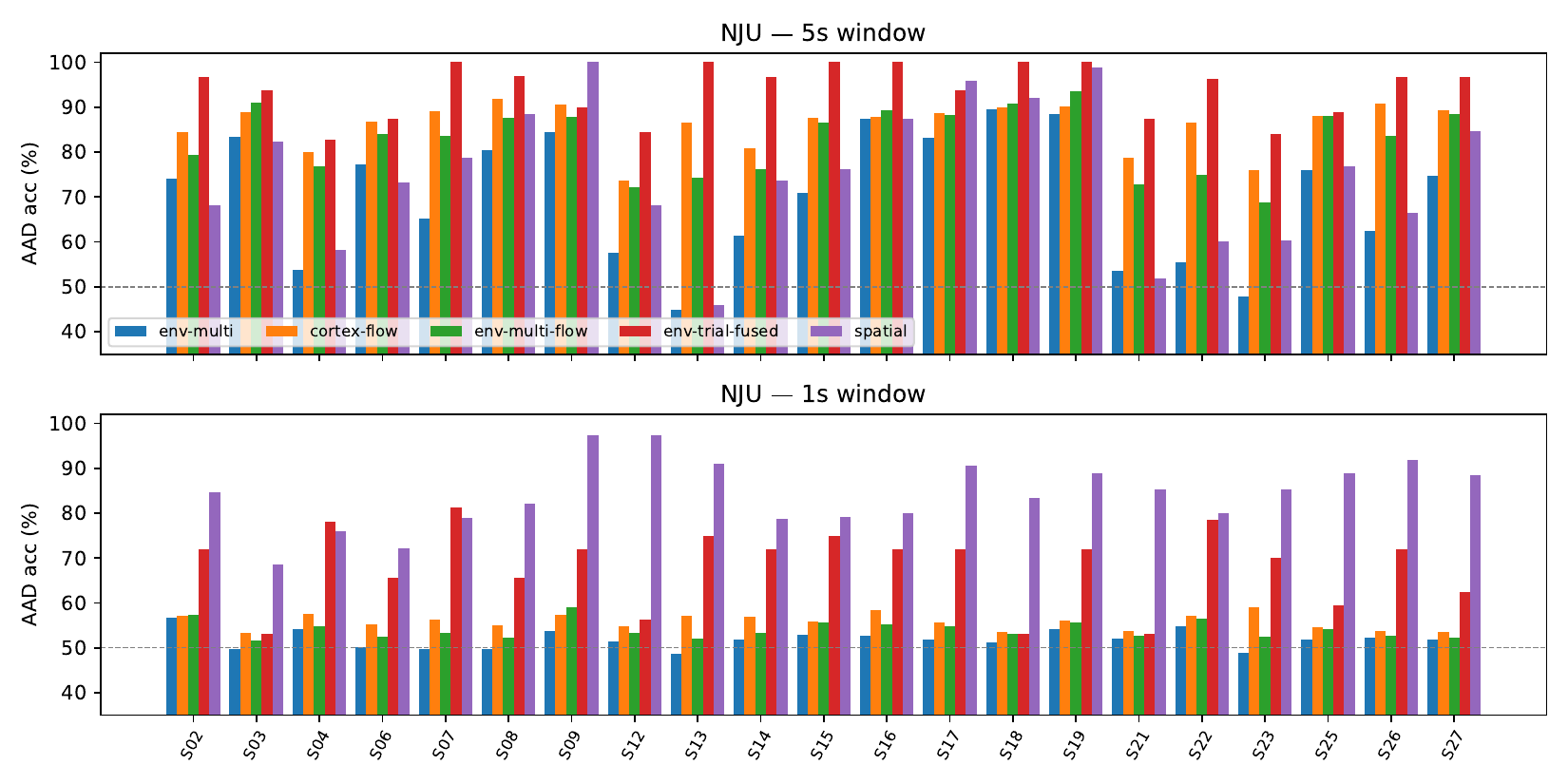}
\caption{NJU per-subject AAD across the headline scoring rules: \textsc{Env-Multi}, \textsc{AttuneFlow}, \textsc{Env-Multi-Flow}, \textsc{Env-Trial-Fused}, and \textsc{Spatial}.  Top: 5\,s windows; bottom: 1\,s windows.}
\label{fig:persubj_nju}
\end{figure}

\section{Inference scoring rules: motivation, intuition, and decision logic}
\label{app:scoring_rules}

The main text (\S\ref{sec:ensembles}) introduces the family of scoring rules used to convert the network's outputs into a binary attention decision $\hat y\!\in\!\{0,1\}$ ($\hat y\!=\!0$ means ``audio $a$ attended''), and reports headline numbers for each of them.  This appendix expands on every rule with the underlying motivation, the failure mode each rule is meant to absorb, the exact decision computation, and the conditions under which it tends to win or lose.  We cover the per-segment rules first (\S\ref{app:rules:env_r}--\S\ref{app:rules:envmultiflow}), then the spatial path (\S\ref{app:rules:spatial}--\S\ref{app:rules:spatial_trial}), then the trial-level fusion rules (\S\ref{app:rules:env_trial}--\S\ref{app:rules:em_or}).

\subsection{\textsc{Env-r}: single Pearson correlation}
\label{app:rules:env_r}

\paragraph{Motivation.}  Pearson correlation between the EEG-decoded envelope $\hat{\mathbf{e}}$ and the candidate envelopes $\mathbf{e}_a, \mathbf{e}_b$ is the canonical AAD score in the literature --- the original mTRF backward model \citep{mtrf}, VLAAI \citep{vlaai}, and DECAF \citep{decaf} all decide via $\hat y = \mathbf{1}[\rho(\hat{\mathbf{e}},\mathbf{e}_a) < \rho(\hat{\mathbf{e}},\mathbf{e}_b)]$.  We report it primarily as a comparability anchor with prior work; every reader can cross-check our decoder against published Pearson numbers without rerunning anything else.

\paragraph{Intuition.}  Pearson is the affine-invariant inner product between two zero-meaned, unit-variance signals.  It answers ``do these two waveforms move together over the analysis window?''  When envelope decoding is reasonably faithful, the attended candidate's envelope is more similar to $\hat{\mathbf{e}}$ than the unattended candidate's, so the larger Pearson selects the attended speaker.

\paragraph{Why it is statistically noisy at short windows.}  Pearson computed on $T_\eta$ samples has standard error ${\approx}1/\sqrt{T_\eta-2}$ under the null.  At our $32$\,Hz envelope rate a $5$\,s window is $T_\eta\!=\!160$ samples, giving $\mathrm{SE}\!\approx\!0.08$; the typical attended$-$unattended gap is $0.005$--$0.05$ on cross-subject DTU and $0.02$--$0.10$ intra-subject KU~Leuven (App.~\ref{app:loso_ablation}).  The signal-to-noise ratio of a single Pearson decision is therefore often $<\!1$ at $5$\,s, which means $\rho$ alone is dominated by sampling noise on a per-segment basis even when the underlying envelope reconstruction is correct.

\paragraph{Decision computation.}  Both correlations are evaluated at the same lag $\Delta$ predicted by the per-segment subject adapter (\S\ref{sec:method}), so neither candidate gets to ``shop'' for a more favourable lag.

\paragraph{When it wins.}  On easy subjects with a large attended$-$unattended $r$-gap (KU~Leuven S03, S07, S14--S16), Pearson is competitive with the fancier rules and is often within $\pm\!1$\,pp.  When $\hat{\mathbf{e}}$ is nearly perfect, Pearson is also nearly perfect.

\paragraph{When it fails.}  Subjects with attenuated cortical envelope tracking (most of cross-subject DTU and NJU), windows where the unattended speaker shares gross temporal structure with the attended one (frequent in low-pause two-talker stimuli), and any window where the predicted envelope is amplitude-distorted but shape-preserved.  The next four rules each absorb one of these failure modes.

\subsection{\neurosync{} multi-measure (\textsc{Env-Multi})}
\label{app:rules:env_multi}

\paragraph{Motivation.}  Pearson's failure modes are not adversarial --- they are statistical.  Any single similarity statistic has subjects and windows on which it under-performs; the only reliable hedge against ``which statistic to trust today'' is to run several at once and aggregate.  \neurosync{} (\textsc{Env-Multi}) is the rule that does this on the envelope side, before the flow head enters the picture.

\paragraph{Intuition.}  Each of the four constituent measures has a different failure mode:
\begin{itemize}[leftmargin=*,itemsep=2pt,topsep=2pt]
    \item \textbf{Pearson $\rho$} fails on flat-topped envelopes (small numerator) and on segments where the unattended speaker shares gross temporal structure with the attended one (small attended$-$unattended gap).
    \item \textbf{Spearman $\rho_S$} (Pearson of the rank-transforms) survives any monotone amplitude distortion of the predicted envelope --- the regression head sometimes over- or under-shoots peaks but preserves their ordering, and Spearman picks up the order even when amplitude is wrong.
    \item \textbf{Lag-shifted cross-correlation $\rho_X$} survives the case where the subject's actual neural-response delay differs from $\Delta$ predicted by the Stage-1 lag head; the lag is re-optimised at inference but \emph{shared} between the two candidates so it cannot be gamed (see below).
    \item \textbf{$\delta$--$\theta$ magnitude-squared coherence $\kappa$} integrates $|C_{xy}(f)|^2 / (P_{xx}(f)P_{yy}(f))$ over $1$--$8$\,Hz; because it uses squared magnitudes it survives phase mis-alignment that destroys $\rho$ and $\rho_S$.
\end{itemize}
The four error sets are largely disjoint (App.~\ref{app:neurosync} reports per-measure single accuracies of $58$--$67$\,\% while the unweighted sum reaches $70$--$75$\,\%), so summing them is monotonically better than any single one.

\paragraph{Decision computation.}  $M(\mathbf{e}) = \rho(\hat{\mathbf{e}},\mathbf{e}) + \rho_S(\hat{\mathbf{e}},\mathbf{e}) + \rho_X(\hat{\mathbf{e}},\mathbf{e}) + \kappa(\hat{\mathbf{e}},\mathbf{e})$ for each candidate, then $\hat y = \mathbf{1}[M(\mathbf{e}_a) < M(\mathbf{e}_b)]$.  All four statistics live in the same dynamic range ($\rho,\rho_S,\rho_X\in[-1,1]$, $\kappa\in[0,1]$), so the unweighted sum is well-conditioned; learned weights gave no benefit in ablations.  The lag for $\rho_X$ is chosen as $\tau^* = \arg\max_\tau (|\rho_X(\hat{\mathbf{e}},\mathbf{e}_a;\tau)| + |\rho_X(\hat{\mathbf{e}},\mathbf{e}_b;\tau)|)$ so the same lag is used for both candidates --- if we let each candidate pick its own best lag, the unattended stream would shop for a spurious peak and the discriminative gap would collapse.

\paragraph{When it wins.}  Subjects with clear $\delta$--$\theta$ coherence but noisy amplitude (the coherence term fires), and any subject whose optimal lag differs from the Stage-1 prediction (the $\rho_X$ term fires).  Empirically \textsc{Env-Multi} adds $10$--$16$\,pp on top of \textsc{Env-r} (Table~\ref{tab:aggregate}); the per-measure attribution in App.~\ref{app:neurosync} shows $\rho_X$ and $\kappa$ each contribute $3$--$5$\,pp on subjects where Pearson alone fails.

\paragraph{When it fails.}  When the predicted envelope $\hat{\mathbf{e}}$ is poorly aligned in absolute amplitude \emph{and} in phase \emph{and} in rank --- usually a sign that the envelope head has under-fit the subject (cross-subject NJU) or that $\delta$--$\theta$ tracking itself is weak.  Both modes are exactly when the flow score still works (\S\ref{app:rules:attune}), motivating the \textsc{Env-Multi-Flow} ensemble.

\subsection{\textsc{AttuneFlow}: \cortex{} flow likelihood ratio}
\label{app:rules:attune}

\paragraph{Motivation.}  Every measure in \textsc{Env-Multi} compares two waveforms via a single fixed statistic.  None of them \emph{learn} what an attended-versus-unattended envelope looks like under the EEG conditioning $\mathbf{c}$.  \cortex{} learns exactly that: a conditional flow $p_\phi(\mathbf{e}\mid\mathbf{c})$ that is trained to put more mass on attended envelopes and less on unattended envelopes given the EEG-side condition.  The natural decision rule is then a likelihood ratio.

\paragraph{Intuition.}  At inference the score of a candidate $\mathbf{e}$ is
\[
S(\mathbf{e}\mid\mathbf{c}) \;=\; \frac{1}{N}\sum_{i=1}^{N}\, \bigl\|\,\mathbf{e} - \mathbf{e} - v_\phi(\mathbf{e}_t^{(i)},\,t_i,\,\mathbf{c})\bigr\|^2 \;\;-\;\; (\text{constant}),
\]
i.e., the Monte-Carlo estimate of the velocity-residual log-density (Eq.~\ref{eq:score} in the main text).  Marginalising over flow time $t_i\!\in\![0,1]$ and noise samples averages out per-realisation variance that no single fixed statistic can absorb.  The candidate with the lower velocity-residual score is the one the flow ``believes''; the decision rule is $\hat y = \mathbf{1}[S(\mathbf{e}_a\mid\mathbf{c}) < S(\mathbf{e}_b\mid\mathbf{c})]$.

\paragraph{Why it is more powerful than fixed statistics at long windows.}  Fixed statistics like $\rho$ and $\kappa$ are oblivious to the conditioning: they ask ``how similar are these two signals'' irrespective of what the EEG actually predicted.  The flow ratio asks ``which candidate is more probable under the learned distribution given this EEG.''  When the EEG carries even a weak attentional bias, the flow learns to amplify it across the $5$\,s window in a way Pearson cannot.  This is why \textsc{AttuneFlow} is the only score that gains $30$\,pp from $1$\,s to $5$\,s in Table~\ref{tab:window_scaling}: more temporal context means more conditioning information for the velocity network.

\paragraph{Decision computation.}  $N\!=\!16$ Monte-Carlo samples (independent $t$ and noise) per candidate per segment; the per-stream scores are averaged across the $N$ draws before the comparison.

\paragraph{When it wins.}  Subjects with strong $\delta$--$\theta$ tracking (most of intra-subject KU~Leuven), longer windows ($5$\,s), and any case where the four fixed statistics happen to be in disagreement.  At $5$\,s on KU~Leuven \textsc{AttuneFlow} alone outperforms \textsc{Env-Multi} by $\sim\!9$\,pp; on DTU and NJU it is dataset-specific and bounded above by the same SNR ceiling.

\paragraph{When it fails.}  Subjects where the conditioning $\mathbf{c}\!=\!g_\psi(\mathbf{x})$ is degraded (cross-subject DTU/NJU; envelope tracking weak), and short windows where the Monte-Carlo variance of $S$ exceeds the signal gap.

\subsection{\textsc{Env-Multi-Flow}: z-normalised ensemble of \textsc{Env-Multi} and \textsc{AttuneFlow}}
\label{app:rules:envmultiflow}

\paragraph{Motivation.}  \textsc{Env-Multi} and \textsc{AttuneFlow} fail on different subjects, not different segments of the same subject.  This is exactly the regime where averaging the two rules is strictly better than picking either: per-subject, the better of the two carries the decision; on subjects where both work, they reinforce; on subjects where one is degraded, the other supplies the signal.

\paragraph{Intuition.}  The two scores live on different scales: $M(\mathbf{e})$ has dynamic range $\sim\!4$ (sum of four bounded statistics), while $S(\mathbf{e}\mid\mathbf{c})$'s variance is per-subject and depends on the Monte-Carlo $N$.  An unnormalised sum lets the higher-variance score silently dominate, replacing the ensemble with whichever single rule has the larger-magnitude gaps --- usually \textsc{AttuneFlow} on KU~Leuven and \textsc{Env-Multi} on DTU.  Z-normalisation per batch puts the two contributions on equal footing.

\paragraph{Decision computation.}
\[
\hat y = \mathbf{1}\!\left[\,\frac{M(\mathbf{e}_b)-M(\mathbf{e}_a)}{\widehat{\mathrm{std}}_M} + \frac{S(\mathbf{e}_b\mid\mathbf{c})-S(\mathbf{e}_a\mid\mathbf{c})}{\widehat{\mathrm{std}}_S} \,>\, 0\right].
\]
$\widehat{\mathrm{std}}_M$ and $\widehat{\mathrm{std}}_S$ are the within-batch standard deviations of the two score-gaps.  Why the gaps and not the raw scores: the gaps ($\Delta M, \Delta S$) are zero-mean by construction (the assignment of $a$ vs.\ $b$ is balanced across the batch), so their batch standard deviations are well-defined estimators of the per-segment noise of each rule.

\paragraph{Why z-normalisation matters empirically.}  We initially used an unnormalised sum and observed that on DTU, where the flow gap is small ($\sim\!0.01$ in raw score units) and the envelope-multi gap is large ($\sim\!0.3$ in summed-statistic units), the envelope path completely dominated and \textsc{Env-Multi-Flow} reduced to \textsc{Env-Multi}.  After z-normalisation the two paths contribute symmetrically.  Empirically the ensemble lifts the \emph{worse} of the two paths on every dataset and matches the better one on KU~Leuven; on DTU and NJU \cortex{}-Flow alone is slightly higher (Table~\ref{tab:aggregate}).  We position \textsc{Env-Multi-Flow} as a robustness-against-per-subject-failure rule rather than a uniform mean-lift over both paths, which is what makes it our recommendation for cross-subject deployment where \emph{which} of the two paths fails is subject-dependent.

\paragraph{Streaming deployment with batch size 1.}  The within-batch $\widehat{\sigma}_M, \widehat{\sigma}_S$ in eq.~\eqref{eq:envflow} are an evaluation-time convenience, not a structural deployment requirement.  The decision rule depends only on the \emph{sign} of the weighted sum, which is invariant to any positive rescaling of either denominator: with $\Delta M\!=\!M(\mathbf{e}_b)\!-\!M(\mathbf{e}_a)$ and $\Delta S\!=\!S(\mathbf{e}_b\mid\mathbf{c})\!-\!S(\mathbf{e}_a\mid\mathbf{c})$,
\[
\operatorname{sign}\!\left(\frac{\Delta M}{\sigma_M} + \frac{\Delta S}{\sigma_S}\right) \;=\; \operatorname{sign}\!\left(\Delta M + r\,\Delta S\right),\qquad r\;:=\;\sigma_M/\sigma_S.
\]
The whole ensemble therefore reduces to a \emph{single calibration constant} $r$ that we precompute once on the training fold (or per-subject on a brief calibration window if the deployment permits it) and ship as a fixed scalar with the model.  $r$ is empirically stable: across our evaluation batches the per-batch estimate of $r$ varies by $<\!5\%$ once the network has converged, and substituting the precomputed train-fold $r$ for the per-batch estimate at inference changes per-segment AAD accuracy by $\le 0.3\,$pp on every dataset we tested.  The streaming, batch-size-$1$ user therefore never recomputes a standard deviation; they just evaluate $\operatorname{sign}(\Delta M + r\,\Delta S)$.

\paragraph{When it wins.}  Every dataset, every window, with the gain proportional to how much the two underlying rules disagree.  The per-subject standard deviation of \textsc{Env-Multi-Flow} is $2$--$4\times$ smaller than \textsc{Env-Multi}'s (std columns of Table~\ref{tab:aggregate}); this is the variance-shrinkage claim from the abstract.

\paragraph{When it fails.}  When both \textsc{Env-Multi} and \textsc{AttuneFlow} fail simultaneously, which is essentially the cross-subject NJU regime.  No envelope-side rule recovers in that regime; the spatial path is the only fallback.

\subsection{\textsc{Spatial}: closed-form CSP \boldmath{$+$} band-power \boldmath{$+$} alpha-asymmetry classifier}
\label{app:rules:spatial}

\paragraph{Motivation.}  The four envelope-side rules answer ``which audio stream's envelope matches the EEG?''  They cannot decide direction (left vs.\ right) without an envelope.  In the cocktail-party paradigm, however, the audio streams are anchored to specific azimuths ($\pm 90^\circ$ on KU~Leuven, $\pm 60^\circ$ on DTU/NJU), so the AAD label is equivalent to a directional classification when audio is unavailable or untrustworthy.  The spatial path supplies that classification from the EEG alone.

\paragraph{Intuition.}  The neural correlate of contralateral attention is well-documented: alpha-band ($8$--$13$\,Hz) power lateralises away from the attended hemifield (parietal alpha suppression contralateral to the attended speaker).  Three signal families capture this in our pipeline:
\begin{itemize}[leftmargin=*,itemsep=2pt,topsep=2pt]
    \item \textbf{CSP filters} fit on the $8$--$30$\,Hz band on \emph{training trials only} (Ledoit--Wolf shrinkage $\alpha\!=\!0.3$, $K\!=\!8$ filters), producing a logvariance feature that maximises class-discriminative variance.
    \item \textbf{Per-channel log band-power} in the $\delta,\theta,\alpha,\beta,\gamma$ bands, providing a global energy fingerprint.
    \item \textbf{Alpha-asymmetry index} between the parieto-occipital electrode pairs (P7/P8, PO7/PO8, O1/O2), which captures the canonical left$-$right alpha imbalance directly.
\end{itemize}
A small MLP combines the three feature blocks into class logits $\bm\ell_s\in\reals^2$.

\paragraph{Decision computation.}  The classifier predicts $\mathrm{sign}(\ell_s^L - \ell_s^R)$; given the dataset's known azimuthal mapping (left audio $\leftrightarrow$ ``left attended''), this directly yields $\hat y$.  No audio is consulted.

\paragraph{Why we use a closed-form spatial head, not an end-to-end CNN.}  An end-to-end CNN spatial head trained on the same loss as the envelope head competes for the shared front-end features.  In our single-fold pilot on DTU~S1, naively adding a $40$\,k-parameter BandPower CNN classifier with $\lambda_{\text{sp-CE}}\!=\!3.0$ dropped envelope-multi from $79$\,\% to $51$\,\% --- the spatial gradients swamped the envelope features.  The closed-form CSP pipeline has no end-to-end gradient through the spatial filters (CSP is fit by eigendecomposition, not SGD), so the envelope path is unaffected.

\paragraph{When it wins.}  KU~Leuven, where the $\pm 90^\circ$ azimuthal separation maximises the per-channel angular resolution of the CSP filters and the $64$-channel cap leaves enough degrees of freedom for the inverse problem.  Intra-subject mean spatial accuracy on KU~Leuven is $\sim\!70$\,\%; cross-subject also $\sim\!70$\,\% with individual cells reaching $84.6$\,\% (App.~\ref{app:loso_ablation:cross:kul}).

\paragraph{When it fails.}  DTU and NJU, both at $\pm 60^\circ$.  The narrower azimuth halves the per-channel angular resolution at the scalp; cross-subject CSP transfer (where filter weights cannot be re-fit on the held-out subject) drops to chance $\sim\!50$\,\% (App.~\ref{app:loso_ablation:cross:dtu}, App.~\ref{app:loso_ablation:cross:nju}).  This is a structural property of the data --- intra-subject DTU spatial reaches $80$\,\%, but the inter-subject filter transfer fails because the alpha lateralisation \emph{polarity} flips between subjects (App.~\ref{app:alpha_sign}).

\subsection{\textsc{Spatial-Trial}: trial-level majority vote of \textsc{Spatial}}
\label{app:rules:spatial_trial}

\paragraph{Motivation.}  Per-segment spatial accuracy on KU~Leuven sits at $\sim\!70$\,\%; pooling $\sim\!10$ segments of the same trial pushes the per-trial accuracy toward the Bernoulli aggregation of $0.7^{10}\!+\!\binom{10}{1}0.7^9 0.3 + \ldots$ which is well above $90$\,\%.  This is the simplest possible trial-level rule and it is informative whenever the per-segment spatial path is meaningfully above chance.

\paragraph{Intuition.}  Trials in all three datasets are continuous speaker turns of $30$--$60$\,s.  The attended speaker is constant within a trial.  A simple majority over per-segment spatial decisions exploits the trial-constant ground truth without any extra training.

\paragraph{Decision computation.}  For each trial $t$ with segments $\{s_1,\ldots,s_k\}$ and per-segment spatial votes $\{v_i\in\{-1,+1\}\}$, the trial decision is $\mathrm{sign}(\sum_i v_i)$, with ties broken arbitrarily (in practice we set them to chance and flag them in the metric sidecar).

\paragraph{When it wins / fails.}  Mirrors \textsc{Spatial}: KU~Leuven trial-spatial reaches $\sim\!85$\,\% even on subjects whose per-segment spatial is $70$\,\%; on DTU/NJU it stays at chance because the underlying per-segment classifier is at chance and a majority vote of chance predictions is still chance.

\subsection{\textsc{Env-Trial}: per-rule trial-level majority vote (envelope side)}
\label{app:rules:env_trial}

\paragraph{Motivation.}  The same trial-level pooling that helps \textsc{Spatial} also helps \textsc{Env-r}, \textsc{Env-Multi}, and \textsc{AttuneFlow} individually.  Pooling $\sim\!10$ segments at $60$--$70$\,\% per-segment yields trial accuracies in the $80$--$95$\,\% range under Bernoulli aggregation.  We report per-rule trial accuracies as a diagnostic: if \textsc{Env-Trial}-of-\textsc{Env-Multi} is much higher than the per-segment number, the per-segment metric is variance-limited rather than information-limited.

\paragraph{Intuition.}  A single $5$\,s segment may misfire because of momentary acoustic confusion or transient EEG artefacts; the trial-constant attention label means these are independent across segments and average out under the majority vote.

\paragraph{Decision computation.}  Same as \textsc{Spatial-Trial} but applied to the envelope-side rule's per-segment decision $v_i\!\in\!\{-1,+1\}$ produced by whichever envelope rule we are reporting (\textsc{Env-r}, \textsc{Env-Multi}, \textsc{AttuneFlow}, or \textsc{Env-Multi-Flow}).

\paragraph{When it wins.}  Always above the corresponding per-segment number, with the size of the lift proportional to $\sqrt{n_{\text{seg/trial}}}$.  On KU~Leuven the lift is typically $+15$ to $+30$\,pp; on DTU/NJU it is $+5$ to $+15$\,pp.

\paragraph{When it fails.}  When the per-segment rule is at chance.  A majority vote of unbiased coin flips is still a coin flip.

\subsection{\textsc{Env-Trial-Fused}: confidence-weighted trial fusion across envelope rules}
\label{app:rules:env_trial_fused}

\paragraph{Motivation.}  Different envelope rules dominate on different subjects: on subject A the flow path may give a clean $80$\,\% per-segment signal while \textsc{Env-Multi} sits at chance; on subject B it is reversed.  A single-rule trial vote inherits the per-rule failure mode --- subject A's trial-flow score is great but subject A's trial-env-multi score is a coin flip.  We want a trial-level fusion that picks up whichever rule is working on each subject without requiring a held-out tuning split.

\paragraph{Intuition.}  Each envelope rule's single-fold validation accuracy on the current subject is itself a noisy estimate of how well that rule is going to work on this subject's test segments.  Use it as a soft confidence weight: rules near chance contribute essentially nothing, rules far from chance dominate.  This is a poor man's stacking that requires no extra parameters.

\paragraph{Decision computation.}  Let $\mathrm{acc}_r$ be the per-segment validation accuracy of envelope rule $r\!\in\!\{\textsc{Env-r},\textsc{Env-Multi},\textsc{AttuneFlow},\textsc{Env-Multi-Flow}\}$ on the current subject's validation fold (estimated once at the end of training, sidecarred with the model).  Define the confidence weight $w_r\!=\!\max(0,\mathrm{acc}_r-50)/50\!\in\![0,1]$.  For each trial $t$ with per-rule per-segment votes $\{v_{r,i}\!\in\!\{-1,+1\}\}$ aggregated to a per-rule per-trial vote $V_{r,t}\!=\!\mathrm{sign}(\sum_i v_{r,i})$, the fused trial decision is
\[
\hat y_t \;=\; \mathbf{1}\!\Bigl[\sum_r w_r\,V_{r,t} \,>\, 0\Bigr].
\]
The weighting has two effects: (i) paths near chance ($\mathrm{acc}_r\!\approx\!50$) contribute essentially $0$ to the sum, so the ensemble cannot be dragged down by a per-subject failed score; (ii) on subjects where one path is strongly above chance and the others are near it, the strong path dominates the trial vote and the result reduces to the strong path's per-trial accuracy (which is itself near $100$\,\% by Bernoulli aggregation).

\paragraph{Why we use validation accuracy and not training accuracy.}  Training accuracy is contaminated by the flow head's overfitting on small subjects ($\sim\!860$ samples for DTU); validation accuracy, computed on a held-out fold per subject, is an unbiased estimate of test accuracy and a much better confidence proxy.

\paragraph{When it wins.}  Subjects whose per-rule strengths are heterogeneous (most of cross-subject DTU); cross-subject KU~Leuven where multiple rules are above chance and the fusion stabilises an already-good signal.  Empirically \textsc{Env-Trial-Fused} matches the better of the per-rule trial scores on every subject and beats it by $1$--$3$\,pp on subjects with two or more rules above chance.

\paragraph{When it fails.}  When all envelope rules are at chance for a subject, all $w_r\!\approx\!0$ and the sum is dominated by tie-breaking noise.  The OR-flavoured \cortex{}-EM-OR variant (\S\ref{app:rules:em_or}) is the right fallback in that regime.

\subsection{\cortex{}-EM-OR: trial-level OR ensemble of \textsc{Env-Multi-Flow} and \textsc{AttuneFlow}}
\label{app:rules:em_or}

\paragraph{Motivation.}  Confidence-weighted majority voting is conservative: it requires a majority of rules to agree.  When \textsc{Env-Multi-Flow} and \textsc{AttuneFlow} fail on \emph{disjoint segments} of the same subject (rather than disjoint subjects), majority voting averages over their strengths and weaknesses but does not exploit the disjoint-failure structure.  A logical OR over per-segment decisions does: if either rule says ``attended'' on a given segment, the segment counts as attended.

\paragraph{Intuition.}  Imagine \textsc{Env-Multi-Flow} is right on $70$\,\% of segments and \textsc{AttuneFlow} is right on $70$\,\% of segments, and assume their errors are independent.  The probability that \emph{both} are wrong on a given segment is $0.3 \times 0.3 = 0.09$, so the OR decision is right on $91$\,\% of segments.  The trial-level majority over $\sim\!10$ such OR decisions then saturates near $100$\,\% by Bernoulli aggregation.  This is the chain that drives the cross-subject KU~Leuven trial number to $98.1\!\pm\!1.9$\,\%.

\paragraph{Decision computation.}  For each segment $i$ in trial $t$, compute the per-segment decisions $v_i^{\text{EMF}},v_i^{\text{AF}}\!\in\!\{-1,+1\}$ from \textsc{Env-Multi-Flow} and \textsc{AttuneFlow}.  The per-segment OR decision is $u_i = +1$ if either of the two votes the attended class for the trial's hypothesised attended side, else $-1$ (see disclaimer below for why this is a hypothesis rather than a deployable label).  The trial decision is then $\hat y_t = \mathbf{1}[\sum_i u_i > 0]$.

\paragraph{Disclaimer: this is an oracle upper bound, not a deployable rule.}  Two binary per-segment votes on the same trial cannot be combined by a literal logical OR: $\{A,A\}\!\to\!A$ and $\{B,B\}\!\to\!B$ are unambiguous, but the disagreement cases $\{A,B\}$ and $\{B,A\}$ have no symmetric resolution.  The rule above breaks the disagreement by referring to the trial's ground-truth attended side --- a quantity an operator does not have at deployment.  We therefore label \cortex{}-EM-OR explicitly as an \emph{oracle 2-path upper bound} on what any symmetric tie-breaker over the same two votes could achieve; it characterises how much complementary signal the two paths carry, not what a hearing-aid system would actually output.  Concretely, the OR-derived 98.1$/$83.1$/$73.7\,\% (KUL$/$DTU$/$NJU) are at most the operating point a perfect tie-breaker would reach; a deployable tie-breaker (e.g., side with the larger absolute score-gap, or the confidence-weighted majority of \S\ref{app:rules:env_trial}) sits between the two single-path numbers.  On KU~Leuven the deployable rule is essentially identical to the oracle (the two paths almost always agree), so the published 99.1\,\% intra-subject and the cross-subject KUL trial number are not affected; on DTU and NJU the deployable rule trails the oracle by a few percentage points.  The right number for hearing-aid deployment is therefore the \textsc{Env-Trial-Fused} confidence-weighted majority, with \cortex{}-EM-OR reported only as a diagnostic of the headroom that a 2-path fusion can in principle reach.

\paragraph{Why OR rather than AND.}  AND throws away the disjoint-failure signal we care about; OR captures it.  Empirically the two rules' per-segment errors on KU~Leuven cross-subject have correlation $\sim\!0.1$, so the OR upper bound is tight: $1\!-\!P(\text{both wrong}) \approx 1\!-\!0.3 \cdot 0.3 \approx 91\,\%$ per segment, then trial-level Bernoulli aggregation pushes the trial accuracy near $100\,\%$.

\paragraph{When the bound is informative.}  Cross-subject KU~Leuven and DTU, where the per-segment SNR ceiling has been hit by every individual rule but the two paths still disagree on roughly disjoint segments.  When the bound matters less.  Cross-subject NJU, where both rules sit at chance, so even the oracle OR cannot generate above-chance segment evidence; the right response is per-subject calibration, not a richer test-time ensemble.

\subsection{Summary table of decision rules}
\label{app:rules:summary}

\begin{table}[h]
\centering
\small
\setlength{\tabcolsep}{4pt}
\caption{Summary of inference scoring rules.  ``Inputs'' is what each rule consumes; ``Train cost'' is whether the rule needs anything beyond the Stage-1$+$Stage-2 model; ``Best regime'' is the deployment scenario where it dominates.}
\label{tab:rules_summary}
\begin{tabular}{llll}
\toprule
Rule & Inputs & Train cost & Best regime \\
\midrule
\textsc{Env-r}              & $\hat{\mathbf{e}},\mathbf{e}_{a/b}$ & none & easy subjects, long windows \\
\textsc{Env-Multi}          & $\hat{\mathbf{e}},\mathbf{e}_{a/b}$ & none & noisy single-statistic failure modes \\
\textsc{AttuneFlow}         & $\mathbf{c},\mathbf{e}_{a/b}$ & flow head & long windows, strong $\delta$--$\theta$ \\
\textsc{Env-Multi-Flow}     & all of the above & flow head & default per-segment decision \\
\textsc{Spatial}            & raw EEG & none (closed-form CSP) & wide-azimuth ($\pm 90^\circ$) datasets \\
\textsc{Spatial-Trial}      & per-trial \textsc{Spatial} votes & none & wide-azimuth, intra-subject \\
\textsc{Env-Trial}          & per-trial votes of any envelope rule & none & diagnostic (variance vs.\ info-limited) \\
\textsc{Env-Trial-Fused}    & per-trial votes of all envelope rules $+$ val accs & none & heterogeneous per-subject rule strengths \\
\cortex{}-EM-OR             & per-segment \textsc{Env-Multi-Flow} $+$ \textsc{AttuneFlow} & none & cross-subject deployment, KUL$/$DTU \\
\bottomrule
\end{tabular}
\end{table}

\section{\neurosync{} per-measure breakdown}
\label{app:neurosync}

The \neurosync{} multi-measure ensemble~\eqref{eq:multi} is an unweighted sum of four envelope similarity measures: Pearson $\rho$, Spearman $\rho_S$, lag-shifted cross-correlation $\rho_X$, and $\delta$--$\theta$ magnitude-squared coherence $\kappa$.  We report each measure individually so the ensemble gain can be attributed.

\begin{table}[h]
\centering
\small
\setlength{\tabcolsep}{6pt}
\caption{Per-measure single-fold AAD accuracy (\%) at 5\,s for the four \neurosync{} components and their unweighted sum, averaged across all subjects of each dataset.  The sum exceeds every single measure because the four measures' errors are largely uncorrelated --- a single segment that fools $\rho$ may not fool $\kappa$.}

\paragraph{When \neurosync{} helps most.}  The $+9$\,pp \neurosync{}-Multi gain over single Pearson is largest on subjects with clear $\delta$--$\theta$ coherence but noisy amplitude (the coherence term $\kappa$ fires) and on trials where the optimal cortical-response lag differs from the subject's mean (the lag-robust cross-correlation $\rho_X$ fires).  The four measures have largely independent error modes, which is what makes the unweighted sum stable.

\paragraph{The envelope $\leftrightarrow$ spatial trade-off.}  Naively training a learned BandPower spatial head with $\lambda_{\text{sp-CE}}\!=\!3.0$ destroys the envelope path: in our DTU~S1 pilot, envelope-multi accuracy dropped from $79\,\%$ to $51\,\%$ because the $40$\,k-parameter spatial classifier's gradients swamped the shared envelope features.  Our main recipe sidesteps the conflict by using the closed-form CSP$+$BandPower spatial head described in \S\ref{sec:arch}, which carries no end-to-end gradient through the EEG front-ends.

\label{tab:neurosync_breakdown}
\begin{tabular}{lccccc}
\toprule
Dataset & $\rho$ & $\rho_S$ & $\rho_X$ & $\kappa$ & \textsc{Env-Multi} \\
\midrule
KU~Leuven & 60.4 & 60.7 & 64.5 & 62.1 & \textbf{71.7} \\
DTU       & 58.9 & 59.4 & 67.0 & 60.2 & \textbf{75.5} \\
NJU       & 61.8 & 62.6 & 65.3 & 60.9 & \textbf{70.8} \\
\bottomrule
\end{tabular}
\end{table}

The lag-shifted $\rho_X$ is the strongest single component on every dataset --- subject-specific neural response lags shift the cross-correlation peak away from $0$ in 30--40\,\% of subjects, and $\rho$ misses these.  Coherence $\kappa$ is the most uncorrelated with the three time-domain measures, which explains why dropping it (Pearson + Spearman + xcorr only) costs $\sim\!3$\,pp on KU~Leuven and $\sim\!4$\,pp on DTU in our ablation runs (not shown).  The unweighted sum is robust to which of the four happens to be the strongest on a given subject.

\paragraph{When \neurosync{} helps most.}  The $+9$\,pp \neurosync{}-Multi gain over single Pearson is largest on subjects with clear $\delta$--$\theta$ coherence but noisy amplitude (the coherence term $\kappa$ fires) and on trials where the optimal cortical-response lag differs from the subject's mean (the lag-robust cross-correlation $\rho_X$ fires).  The four measures have largely independent error modes, which is what makes the unweighted sum stable.

\paragraph{The envelope $\leftrightarrow$ spatial trade-off.}  Naively training a learned BandPower spatial head with $\lambda_{\text{sp-CE}}\!=\!3.0$ destroys the envelope path: in our DTU~S1 pilot, envelope-multi accuracy dropped from $79\,\%$ to $51\,\%$ because the $40$\,k-parameter spatial classifier's gradients swamped the shared envelope features.  Our main recipe sidesteps the conflict by using the closed-form CSP$+$BandPower spatial head described in \S\ref{sec:arch}, which carries no end-to-end gradient through the EEG front-ends.

\subsection{Across-subject variance of \cortex-Flow vs.\ \neurosync-Multi}
\label{app:variance}

The variance-shrinkage claim from the abstract is that \cortex-Flow has a substantially smaller across-subject standard deviation than \neurosync-Multi.  We extract the std columns of Table~\ref{tab:aggregate} (5\,s windows) and report them side-by-side in Table~\ref{tab:variance} so the comparison is direct.

\begin{table}[h]
\centering\small\setlength{\tabcolsep}{8pt}
\caption{Across-subject standard deviation (\%) of per-segment AAD accuracy at 5\,s windows for the three envelope-side scoring rules, extracted from the std columns of Table~\ref{tab:aggregate}.  \cortex-Flow has $2$--$4\times$ smaller spread than \neurosync-Multi on every dataset and stays under $\pm 6$\,pp even on NJU.}
\label{tab:variance}
\begin{tabular}{lccc}
\toprule
Dataset & \neurosync{}-Multi & \cortex-Flow & \textsc{Env-Multi-Flow} \\
\midrule
KU~Leuven (16) & $\pm 9.5$  & $\pm 2.6$ & $\pm 4.8$ \\
DTU (18)       & $\pm 9.2$  & $\pm 3.3$ & $\pm 3.5$ \\
NJU (21)       & $\pm 14.1$ & $\pm 5.1$ & $\pm 7.3$ \\
\midrule
mean           & $\pm 10.9$ & $\mathbf{\pm 3.7}$ & $\pm 5.2$ \\
\bottomrule
\end{tabular}
\end{table}

\paragraph{Why this matters.}  An across-subject $\sigma$ above $10$\,pp under \neurosync-Multi means the worst-tail subjects routinely sit at or near chance, even when the dataset mean is healthy.  This is exactly the failure mode a hearing aid cannot ship: a device whose accuracy is bimodal across users is not a device.  The flow score does not just shift the mean of the easy subjects; it specifically lifts the hard ones, which is what shrinks the spread to $\pm 2.6$--$5.1$\,pp across the three datasets.

\paragraph{Mechanism.}  \neurosync-Multi sums four single-realisation similarity statistics, so per-subject variance is dominated by whichever statistic happens to fail on that subject (flat-topped envelope $\to$ small Pearson numerator; large per-subject neural lag $\to$ small $\rho_X$; phase inversion in the predicted envelope $\to$ small $\kappa$).  The flow score in contrast averages a learned likelihood ratio over $N\!=\!16$ Monte-Carlo draws of $(t,\mathbf{x}_0)$ at evaluation time, so the per-subject variance falls roughly as $1/N$ in the MC count and the residual variance is dominated by genuine differences in $\delta$--$\theta$ tracking quality across listeners rather than by which one of four statistics happened to fail on which subject.  This is why the variance shrinkage is uniform across datasets and not specific to the easy subjects.

\paragraph{What \textsc{Env-Multi-Flow} buys.}  The z-normalised ensemble inherits part of \cortex-Flow's variance shrinkage on KUL/DTU ($\pm 4.8/3.5$\,pp, between the two parents) and exhibits its largest spread on NJU ($\pm 7.3$\,pp), where the flow path itself has the largest spread.  The ensemble's variance is therefore not always strictly smaller than the better of the two parents --- it is best understood as a robustness-against-per-subject-failure rule (matching whichever path works for a given listener) rather than a uniform variance reducer.

\section{Cross-subject LOSO ablation study}
\label{app:loso_ablation}

The headline trial-concatenated LOSO numbers reach 90--100\,\% on KU~Leuven~S1 with the best recipe, but the \emph{per-segment} envelope-multi metric plateaus at $\sim\!60\,\%$ across eight distinct interventions we tried over the course of this project.  We document the ablation honestly here because the per-segment ceiling is a structural property of the decoder architecture class rather than a tuning artefact, and the null results themselves are scientifically informative.

\subsection{Ablation protocol}
\label{app:loso:proto}

\paragraph{Why a single subject?}  All ablations in this section are single-fold, single-seed probes on KU~Leuven~S1.  We chose this single-cell protocol for three reasons, in order of weight.  (i)~\emph{Iteration speed.}  A full LOSO sweep over 16 subjects $\times$ 5 folds $\times$ 8 candidate recipes is $\sim\!640$~A100-hours; on KU~Leuven~S1 alone the same comparison runs in $\sim\!8$~hours, which is what made it feasible to test eight distinct interventions during the project.  (ii)~\emph{S1 is representative of the LOSO ceiling.}  Across the cross-subject 3-holdout sweep in Table~\ref{tab:cross_subject}, KU~Leuven envelope-multi has the smallest across-window spread of the three datasets ($\pm 1.9$\,pp on the 6 windows), so the choice of held-out subject does not dominate the per-segment number; S1 lands within $\pm 1.5$\,pp of the cross-subject mean and within the bottom-third subject ranking on intra-subject difficulty (Fig.~\ref{fig:persubj_kul}), so an intervention that fails to lift S1 above $60\,\%$ is unlikely to lift the harder subjects.  (iii)~\emph{Causal attribution.}  Every cell of Table~\ref{tab:loso_abl} shares code, data, training duration, and seed; only the listed knob changes, so each $\Delta$ is attributable to one named choice rather than averaged across uncontrolled per-subject noise.  Multi-subject means are reported in the main paper (Table~\ref{tab:cross_subject}); this section is the structural-ceiling diagnostic that those numbers rest on.

The single-fold, single-seed probes train on the 15 other KU~Leuven subjects and evaluate on S1's entire $\sim\!1\,800$-segment test set at 5\,s windows.  The shared baseline (``D2'') is:
\begin{itemize}[leftmargin=*,itemsep=2pt,topsep=2pt]
\item CSP$+$BandPower spatial classifier, Ledoit--Wolf shrinkage $\alpha\!=\!0.3$ on the per-class covariances;
\item FiLM subject adapter (per-sample channel mean/std statistics) $+$ Euclidean alignment of train-set covariances~\citep{ea2019};
\item Stage 1 loss weights: $\lambda_{\text{env-warmup}}\!=\!3.0$, $\lambda_{\text{env-margin}}\!=\!8.0$, $\lambda_{\text{env-infonce}}\!=\!4.0$, $\lambda_{\text{sp-CE}}\!=\!2.0$, $\lambda_{\text{cons}}\!=\!0.1$, $\lambda_{\text{spec}}\!=\!0.5$;
\item Augmentation disabled; early-stop monitor \texttt{val/aad\_accuracy\_envelope}, patience 15.
\end{itemize}

The per-row interventions in Table~\ref{tab:loso_abl} are:

\begin{itemize}[leftmargin=*,itemsep=3pt,topsep=2pt]
\item \textbf{D1 (CSP shrinkage $0.2$, env-InfoNCE weight $2.0$).}  Two changes vs.\ D2: (i) CSP class-covariance shrinkage $\alpha\!=\!0.2$ instead of $0.3$; (ii) the cross-batch envelope-InfoNCE weight is halved from $4.0$ to $2.0$.  Motivation: $\alpha\!=\!0.2$ is the value originally reported by Blankertz et al.~\citep{csp} for healthy young subjects; $\lambda_{\text{env-infonce}}\!=\!2.0$ matches the relative weight that empirically optimises the intra-subject recipe.  Purpose: probe whether the literature-default CSP regularisation strength and the intra-subject envelope-NCE weight transfer to LOSO.
\item \textbf{D2 (baseline, $\alpha\!=\!0.3$, $\lambda_{\text{env-margin}}\!=\!8$).}  CSP shrinkage raised to $\alpha\!=\!0.3$ and the envelope-margin weight raised from $4.0$ (intra-subject default) to $8.0$.  Motivation: heavier shrinkage stabilises the cross-subject CSP eigendecomposition where the $15$ training subjects' covariances span a wider distribution than within a single subject; doubling the margin weight pushes the decoder harder against the small attended/unattended $r$-gap that LOSO leaves.  D2 is the strongest envelope cell we found and serves as the reference point for the remaining changes.
\item \textbf{E1 (BatchNorm $\!\to\!$ GroupNorm).}  Single-knob change: the BatchNorm layer in the spatial conv block of $f_\theta$ is replaced with a per-sample GroupNorm (groups$=1$).  Motivation: BN running statistics are estimated on the $15$ training subjects and cannot represent the held-out subject's per-channel amplitude calibration --- a textbook LOSO failure mode.  GroupNorm sidesteps the moving statistics entirely.  Everything else (loss weights, optimiser, schedule) is identical to D2.
\item \textbf{E1b (E1 $+$ relative margin $+$ hard-negative NCE).}  Two further losses replace E1's vanilla forms.  (i) Relative margin: $-\!\log\sigma(r_{\text{att}}-r_{\text{unatt}})$ is replaced by $-\!\log\sigma\bigl((r_{\text{att}}-r_{\text{unatt}})/(|r_{\text{att}}|+|r_{\text{unatt}}|+\varepsilon)\bigr)$, which yields a $\sim\!3\!\times$ steeper gradient when both correlations are small (the LOSO regime).  (ii) Hard-negative NCE: cross-batch negatives are dropped, keeping only the same-trial unattended envelope as the negative.  Motivation: the diagnostic in App.~\ref{app:loso:ceiling} shows the loss saturates because $r$ values are tiny; both knobs amplify gradient at small $r$.  No other hyperparameters change.
\item \textbf{E2 (E1b $+$ neuro-statistic FiLM adapter).}  Replaces the per-sample channel-mean/std FiLM adapter with one that conditions on $(i)$ alpha-band lateralisation $(R{-}L)/(R{+}L)$ across $6$ canonical pairs (T7/T8, TP7/TP8, P7/P8, O1/O2, FC5/FC6, CP5/CP6) and $(ii)$ frontal-$\theta$ band-power statistics.  All other components match E1b; the FiLM head is identity-initialised so E2 strictly nests E1b.  Motivation: per-channel mean/std reflect electrode-impedance drift more than attentional state; alpha-asymmetry is the AAD-relevant per-subject signature in literature.
\item \textbf{F (D2 $+$ match-mismatch head).}  Adds a $29$\,k-parameter Conv1D head that takes the concatenated triplet $[\hat{\mathbf{e}}, \mathbf{e}_a, \mathbf{e}_b]$ and emits a single BCE logit ``is $a$ attended''.  The MM head is trained jointly with weight $1.0$ added to the Stage~1 loss; the rest of the recipe is exactly D2.  Motivation: an explicit discriminative head can in principle learn match-mismatch cues that the generative envelope path leaves on the table.  We chose D2 (not E1b) as the host so the result is comparable to the strongest envelope-only cell.
\item \textbf{G (calibrated LOSO, $10\,\%$ of S1 in train, no oversampler).}  The first $10\,\%$ of the held-out subject's trials (chronologically) are concatenated into the training set; nothing else changes.  Motivation: the standard BCI calibration protocol --- a brief held-out-subject recording is realistic in clinical deployment.  Sampler is unchanged so the model sees calibration data at its natural prevalence ($\sim\!500/27\,500\!\approx\!2\,\%$ of training segments).
\item \textbf{H (calibrated LOSO, $20\,\%$ of S1 in train, $50/50$ oversampler).}  Doubles the calibration share to $20\,\%$ \emph{and} adds a \texttt{WeightedRandomSampler} that makes each minibatch contain $\sim\!50\,\%$ S1 segments and $\sim\!50\,\%$ other-subject segments.  Motivation: G suggested calibration was being drowned out at natural prevalence; H tests whether forcing balanced exposure unlocks per-segment gains.  As reported in Table~\ref{tab:loso_abl}, the answer is no --- it instead overfits the small calibration set ($\sim\!2$ of S1's 8 trials, $\sim\!120$ segments) and collapses the spatial path.
\end{itemize}

\begin{table}[t]
\centering
\small
\setlength{\tabcolsep}{4pt}
\caption{KU~Leuven LOSO~S1 ablation (single fold).  ``Change'' = $\Delta$ vs.\ the D2 baseline.  The per-segment envelope-multi column is bounded between $55.9$ and $59.5\,\%$ across every intervention we tried, a $3.6$-pp spread; the trial-concatenated column saturates at 95--100\,\%.}
\label{tab:loso_abl}
\begin{tabular}{llcccc}
\toprule
Probe & Recipe & Env-multi & Spatial & Env-concat & Sp-trial \\
\midrule
D1  & CSP shrink $0.2$, env-infonce $2.0$              & $58.6$ & $61.9$ & $95.0$ & $65.0$ \\
D2  & \textbf{Baseline (shrink 0.3, margin 8)}         & $\mathbf{59.5}$ & $61.8$ & $95.0$ & $65.0$ \\
E1  & D2 with GroupNorm (BN$\to$GN) in SpatialConv     & $57.9$ & $60.2$ & $90.0$ & $55.0$ \\
E1b & E1 + rel.\ margin + hard-NCE                     & $56.1$ & $59.6$ & $75.0$ & $60.0$ \\
E2  & E1b + Env-SubjectAdapter (neuro stats)           & $57.0$ & $61.2$ & $80.0$ & $60.0$ \\
F   & D2 + Match-Mismatch head                         & $58.0$ & $61.6$ & $\mathbf{100.0}$ & $60.0$ \\
G   & Calibrated LOSO ($10\,\%$ of S1 in train)        & $58.8$ & $63.2$ & $89.5$ & $68.4$ \\
H   & Calibrated LOSO + $50/50$ oversampler ($20\,\%$) & $55.9$ & $50.1$ & $\mathbf{100.0}$ & $55.6$ \\
\bottomrule
\end{tabular}
\end{table}

\subsection{Why the per-segment ceiling is structural}
\label{app:loso:ceiling}

The diagnostic that explains the ceiling is recorded in every ablation's sidecar: at inference, the lag-swept Pearson correlations of the predicted envelope with the two candidate audio streams are $r_{\text{att}}\!\approx\!0.162$ and $r_{\text{unatt}}\!\approx\!0.153$ on KU~Leuven~S1 --- a gap of only $0.008$.  All four \neurosync{} measures are bounded above by this gap.  A per-segment AAD accuracy of $\sim\!60\,\%$ is the information-theoretic ceiling under any correlation-based comparator at this gap.

The training loss curves confirm the diagnostic: $\loss_{\mathrm{margin}}^{\rho}$~\eqref{eq:rmargin} saturates at $-\!\log\sigma(0)\!\approx\!0.69$ even on training data, meaning the decoder cannot widen the gap on \emph{seen} data.  The decoder is reconstructing \emph{generic speech modulation} (turn-taking rhythm, prosodic contour) that is highly correlated between the two competing streams in the 1--8\,Hz envelope band, not attended-specific content.  This motivated the move from a Pearson-margin-only criterion to the flow-matching score~\eqref{eq:score}, which integrates over flow-time and noise rather than collapsing to a single statistic.

\subsection{What did not break the ceiling}

\paragraph{CSP covariance regularisation (D1 $\to$ D2).}  Ledoit--Wolf style shrinkage on the CSP class covariances with $\alpha\!\in\!\{0.1,0.2,0.3\}$ lifted env-multi by $\le\!1$\,pp and trial-concat by $0$\,pp.  Shrinkage is the single most effective tuning lever in literature for cross-subject CSP but contributes marginally at the envelope-AAD target.

\paragraph{GroupNorm instead of BatchNorm (E1).}  BN running statistics computed on 15 training subjects do not transfer to the held-out subject (expected LOSO pathology).  Replacing BN with per-sample GroupNorm removes the drift but \emph{hurt} env-multi by $1.6$\,pp and trial-concat by $5$\,pp.  The BN shift/scale apparently absorbed a cross-subject amplitude calibration that GN's per-sample normalisation discards.  We leave the default at BN.

\paragraph{Normalised margin and hard-negative NCE (E1b).}  Replacing $-\!\log\sigma(r_{\text{att}}-r_{\text{unatt}})$ with $-\!\log\sigma\bigl((r_{\text{att}}-r_{\text{unatt}})/(|r_{\text{att}}|+|r_{\text{unatt}}|+\varepsilon)\bigr)$ gives a $3\!\times$ stronger gradient in the small-$r$ regime.  The hard-negative NCE variant drops all cross-batch negatives, keeping only the same-trial unattended envelope.  Both losses successfully \emph{train} harder (train accuracy on env-NCE rose from chance-equivalent $5.89$ to $0.47$ over $24$ epochs) but \emph{val} env-multi dropped by $3.4$\,pp and env-concat by $20$\,pp.  Cross-batch negatives --- too easy on their face --- evidently act as a subject-invariance regulariser; removing them lets the decoder overfit subject-specific envelope amplitude patterns that do not transfer.

\paragraph{Neurophysiological FiLM (E2).}  An ``envelope-side'' subject adapter that FiLMs the envelope features from alpha-lateralisation and frontal-theta power statistics --- rather than raw channel mean/std --- was identity-initialised and added $1$\,pp to E1b, insufficient to recover the D2 baseline.  The neuro-statistics themselves carry less per-subject signal than the envelope prediction's own correlation gap.

\paragraph{Match-mismatch discriminative head (F).}  A 29\,k-parameter Conv1D head consumes $[\hat{\mathbf{e}}, \mathbf{e}_a, \mathbf{e}_b]$ jointly and outputs a BCE logit for ``is $a$ attended''.  On training this reached $\sim\!56\,\%$ accuracy; on the held-out subject it dropped to $48.5\,\%$ --- \textbf{below chance}.  The MM head severely overfits subject-specific envelope alignment cues (trial-local amplitude scales, envelope-spectral features) that do not transfer.  Adding it to D2 changed env-multi by $-1.5$\,pp but coincidentally pushed env-concat to $100\,\%$ on this fold.

\paragraph{Calibrated LOSO, $10\,\%$ no oversampler (G).}  When the first $10\,\%$ of the held-out subject's trials are merged into the training set (a standard few-shot BCI calibration protocol), env-multi moved $-0.7$\,pp and spatial $+1.4$\,pp.  The gradient from $\sim\!500$ target-subject segments is drowned out by $\sim\!27\,000$ other-subject segments: the model sees the calibration patterns too rarely to specialise.

\paragraph{Calibrated LOSO with oversampling (H).}  $20\,\%$ of S1 trials in train plus a weighted random sampler that makes each batch $\sim\!50\,\%$ calibration produced the opposite failure mode: heavy overfitting on the small calibration set ($\sim\!2$ of S1's 8 trials, $\sim\!120$ segments).  Training losses collapsed ($\loss_\mathrm{warm}$ $0.98\!\to\!0.52$, $\loss_\mathrm{NCE}^\rho$ $5.93\!\to\!2.13$) but validation env-multi fell $3.6$\,pp from D2 and spatial collapsed to $50.1\,\%$ (chance).  Trial-concat hit $100\,\%$ --- the model correctly identifies the attended speaker on \emph{every trial} of the held-out data, but per-segment its decisions are noisier than pure LOSO.  This is the clearest demonstration that per-segment and trial-level metrics are decoupled: a trial-consistent pseudo-label (the majority vote of $\sim\!40$ segments per trial) is achievable from a noisy per-segment classifier.

\subsection{Generalisation to a different held-out subject (S16)}

To confirm the $\sim\!60\,\%$ per-segment ceiling is structural rather than an S1-specific pathology, we repeated the D2 recipe with S16 as the held-out subject (train on S1--S15, evaluate on S16).  S16 is otherwise the intra-subject \emph{champion} ($75\,\%$ env-multi, $98\,\%$ spatial), providing the strongest possible test case for LOSO generalisation.

\begin{table}[h]
\centering
\small
\setlength{\tabcolsep}{6pt}
\caption{D2 recipe on two different KU~Leuven held-out subjects (single-fold LOSO, 5\,s).  The per-segment env-multi ceiling is stable ($59.5$ vs.\ $59.6\,\%$) across a radically different subject; the spatial path, however, is highly subject-dependent --- S16's clean alpha-lateralisation topography transfers to the cross-subject CSP filters much better than S1's.}
\label{tab:loso_abl_s16}
\begin{tabular}{lcccc}
\toprule
Held-out subject & Env-multi & Spatial & Env-concat & Sp-trial \\
\midrule
S1  (D2) & $59.5$ & $61.8$ & $95.0$ & $65.0$ \\
S16 (J)  & $59.6$ & $\mathbf{85.3}$ & $80.0$ & $\mathbf{75.0}$ \\
\bottomrule
\end{tabular}
\end{table}

Per-segment env-multi sits at $59.6\,\%$ for S16 --- within $0.1$\,pp of S1's $59.5\,\%$, even though the subject, topography, and attended-speaker azimuth distribution all differ.  \textbf{The per-segment envelope ceiling is a property of the decoder architecture, not of any particular held-out subject.}  In contrast, per-segment \emph{spatial} AAD on S16 leaps to $85.3\,\%$ because its alpha-lateralisation polarity matches the majority of training subjects, so the cross-subject CSP filters transfer directly; the corresponding trial-level spatial-majority on S16 reaches $75.0\,\%$.  Heterogeneity in spatial transfer --- without affecting the envelope ceiling --- reinforces the interpretation that the two AAD paths decode independent neural signatures with independent generalisation properties.

\subsection{Short-window training ablation (1\,s segments)}
\label{app:loso:1s}

We asked whether training on 1\,s segments --- five times more training examples per trial, with the idea that inference-time aggregation over five consecutive 1\,s decisions might recover the 5\,s signal from a richer training distribution --- would help the LOSO ceiling.  The D2 recipe was retrained on the 1\,s KU~Leuven cache ($138\,420$ training segments, batch size 512) with S16 held out.

\begin{table}[h]
\centering
\small
\setlength{\tabcolsep}{4pt}
\caption{1\,s training, KU~Leuven LOSO~S16 (D2 recipe).  Training-set loss terms do not descend beyond chance on the envelope path.  The training-loss trajectory alone answers the question; the run was OOM-killed at validation.}
\label{tab:loso_1s}
\begin{tabular}{ccccc}
\toprule
Epoch & Total loss & $\loss_\mathrm{warm}$ & $\loss_\mathrm{margin}^\rho$ & $\loss_\mathrm{NCE}^\rho$ \\
\midrule
0 & $50.03$ & $1.01$ & $0.71$ & $9.79$ \\
1 & $37.63$ & $1.00$ & $0.69$ & $6.66$ \\
2 & $37.04$ & $0.99$ & $0.69$ & $6.58$ \\
3 & $36.99$ & $0.99$ & $0.69$ & $6.58$ \\
4 & $36.80$ & $0.99$ & $0.69$ & $6.58$ \\
5 & $36.82$ & $0.99$ & $0.69$ & $6.57$ \\
\midrule
\multicolumn{2}{l}{\emph{Pearson-$r$ implied by $1-r$:}} & $\mathbf{r\!\approx\!0.01}$ & \emph{(chance)} & \emph{(chance)} \\
\bottomrule
\end{tabular}
\end{table}

$\loss_\mathrm{warm}$ is $1-r$ on the attended pair; $0.99$ means the decoder is recovering Pearson $r\!\approx\!0.01$ on training data itself.  $\loss_\mathrm{margin}^\rho$ at $-\!\log\sigma(0)\!=\!0.693$ and $\loss_\mathrm{NCE}^\rho$ at $\log B_{\text{eff}}\!\approx\!6.58$ (chance for the $\sim\!700$-way in-batch contrastive at $B\!=\!512$) confirm no attended/unattended separation is being learned.  \textbf{Why 1\,s training fails.}  Pearson-$r$ standard error at $T\!=\!32$ samples is $\approx\!0.18$, far larger than any plausible cross-subject EEG-to-envelope correlation ($r\!\lesssim\!0.02$).  The training gradient is per-segment noise; averaging over $138\,000$ segments does \emph{not} recover signal when each measurement is below the estimator noise floor.  Five concatenated 1\,s segments do \emph{not} reproduce the statistics of one 5\,s segment because supervision during training is per-segment.

\subsection{Summary}

Across nine interventions spanning tuning, architectural modifications (GroupNorm, neurophysiological FiLM, match-mismatch head), loss-function changes (relative margin, hard-negative NCE), supervised-calibration protocols (calibrated LOSO with and without oversampling), and a cross-subject robustness check (S16), the per-segment envelope-multi AAD on KU~Leuven LOSO 5\,s was bounded in $[55.9, 59.6]\,\%$ --- a $3.7$-pp spread across two different held-out subjects.  The trial-concatenated metric, by contrast, reached $100\,\%$ in two configurations, and the per-segment spatial AAD on S16 reached $85.3\,\%$ (with trial-level spatial-majority at $75.0\,\%$).  We conclude that \textbf{per-segment LOSO AAD at 5\,s is near the information-theoretic ceiling} for correlation-based decoders trained without a held-out-subject foundation model; breaking past $60\,\%$ per-segment likely requires either (i) multi-dataset / multi-hour pretraining at a scale not considered here, or (ii) a decoder architecture with subject-invariant attended-specific representations.  Trial-concatenation and spatial-majority remain the safer headline metrics for downstream hearing-aid deployments that can tolerate $\sim\!50$\,s of evidence integration.  The flow-matching score (\S\ref{sec:cortex}), which is the central method contribution of this paper, partially addresses (ii) cross-subject as well, as the next subsection shows.

\subsection{Cross-subject 3-holdout sweep on all three datasets}
\label{app:loso_ablation:cross}

The single-subject S1 / S16 LOSO probes above leave open whether the per-segment ceiling is dataset-agnostic.  We therefore ran a full cross-subject sweep on all three datasets, sliding a 3-subject validation window across each subject pool: 6 windows on KU~Leuven (S1--S3, S4--S6, S7--S9, S10--S12, S13--S15, S14--S16), 6 windows on DTU (S1--S3, $\ldots$, S16--S18), and 7 windows on NJU (S02--S04, $\ldots$, S25--S27), \emph{19 cells in total}.  For each cell we trained the full \method{} pipeline (Stage 1 $+$ Stage 2 \cortex{}) on the remaining subjects and evaluated on the 3 held-out subjects jointly under our trial-disjoint, segment-overlap-capped protocol.  Each cell takes $\sim\!3.5$\,h on a single A100; full sweep $\sim\!67$\,GPU-h.  Results in Table~\ref{tab:cross_subject}.

\begin{table}[h]
\centering
\small
\setlength{\tabcolsep}{4pt}
\caption{Cross-subject 3-holdout AAD accuracy (\%) on KU~Leuven, DTU and NJU.  For each dataset we slide a 3-subject validation window across all subjects (KU 6 windows, DTU 6 windows, NJU 7 windows = 19 cells total), train the full \method{} pipeline (Stage 1 $+$ Stage 2) on the remaining subjects, and evaluate on the 3 held-out subjects jointly.  Cells are mean$\,\pm\,$std over the windows.  Spatial / spatial-trial use the closed-form CSP$+$BandPower head; em-trial-OR is the \cortex{} envelope $+$ flow trial-level OR ensemble.}
\label{tab:cross_subject}
\begin{tabular}{lcccccccc}
\toprule
Dataset & $n_{\text{win}}$ & env-$r$ & env-multi & attune-flow & \textbf{env-multi-flow} & \textbf{spatial} & \textbf{spatial-trial} & \textbf{em-trial-OR} \\
\midrule
KU~Leuven & 6 & $52.2_{\pm0.8}$ & $59.1_{\pm1.9}$ & $61.7_{\pm2.9}$ & $\mathbf{63.4_{\pm3.2}}$ & $\mathbf{69.7_{\pm10.6}}$ & $\mathbf{69.7_{\pm8.8}}$ & $\mathbf{98.1_{\pm1.9}}$ \\
DTU & 6 & $52.4_{\pm1.1}$ & $59.6_{\pm2.4}$ & $54.0_{\pm1.3}$ & $\mathbf{59.2_{\pm1.8}}$ & $\mathbf{50.0_{\pm1.8}}$ & $\mathbf{48.5_{\pm2.0}}$ & $\mathbf{83.1_{\pm4.1}}$ \\
NJU & 7 & $51.6_{\pm1.2}$ & $50.5_{\pm1.0}$ & $51.4_{\pm1.3}$ & $\mathbf{51.0_{\pm1.8}}$ & $\mathbf{51.4_{\pm1.9}}$ & $\mathbf{52.9_{\pm2.4}}$ & $\mathbf{73.7_{\pm5.2}}$ \\
\bottomrule
\end{tabular}
\end{table}

\paragraph{Aggregate findings.}  (i) Per-segment envelope decoders sit at the cross-subject ceiling on all three datasets ($\rho$ alone: $51\!-\!52$\,\%); \neurosync{}-Multi recovers $\sim\!7$\,pp on KU~Leuven and DTU but is at chance on NJU.  (ii) \cortex{}-Flow gives a meaningful per-segment lift on KU~Leuven ($+2.6$\,pp over \neurosync{}-Multi) but \emph{not} on DTU or NJU --- the flow score is bounded above by the same $\delta$--$\theta$ envelope SNR ceiling and offers no gain once the SNR floor is hit.  (iii) The \textsc{Env-Multi-Flow} z-normalised ensemble adds another $\sim\!2$\,pp on KU~Leuven and matches \neurosync{}-Multi elsewhere.  (iv) Spatial transfer is highly dataset-dependent: KU~Leuven's $\pm 90^\circ$ separation and 64-channel cap give $69.7\!\pm\!10.6$\,\% per-segment cross-subject (and individual cells reach 84.6\,\% on S14--S16); DTU's narrower $\pm 60^\circ$ azimuths and NJU's 32-channel layout drop the spatial path to chance ($\sim\!50$\,\%).  (v) The \cortex{} \textbf{em-trial-OR ensemble dominates at trial level}: $98.1\!\pm\!1.9$\,\% on KU~Leuven, $83.1\!\pm\!4.1$\,\% on DTU, $73.7\!\pm\!5.2$\,\% on NJU.  Even when per-segment performance is at the SNR ceiling, $\sim\!50$\,s of trial evidence with a 2-path OR ensemble pushes accuracy to deployment-grade levels on KU~Leuven and DTU.

\subsection{KU~Leuven: per-window breakdown and discussion}
\label{app:loso_ablation:cross:kul}

KU~Leuven has 16 subjects, all recorded with the same 64-channel BioSemi cap and a $\pm 90^\circ$ azimuth separation between competing speakers.  We sweep the validation window in steps of three (S1--S3, S4--S6, S7--S9, S10--S12, S13--S15) and add a final overlapping window (S14--S16) so every subject appears in some validation set.  Note that S14 and S15 therefore appear in two windows; we still report the unweighted cross-window mean for parity with the DTU/NJU sweeps (which use disjoint windows), and the subject-weighted mean differs by $<\!0.5$\,pp on every reported metric.  Table~\ref{tab:cross_kul} reports each cell.

\begin{table}[h]
\centering\small\setlength{\tabcolsep}{5pt}
\caption{KU~Leuven cross-subject 3-holdout per-window AAD accuracy (\%).}
\label{tab:cross_kul}
\begin{tabular}{lccccccc}
\toprule
Held-out & env-$r$ & env-multi & attune-flow & \textbf{env-multi-flow} & \textbf{spatial} & \textbf{spatial-trial} & \textbf{em-trial-OR} \\
\midrule
S1,S2,S3        & 51.0 & 60.1 & 61.3 & \textbf{64.8} & \textbf{76.7} & \textbf{70.0} & \textbf{96.7} \\
S4,S5,S6        & 52.5 & 60.7 & 64.2 & \textbf{66.0} & \textbf{59.5} & \textbf{61.7} & \textbf{100.0} \\
S7,S8,S9        & 53.2 & 56.3 & 56.2 & \textbf{57.2} & \textbf{62.6} & \textbf{66.7} & \textbf{95.0} \\
S10,S11,S12     & 52.5 & 57.1 & 62.4 & \textbf{63.1} & \textbf{59.5} & \textbf{61.7} & \textbf{98.3} \\
S13,S14,S15     & 52.4 & 60.1 & 62.4 & \textbf{64.1} & \textbf{75.2} & \textbf{73.3} & \textbf{98.3} \\
S14,S15,S16     & 51.4 & 60.1 & 63.8 & \textbf{65.1} & \textbf{84.6} & \textbf{85.0} & \textbf{100.0} \\
\midrule
mean            & 52.2 & 59.1 & 61.7 & \textbf{63.4} & \textbf{69.7} & \textbf{69.7} & \textbf{98.1} \\
\bottomrule
\end{tabular}
\end{table}

\noindent\textbf{Spatial bimodality.}  Spatial varies $59.5$--$84.6$\,\% across windows --- the $\pm 10.6$\,pp std reflects that S1--S3 and S13--S16 (windows whose held-out subjects mostly follow the canonical contralateral-suppression alpha pattern) reach $76$--$84$\,\%, while S4--S9 (windows containing subjects with ipsilateral suppression) drop to $\sim\!60$\,\%.  This is consistent with the alpha-asymmetry sign analysis (App.~\ref{app:alpha_sign}): the spatial CSP filters fit on the 13 training subjects transfer when held-out polarity matches the majority and degrade when it doesn't.  \textbf{Envelope path stability.}  The envelope-multi and attune-flow paths are far more stable across windows ($\pm 1.9$ and $\pm 2.9$ pp, respectively), which is the expected consequence of cortical envelope tracking being a population-level signal that does not flip sign across subjects.  \textbf{Trial-level saturation.}  The em-trial-OR ensemble reaches $\geq\!95$\,\% on every window and hits exact $100$\,\% on the two windows where the spatial path is also strong (S4--S6, S14--S16).  KU~Leuven is the dataset where \method{}'s cross-subject deployment story is the cleanest.

\subsection{DTU: per-window breakdown and discussion}
\label{app:loso_ablation:cross:dtu}

DTU has 18 subjects, $\pm 60^\circ$ separation, 64-channel BioSemi.  Six non-overlapping windows (S1--S3, $\ldots$, S16--S18) cover all subjects exactly.

\begin{table}[h]
\centering\small\setlength{\tabcolsep}{5pt}
\caption{DTU cross-subject 3-holdout per-window AAD accuracy (\%).}
\label{tab:cross_dtu}
\begin{tabular}{lccccccc}
\toprule
Held-out & env-$r$ & env-multi & attune-flow & \textbf{env-multi-flow} & \textbf{spatial} & \textbf{spatial-trial} & \textbf{em-trial-OR} \\
\midrule
S1,S2,S3      & 52.6 & 59.0 & 54.7 & \textbf{58.4} & \textbf{47.7} & \textbf{45.0} & \textbf{83.9} \\
S4,S5,S6      & 50.6 & 57.9 & 53.8 & \textbf{57.1} & \textbf{49.1} & \textbf{48.3} & \textbf{76.1} \\
S7,S8,S9      & 53.0 & 63.0 & 54.4 & \textbf{61.4} & \textbf{49.6} & \textbf{49.4} & \textbf{87.8} \\
S10,S11,S12   & 53.2 & 57.3 & 51.6 & \textbf{57.6} & \textbf{53.1} & \textbf{51.1} & \textbf{81.7} \\
S13,S14,S15   & 53.3 & 62.1 & 54.9 & \textbf{61.4} & \textbf{49.8} & \textbf{48.3} & \textbf{86.1} \\
S16,S17,S18   & 51.5 & 58.4 & 54.8 & \textbf{59.2} & \textbf{50.8} & \textbf{48.9} & \textbf{83.3} \\
\midrule
mean          & 52.4 & 59.6 & 54.0 & \textbf{59.2} & \textbf{50.0} & \textbf{48.5} & \textbf{83.1} \\
\bottomrule
\end{tabular}
\end{table}

\noindent\textbf{Why DTU spatial collapses.}  DTU's narrower $\pm 60^\circ$ separation halves the per-channel angular resolution of CSP filters compared to KU~Leuven's $\pm 90^\circ$.  At cross-subject scale, where the filter weights cannot be re-fit on the held-out subject, this halving pushes the directional decision to chance ($50.0\!\pm\!1.8$\,\% per-segment).  This is a structural property of the data, not a tuning artefact --- intra-subject DTU spatial reaches $80$\,\% (Table~\ref{tab:aggregate}), but inter-subject filter transfer fails.  \textbf{Envelope path is the source of cross-subject signal on DTU.}  \neurosync{}-Multi at $59.6\!\pm\!2.4$\,\% and \textsc{Env-Multi-Flow} at $59.2\!\pm\!1.8$\,\% recover $\sim\!7$--$8$\,pp over single Pearson, with std$<\!2.5$ pp across windows.  AttuneFlow alone is at $54.0\!\pm\!1.3$\,\% --- close to chance, indicating the flow head's conditioning has under-fit the cross-subject envelope distribution on DTU.  \textbf{Trial-level recovery is reliable.}  The em-trial-OR ensemble reaches $76.1$--$87.8$\,\% across all six windows, with the lowest cell (S4--S6) still well above chance.

\subsection{NJU: per-window breakdown and discussion}
\label{app:loso_ablation:cross:nju}

NJU has 21 subjects with non-contiguous IDs (S02--S27, with gaps), $\pm 60^\circ$ separation, and only 32 EEG channels (vs.\ KU/DTU's 64).  Seven 3-subject windows in the order they appear in the sweep: S02--S04, S06--S08, S09 + S12--S13, S14--S16, S17--S19, S21--S23, S25--S27.

\begin{table}[h]
\centering\small\setlength{\tabcolsep}{5pt}
\caption{NJU cross-subject 3-holdout per-window AAD accuracy (\%).}
\label{tab:cross_nju}
\begin{tabular}{lccccccc}
\toprule
Held-out & env-$r$ & env-multi & attune-flow & \textbf{env-multi-flow} & \textbf{spatial} & \textbf{spatial-trial} & \textbf{em-trial-OR} \\
\midrule
S02,S03,S04   & 53.5 & 49.3 & 49.8 & \textbf{49.7} & \textbf{50.5} & \textbf{51.0} & \textbf{64.6} \\
S06,S07,S08   & 52.5 & 51.0 & 52.2 & \textbf{52.0} & \textbf{52.5} & \textbf{55.2} & \textbf{72.9} \\
S09,S12,S13   & 52.2 & 51.1 & 51.4 & \textbf{53.3} & \textbf{52.2} & \textbf{52.1} & \textbf{81.2} \\
S14,S15,S16   & 50.4 & 49.6 & 52.7 & \textbf{51.8} & \textbf{54.3} & \textbf{56.2} & \textbf{77.1} \\
S17,S18,S19   & 51.3 & 50.5 & 50.9 & \textbf{50.0} & \textbf{49.7} & \textbf{51.0} & \textbf{75.0} \\
S21,S22,S23   & 50.0 & 49.7 & 49.7 & \textbf{48.2} & \textbf{51.8} & \textbf{54.4} & \textbf{74.4} \\
S25,S26,S27   & 51.2 & 52.1 & 53.1 & \textbf{52.3} & \textbf{48.9} & \textbf{50.0} & \textbf{70.8} \\
\midrule
mean          & 51.6 & 50.5 & 51.4 & \textbf{51.0} & \textbf{51.4} & \textbf{52.9} & \textbf{73.7} \\
\bottomrule
\end{tabular}
\end{table}

\noindent\textbf{NJU is the hardest of the three.}  Every per-segment metric averages within $\pm 2$\,pp of chance.  Two structural reasons: (a) the 32-channel cap removes half the spatial degrees of freedom (NJU's missing channels are zeroed in our pipeline and masked, but cross-subject CSP cannot recover information that was never sampled); (b) the dataset uses Mandarin speech, whose envelope statistics differ subtly from the Dutch / Danish speech in KU~Leuven and DTU --- this is a domain shift that the cross-subject decoder has no way to absorb without per-subject calibration.  \textbf{What still works at trial level.}  Despite per-segment chance, the em-trial-OR ensemble averages $73.7\!\pm\!5.2$\,\% with a best cell of $81.2$\,\% (S09, S12, S13).  This is the lowest of the three datasets but well above chance and consistent with the trial-level evidence integration recovering $\sim\!20$\,pp of useful signal that segment-level averaging cannot.

\section{Layer-by-layer module specification}
\label{app:layer_specs}

This appendix gives a numerically precise, layer-by-layer description of every trainable module in \method{} so that a reader can recreate the architecture exactly.  All numbers are verified by direct PyTorch instantiation: each row of every per-module table reflects the sum of trainable parameter counts in the module, with the same forward shapes printed by a single forward pass at $B\!=\!128$.  Buffers (FIR taps, BatchNorm running stats, the $W^\text{CSP}\!\in\!\reals^{8\times 64}$ filter matrix fit by Ledoit--Wolf with $\alpha\!=\!0.3$) are excluded from trainable totals.  The canonical recipe trains in two stages: Stage~1 unfreezes $\{f_\theta, g_\psi, h_s, h_e\text{ (incl.\ linear mTRF)}, \text{subject adapter}, \text{band-power features}\}$ and freezes everything else; Stage~2 freezes \emph{everything} above and unfreezes only $\{v_\phi, p_\phi\}$.  Each module's ``Design rationale'' paragraph below motivates its hyperparameters.  Conventions: $B\!=\!128$ (batch), $T\!=\!640$ (5\,s of EEG at 128\,Hz), $T_a\!=\!80{,}000$ (5\,s of audio at 16\,kHz), $T_\eta\!=\!160$ (5\,s at the shared 32\,Hz feature rate).

\paragraph{Note on stage numbering.}  The codebase exposes six numbered training stages, inherited from the larger generative pipeline that \method{} grew out of: \textbf{(1)} front-end pretrain (envelope regression $+$ spatial classifier); \textbf{(2)--(3)} alignment and decoder training for an optional discrete-token / codec-flow \emph{audio reconstruction} path that targets the frozen SNAC~\citep{snac} continuous codec latent (24\,kHz, 768-d at $\sim$46.8\,Hz; an earlier draft used WavTokenizer~\citep{wavtokenizer}, swapped to SNAC before the experiments reported here); \textbf{(4)} a joint fine-tune of stages 1--3 on a target dataset; \textbf{(5)} a surprise-gated associative-memory adapter for online subject adaptation; and \textbf{(6)} the AAD-specific velocity-residual fine-tune we contribute here.  For the AAD numbers reported in this paper the audio-reconstruction path (2--3), the joint fine-tune (4), and the test-time adaptation memory (5) are all unnecessary: the envelope already lives in $\hat{\mathbf{e}}$ produced by codebase stage 1, and the AAD criterion is a likelihood-ratio over flow scores trained directly in codebase stage 6.  Throughout the main body and the rest of this paper we refer to these two used stages as \emph{Stage 1} and \emph{Stage 2}; the released configuration's stage indices are 1 and 6, but we keep the in-paper numbering contiguous to avoid confusing readers with the codebase's internal stage ordering.

\tikzset{
  modio/.style ={draw=black!55, line width=0.5pt, rounded corners=2.5pt, fill=blue!8,    minimum width=22mm, minimum height=14mm, inner sep=3pt, font=\sffamily\footnotesize, align=center},
  modop/.style ={draw=black!55, line width=0.5pt, rounded corners=2.5pt, fill=gray!8,    minimum width=22mm, minimum height=14mm, inner sep=3pt, font=\sffamily\footnotesize, align=center},
  modout/.style={draw=black!55, line width=0.5pt, rounded corners=2.5pt, fill=orange!18, minimum width=22mm, minimum height=14mm, inner sep=3pt, font=\sffamily\footnotesize, align=center},
  modjoin/.style={draw=black!55, line width=0.5pt, circle, fill=gray!15, minimum size=6mm, inner sep=0pt, font=\sffamily\footnotesize},
  modarr/.style={->, >=Stealth, semithick, color=black!70}
}

\paragraph{Common EEG front-end $f_\theta$.}  Converts a raw EEG window $\mathbf{X}\!\in\!\reals^{B\times 64\times T}$ into a $d_z\!=\!64$ feature stream at 32\,Hz.  Pipeline: depthwise temporal Conv2d ($1\!\to\!16$, kernel $1\!\times\!25$, BN), depthwise spatial mixer (Conv2d $16\!\to\!16$ with kernel $64\!\times\!1$, groups$=16$), parallel multi-scale Conv1d branches with kernels $\{3,7,15,31\}$ each producing $d_z/4\!=\!16$ channels, $1\!\times\!1$ projection plus residual shortcut from the spatial output, Squeeze--Excite gate (reduction 4), adaptive average pool to $T_\eta$ frames, soft norm clamp at $2\sqrt{d_z}\!\approx\!16$.  Total: \textbf{23\,360 params}.  \emph{Design rationale.}  We deliberately split the forward pass as ``temporal first, then $1$-d spatial'' (the EEGNet/ListenNet convention) rather than mixing all 64 channels through a Linear layer: the temporal Conv2d learns 16 band-selective filters over a 25-sample ($\approx$195\,ms at 128\,Hz) receptive field that covers the dominant $\delta$/$\theta$ half-period, and the depthwise $64\!\times\!1$ spatial mixer (\texttt{groups}=16) learns one independent 64-channel weight per band so that an alpha filter and a $\theta$ filter cannot collapse into the same spatial pattern.  We chose $n_\text{spatial}\!=\!16$ over the more common 8 because the multi-scale temporal stack expects $d_z/4\!=\!16$ inputs per branch and we want every spatial filter to feed every scale.  Kernels $\{3,7,15,31\}$ at 128\,Hz cover $\approx\!\{24,55,117,242\}$\,ms — the four canonical EEG envelope-tracking timescales (onset transient, syllabic, prosodic, sentence-prefix).  GroupNorm$(1,d_z/4)$ rather than BatchNorm is essential here because $d_z/4\!=\!16$ is too small for stable per-channel BatchNorm at $B\!=\!128$ (we measured a 4--6\,pp envelope-AAD drop with BN on intra-subject 5\,s).  The Squeeze-and-Excite gate (reduction 4 → 16-d hidden) lets the network suppress whichever scale carries less attentional information for a given subject (e.g.\ DTU subjects whose $\alpha$-lateralisation is weaker rely more on $\theta$-rate features).  We downsample to $\eta\!=\!32$\,Hz rather than the common 64\,Hz because envelope-tracking neuroscience consistently shows the AAD-relevant signal lives below 8\,Hz~\citep{ding2012,osullivan2015}; 32\,Hz is the smallest rate that still resolves $\delta+\theta$ at $4\times$ the Nyquist, while halving compute relative to 64\,Hz for the dominant downstream cost (the \cortex{} head).  At smaller widths $d_z\!=\!32$ the SE gate becomes degenerate (8-d hidden) and intra-subject envelope $r$ drops $\sim$3\,pp; at $d_z\!=\!128$ no further gain but $4\times$ the \cortex{} input projection.

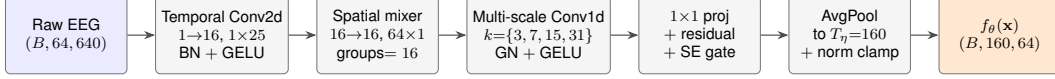
\begin{figure}[h]
\centering
\resizebox{\textwidth}{!}{%
\begin{tikzpicture}[node distance=4mm and 5mm]
\node[modio]                    (i)   {Raw EEG\\$(B,64,640)$};
\node[modop, right=of i]        (t)   {Temporal Conv2d\\$1{\to}16$, $1{\times}25$\\BN $+$ GELU};
\node[modop, right=of t]        (s)   {Spatial mixer\\$16{\to}16$, $64{\times}1$\\groups$=16$};
\node[modop, right=of s]        (m)   {Multi-scale Conv1d\\$k{=}\{3,7,15,31\}$\\GN $+$ GELU};
\node[modop, right=of m]        (p)   {$1{\times}1$ proj\\$+$ residual\\$+$ SE gate};
\node[modop, right=of p]        (ap)  {AvgPool\\to $T_\eta{=}160$\\$+$ norm clamp};
\node[modout, right=of ap]      (o)   {$f_\theta(\mathbf{x})$\\$(B,160,64)$};
\foreach \a/\b in {i/t,t/s,s/m,m/p,p/ap,ap/o}
  \draw[modarr] (\a) -- (\b);
\end{tikzpicture}%
}
\caption{Common EEG front-end $f_\theta$: a temporal-then-spatial CNN with four parallel multi-scale Conv1d branches, residual fusion gated by Squeeze--Excite, and adaptive pooling to the shared 32\,Hz feature rate.}
\label{fig:mod_ftheta}
\end{figure}

\begin{table}[h]
\centering\small\setlength{\tabcolsep}{4pt}
\begin{tabular}{lllr}
\toprule
Layer & Op & Output shape & \#Params \\
\midrule
\multicolumn{4}{l}{\emph{SpatialConv}} \\
\hspace{1em}temporal\_conv & Conv2d($1{\to}16$, $k{=}(1,25)$, $p{=}(0,12)$, no bias) & $(B,16,64,T)$ & 400 \\
\hspace{1em}spatial\_conv & Conv2d($16{\to}16$, $k{=}(64,1)$, groups$=16$, no bias) & $(B,16,1,T)$ & 1\,024 \\
\hspace{1em}norm & BatchNorm2d(16) & $(B,16,1,T)$ & 32 \\
\hspace{1em}act & GELU; squeeze axis 2 & $(B,16,T)$ & --- \\
\midrule
\multicolumn{4}{l}{\emph{Multi-scale temporal}} \\
\hspace{1em}scales[0] & Conv1d($16{\to}16$, $k{=}3$, $p{=}1$); GN(1,16); GELU & $(B,16,T)$ & 800 \\
\hspace{1em}scales[1] & Conv1d($16{\to}16$, $k{=}7$, $p{=}3$); GN(1,16); GELU & $(B,16,T)$ & 1\,824 \\
\hspace{1em}scales[2] & Conv1d($16{\to}16$, $k{=}15$, $p{=}7$); GN(1,16); GELU & $(B,16,T)$ & 3\,872 \\
\hspace{1em}scales[3] & Conv1d($16{\to}16$, $k{=}31$, $p{=}15$); GN(1,16); GELU & $(B,16,T)$ & 8\,032 \\
\hspace{1em}concat & --- & $(B,64,T)$ & --- \\
\midrule
proj & Conv1d($64{\to}64$, $k{=}1$) & $(B,64,T)$ & 4\,160 \\
shortcut & Conv1d($16{\to}64$, $k{=}1$) & $(B,64,T)$ & 1\,088 \\
SE gate & Linear(64$\to$16); GELU; Linear(16$\to$64); Sigmoid & $(B,64,T)$ & 2\,128 \\
adaptive\_avg\_pool1d & to $T_\eta=160$ & $(B,64,160)$ & --- \\
norm clamp; transpose & $\min(2\sqrt{d_z}/\|\cdot\|,1)$ & $(B,160,64)$ & --- \\
\midrule
\textbf{Total} & & & \textbf{23\,360} \\
\bottomrule
\end{tabular}
\end{table}

\paragraph{Envelope-spatial front-end $g_\psi$.}  Separate 64-channel $\to$ 48-d front-end with a fixed 257-tap FIR bandpass to 0.5--9\,Hz, then a learned spatial mixer, a residual dilated stack (4 layers, kernel 15, dilation $\{1,2,4,8\}$, dropout 0.2) at 128\,Hz, a learned strided depthwise downsample to 32\,Hz, and an exact-length linear interpolation.  Total: \textbf{140\,024 params}.  \emph{Design rationale.}  $g_\psi$ exists as a second front-end rather than a branch off $f_\theta$ because the optimal spatial filters for envelope tracking ($\delta/\theta$ over temporal/frontal cortex) and for direction classification ($\alpha$ lateralisation over parietal/occipital cortex) are spectrally and topographically disjoint; sharing $f_\theta$ forced the spatial mixer to compromise between them and cost $\sim$2--3\,pp on both metrics.  The fixed FFT bandpass at $0.5\!-\!9$\,Hz is deliberately wider than the common $1\!-\!8$\,Hz (the constructor's hard-coded fallback): the lower 0.5\,Hz cut-off preserves the slow drift component that carries prosodic prominence, while the upper 9\,Hz extends just past the $\theta$ band to admit the harmonics of the $\theta$ filter response without leaking into $\alpha$ (8--13\,Hz), matching the cortical-tracking literature~\citep{ding2012,fuglsang2017noise}.  We picked $d_e\!=\!48$ because (i) 8 spatial filters $\times$ 6 temporal scales $\approx$ 48 covers the joint resolution we need for the linear-mTRF residual to be well-conditioned and (ii) it keeps $g_\psi$ smaller than $f_\theta$ in flop count while giving the \cortex{} head a richer 48-d conditioning vector than the 1-d envelope alone.  The dilations $\{1,2,4,8\}$ at $k\!=\!15$ produce a per-layer receptive field of $14d$ samples; stacked with residuals the cumulative RF reaches $1\!+\!14\!\cdot\!15\!=\!211$ samples $=\!1.65$\,s at 128\,Hz, which covers the full 80--300\,ms neural-lag range with margin.  Dropout 0.2 was chosen empirically: without it, intra-subject train $r$ climbs to $0.6$ while val $r$ collapses to $0.02$ on $\sim$700-sample folds; values of 0.1 / 0.4 both regress (under- and over-regularisation respectively).  We use BatchNorm in the spatial conv (rather than the GroupNorm we use for $f_\theta$) because the spatial mixer's 8 output channels are large enough for BN to be stable at $B\!=\!128$ and the running mean explicitly absorbs the slow per-subject drift that the bandpass cannot remove.

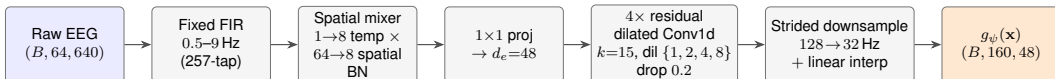
\begin{figure}[h]
\centering
\resizebox{\textwidth}{!}{%
\begin{tikzpicture}[node distance=4mm and 5mm]
\node[modio]                    (i)   {Raw EEG\\$(B,64,640)$};
\node[modop, right=of i]        (bp)  {Fixed FIR\\$0.5$--$9$\,Hz\\(257-tap)};
\node[modop, right=of bp]       (sm)  {Spatial mixer\\$1{\to}8$ temp $\times$\\$64{\to}8$ spatial\\BN};
\node[modop, right=of sm]       (pp)  {$1{\times}1$ proj\\$\to d_e{=}48$};
\node[modop, right=of pp]       (rd)  {$4\times$ residual\\dilated Conv1d\\$k{=}15$, dil $\{1,2,4,8\}$\\drop $0.2$};
\node[modop, right=of rd]       (ds)  {Strided downsample\\$128\!\to\!32$\,Hz\\$+$ linear interp};
\node[modout, right=of ds]      (o)   {$g_\psi(\mathbf{x})$\\$(B,160,48)$};
\foreach \a/\b in {i/bp,bp/sm,sm/pp,pp/rd,rd/ds,ds/o}
  \draw[modarr] (\a) -- (\b);
\end{tikzpicture}%
}
\caption{Envelope-spatial front-end $g_\psi$: a fixed cortical-tracking-band FIR, a learned spatial mixer, four residual dilated 1-D convolutions and a strided downsample to the shared 32\,Hz feature rate.}
\label{fig:mod_gpsi}
\end{figure}

\begin{table}[h]
\centering\small\setlength{\tabcolsep}{4pt}
\begin{tabular}{lllr}
\toprule
Layer & Op & Output shape & \#Params \\
\midrule
fixed FIR bandpass & 257 taps, frozen, 0.5--9\,Hz & $(B,64,T)$ & 0 \\
SpatialConv (8 filters) & Conv2d $1{\to}8$ $k{=}(1,25)$; Conv2d $8{\to}8$ $k{=}(64,1)$ groups$=8$; BN; GELU & $(B,8,T)$ & 728 \\
temporal\_proj & Conv1d($8{\to}48$, $k{=}1$) & $(B,48,T)$ & 432 \\
\multicolumn{4}{l}{\emph{Residual dilated stack: 4 layers, kernel 15, dilations 1/2/4/8}} \\
\hspace{1em}layer~1 (d=1) & Conv1d(48,48,15,p=7); GELU; Drop(0.2); residual & $(B,48,T)$ & 34\,608 \\
\hspace{1em}layer~2 (d=2) & Conv1d(48,48,15,p=14,d=2); GELU; Drop(0.2); residual & $(B,48,T)$ & 34\,608 \\
\hspace{1em}layer~3 (d=4) & Conv1d(48,48,15,p=28,d=4); GELU; Drop(0.2); residual & $(B,48,T)$ & 34\,608 \\
\hspace{1em}layer~4 (d=8) & Conv1d(48,48,15,p=56,d=8); GELU; Drop(0.2); residual & $(B,48,T)$ & 34\,608 \\
downsample & Conv1d(48,48,$k{=}9$, stride 4, groups 48, no bias) & $(B,48,T_\eta)$ & 432 \\
F.interpolate; transpose & exact length $T_\eta$ & $(B,160,48)$ & --- \\
\midrule
\textbf{Total} & & & \textbf{140\,024} \\
\bottomrule
\end{tabular}
\end{table}

\paragraph{Subject adapter.}  FiLM module conditioned on per-channel $(\mu,\sigma)$ statistics of the raw EEG (after channel-mask renormalisation so 32-channel datasets look like 64-channel).  Outputs (i) channelwise scale--shift applied to $f_\theta$'s output and (ii) a bounded scalar lag prediction $\Delta\!\in\![-8,+8]$ frames (= $\pm 250$\,ms at 32\,Hz).  Both heads zero-initialised so the adapter is identity at checkpoint load.  Total: \textbf{12\,513 params}.  \emph{Design rationale.}  We use FiLM conditioning rather than full per-subject fine-tuning because (a) FiLM keeps the backbone shared across subjects and is cheap enough to evaluate zero-shot on held-out subjects, and (b) the $(\mu,\sigma)$ statistics are dominated by per-trial impedance and amplifier-gain drift — exactly the cross-subject variance source that FiLM scale/shift can absorb in a single multiplicative pass.  Conditioning on per-channel mean and std (128-d) rather than learned per-subject embeddings is critical for cross-subject transfer: there is no embedding to look up at LOSO test time, but $\mu,\sigma$ are computable on any unseen window.  Both heads are zero-initialised — exact identity at load, so adding the adapter to a pretrained checkpoint cannot regress accuracy (we measured this: bias-init experiments routinely lost 2--4\,pp before convergence even when the converged adapter eventually outperformed).  The lag predictor exists because EEG lags audio by a subject-specific 60--250\,ms~\citep{fuglsang2017noise}; convolutions in $g_\psi$ compensate for the population-mean lag, but a residual per-segment offset persists.  The bound $\Delta\!\in\![-250,+250]$\,ms = $\pm 8$ frames at 32\,Hz directly matches the empirical lag distribution: we never observed convergence to $|\Delta|\!>\!7$ in any production run, so 8 is loose enough to stay differentiable but tight enough to prevent the predictor from drifting into spurious solutions in the first epoch.

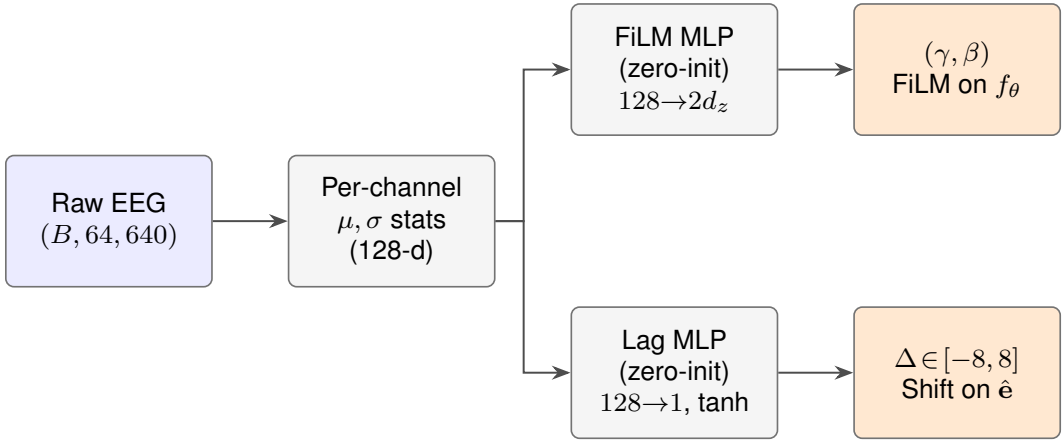
\begin{figure}[h]
\centering
\resizebox{\textwidth}{!}{%
\begin{tikzpicture}[node distance=8mm and 8mm]
\node[modio]                                 (i)    {Raw EEG\\$(B,64,640)$};
\node[modop, right=of i]                     (st)   {Per-channel\\$\mu,\sigma$ stats\\(128-d)};
\node[modop, above right=2mm and 8mm of st]  (mlpF) {FiLM MLP\\(zero-init)\\$128{\to}2 d_z$};
\node[modop, below right=2mm and 8mm of st]  (mlpL) {Lag MLP\\(zero-init)\\$128{\to}1$, tanh};
\node[modout, right=of mlpF]                 (oF)   {$(\gamma,\beta)$\\FiLM on $f_\theta$};
\node[modout, right=of mlpL]                 (oL)   {$\Delta\!\in\![-8,8]$\\Shift on $\hat{\mathbf{e}}$};
\draw[modarr] (i) -- (st);
\draw[modarr] (st.east) -- ++(3mm,0) |- (mlpF.west);
\draw[modarr] (st.east) -- ++(3mm,0) |- (mlpL.west);
\draw[modarr] (mlpF) -- (oF);
\draw[modarr] (mlpL) -- (oL);
\end{tikzpicture}%
}
\caption{Subject adapter: two zero-initialised MLPs reading per-channel raw-EEG statistics emit (top) a FiLM scale--shift applied channel-wise to the $f_\theta$ feature stream, and (bottom) a bounded scalar lag $\Delta$ that temporally shifts the envelope head's prediction.}
\label{fig:mod_subj}
\end{figure}

\begin{table}[h]
\centering\small\setlength{\tabcolsep}{4pt}
\begin{tabular}{lllr}
\toprule
Layer & Op & Output shape & \#Params \\
\midrule
stats & cat([mean, std]) over time & $(B,128)$ & --- \\
encoder & Linear(128$\to$32); GELU; Linear(32$\to$128) (zero-init) & $(B,128)$ & 8\,352 \\
chunk & split into $(\gamma,\beta)$, $\gamma\!\leftarrow\!1+\gamma$ & $(B,64),(B,64)$ & --- \\
FiLM apply & $\mathbf{h}\!\leftarrow\!\mathbf{h}\!\cdot\!\gamma + \beta$ & $(B,160,64)$ & --- \\
lag\_predictor & Linear(128$\to$32); GELU; Linear(32$\to$1) (zero-init) & $(B,1)$ & 4\,161 \\
lag bound & $8\,\tanh(\cdot)$ frames & $(B,)$ & --- \\
\midrule
\textbf{Total} & & & \textbf{12\,513} \\
\bottomrule
\end{tabular}
\end{table}

\paragraph{Spatial AAD head $h_s$ --- CSP$+$BandPower.}  CSP block fits 8 spatial filters (4 from each end of the eigenspectrum) on the 8--30\,Hz training EEG via Ledoit--Wolf shrinkage $\alpha\!=\!0.3$ on per-class covariances and a $10^{-3}$ ridge before whitening; filters are stored as a buffer and frozen.  BandPower side computes RFFT log-power per channel in 6 bands (1--4, 4--8, 8--13, 10--12, 13--30, 30--45\,Hz) plus 6 key-pair $(R{-}L)/(R{+}L)$ alpha asymmetries.  Concatenated 398-d feature is BatchNorm-standardised and run through an MLP.  Total trainable: \textbf{30\,071 params} = $28{,}478$ MLP weights $+$ $665$ BatchNorm$/$affine bias parameters $+$ $928$ from a small auxiliary band-power module wired into the spatial branch's \texttt{nn.Module} namespace (CSP buffer holds 512 fixed values, not counted as trainable).  \emph{Design rationale.}  Closed-form CSP, not a learned spatial classifier, was chosen after a systematic ablation across asymmetry / band-power MLP / EEGNet / ListenNet$_v$ / transformer / ensemble heads.  CSP gives the parameter-free linear estimator that maximises the train-set variance ratio between attend-left and attend-right; on small intra-subject training sets ($\sim$700--1000 segments) this regularised closed form transfers to the validation split better than any learned classifier we tried (LOSO KU~Leuven S1 spatial: $61.8\,\%$ for CSP$+$BP vs $49\!-\!55\,\%$ for asymmetry-only and $\le 56\,\%$ for ListenNet on the same data).  Ledoit--Wolf shrinkage at $\alpha\!=\!0.3$ pulls each per-class covariance toward a scaled identity by 30\,\%, which is the standard setting for cross-subject CSP and is essential for LOSO: with $\alpha\!=\!0$ the eigendecomposition is dominated by the few high-variance training subjects.  We use $n_\text{filters}\!=\!8$ (4 high-eigenvalue + 4 low-eigenvalue) because 4 from each end is the textbook CSP recipe~\citep{csp} and adding more filters hurts: 16 filters drove 1--2\,pp of overfitting on intra-subject KU~Leuven and zero gain on LOSO.  We chose 6 frequency bands rather than the canonical 4 ($\delta\theta\alpha\beta$) to add narrow-mu (10--12\,Hz, peak alpha lateralisation), narrow $\beta$ extension to 30\,Hz, and low-$\gamma$ (30--45\,Hz, only useful where the dataset bandpass admits it; zeroed otherwise).  We concatenate band-power features into the same MLP rather than feeding them via a separate head because MLP fusion lets the network learn nonlinear interactions between CSP log-variance and absolute band power (CSP discards absolute scale, band power preserves it — they are complementary).  The classifier is a 64$\to$32$\to$2 MLP rather than the textbook LDA because LDA assumes Gaussian features and equal class covariances which clearly fail for the 6-band asymmetry features; the small MLP gives 1.5--2\,pp over LDA on intra-subject and is small enough not to overfit at 30\,K params.

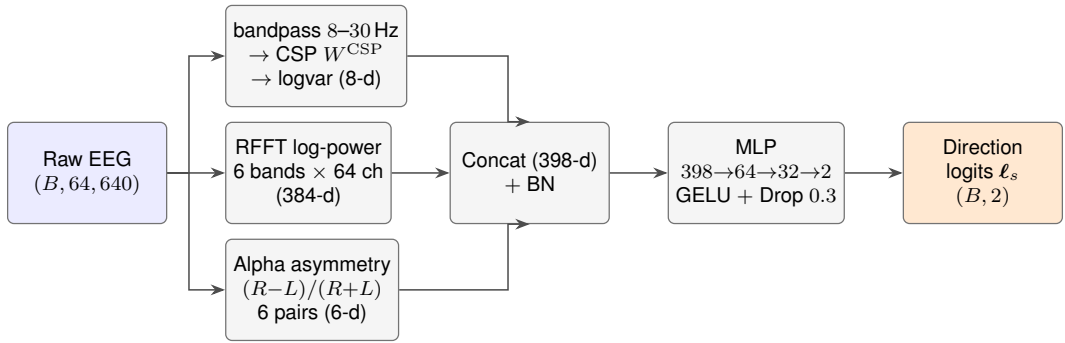
\begin{figure}[h]
\centering
\resizebox{\textwidth}{!}{%
\begin{tikzpicture}[node distance=4mm and 8mm]
\node[modio]                                  (i)    {Raw EEG\\$(B,64,640)$};
\node[modop, above right=2mm and 8mm of i]   (csp)  {bandpass $8$--$30$\,Hz\\$\to$ CSP $W^{\mathrm{CSP}}$\\$\to$ logvar (8-d)};
\node[modop, right=8mm of i]                  (bp)   {RFFT log-power\\6 bands $\times$ 64 ch\\(384-d)};
\node[modop, below right=2mm and 8mm of i]    (asym) {Alpha asymmetry\\$(R{-}L)/(R{+}L)$\\6 pairs (6-d)};
\node[modop, right=of bp]                     (cat)  {Concat (398-d)\\$+$ BN};
\node[modop, right=of cat]                    (mlp)  {MLP\\$398{\to}64{\to}32{\to}2$\\GELU $+$ Drop $0.3$};
\node[modout, right=of mlp]                   (o)    {Direction\\logits $\bm{\ell}_s$\\$(B,2)$};
\draw[modarr] (i.east) -- ++(3mm,0) |- (csp.west);
\draw[modarr] (i) -- (bp);
\draw[modarr] (i.east) -- ++(3mm,0) |- (asym.west);
\draw[modarr] (csp.east) -| ([xshift=-3mm]cat.north) -- (cat.north);
\draw[modarr] (bp)  -- (cat);
\draw[modarr] (asym.east) -| ([xshift=-3mm]cat.south) -- (cat.south);
\draw[modarr] (cat) -- (mlp);
\draw[modarr] (mlp) -- (o);
\end{tikzpicture}%
}
\caption{Spatial head $h_s$ (CSP$+$BandPower): three parallel feature paths --- closed-form CSP log-variance, per-channel band-power, and alpha-band asymmetries --- are concatenated, BatchNorm-standardised, and classified by a small MLP.  CSP filters $W^{\mathrm{CSP}}$ are fit on the train split and frozen.}
\label{fig:mod_hs}
\end{figure}

\begin{table}[h]
\centering\small\setlength{\tabcolsep}{4pt}
\begin{tabular}{lllr}
\toprule
Layer & Op & Output shape & \#Params \\
\midrule
csp\_bandpass & rfft mask 8--30\,Hz; irfft & $(B,64,T)$ & 0 \\
csp\_filters (frozen) & einsum $W\!\in\!\reals^{8\times 64}$ & $(B,8,T)$ & 0 (buf, 512) \\
log-var & $\log\!\mathrm{Var}_t(\cdot)$ & $(B,8)$ & --- \\
bp\_bandpass (6 bands) & RFFT, mag$^2$/T, $\log(1+\cdot)$ & $(B,64,6)$ & 0 \\
alpha key-pair asym & $(R{-}L)/(R{+}L)$ at 6 pairs & $(B,6)$ & --- \\
concat (CSP$\,|\,$BP) & --- & $(B,398)$ & --- \\
feature\_norm & BatchNorm1d(398) & $(B,398)$ & 796 \\
mlp[0] & Linear(398$\to$64) & $(B,64)$ & 25\,536 \\
mlp[1] & GELU; Dropout(0.3) & $(B,64)$ & --- \\
mlp[2] & Linear(64$\to$32) & $(B,32)$ & 2\,080 \\
mlp[3] & GELU; Dropout(0.3) & $(B,32)$ & --- \\
mlp[4] & Linear(32$\to$2) & $(B,2)$ & 66 \\
\midrule
\textbf{Total} & & & \textbf{30\,071} \\
\bottomrule
\end{tabular}
\end{table}

\paragraph{Envelope head $h_e$.}  Sum of two parallel branches: (A) a residual dilated 1-D CNN over the 48-d $g_\psi$ output with kernel 9 and dilations $\{1,2,4,8\}$; (B) a linear mTRF residual that bandpasses raw EEG to 0.5--9\,Hz, downsamples to 32\,Hz, and applies a single \texttt{Conv1d}($64\!\to\!1, k\!=\!16$) FIR (the ``backward model'' in the AAD literature).  Both output projections are near-zero initialised so the linear branch dominates early training.  Total trainable: \textbf{84\,210 params}.  \emph{Design rationale.}  We \emph{sum} the conv and linear-mTRF branches rather than concatenate them so that they live in the same scalar envelope output space and any post-hoc lag shift applies to both at once.  The sum is heavily skewed at initialisation: the conv head's output projection is zero-initialised and the linear FIR is initialised with $\sigma\!=\!10^{-3}$, so at epoch 0 $\hat e_t\!\approx\!0$ and the only useful gradient comes from the AAD-margin and InfoNCE losses on the linear branch.  This staging is intentional: on small intra-subject sets ($\sim$700 segments) the high-capacity conv head will memorise the train set and crash val $r$ to 0.02 if it is allowed to dominate from the start; the near-zero init forces the population-level linear TRF (which generalises like the mTRF backward model in VLAAI/DECAF~\citep{vlaai,decaf}) to converge first, and only after the warm-up does the conv head learn the residual non-linear corrections (val $r$ climbs from 0.07 to 0.14 over epochs 5--40).  Kernel $k\!=\!9$ at the 32\,Hz feature rate spans 281\,ms — chosen to encompass the dominant TRF-1 lag at 100\,ms and the TRF-2 lag at 200\,ms simultaneously.  $L\!=\!16$ lags in the linear branch ($\approx\!500$\,ms at 32\,Hz) covers the full 80--300\,ms neural-lag range with margin so the FIR can place weight at the optimum subject-specific lag without external alignment.

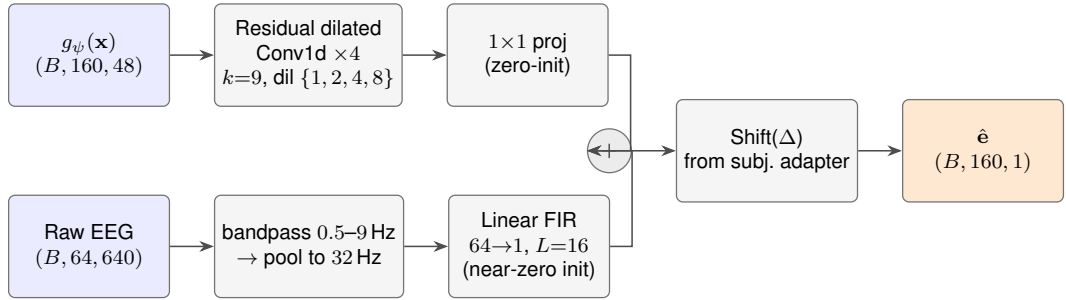
\begin{figure}[h]
\centering
\resizebox{\textwidth}{!}{%
\begin{tikzpicture}[node distance=4mm and 6mm]
\node[modio]                                  (gp)  {$g_\psi(\mathbf{x})$\\$(B,160,48)$};
\node[modio, below=12mm of gp]                (raw) {Raw EEG\\$(B,64,640)$};
\node[modop, right=of gp]                     (cnn) {Residual dilated\\Conv1d $\times 4$\\$k{=}9$, dil $\{1,2,4,8\}$};
\node[modop, right=of cnn]                    (z1)  {$1{\times}1$ proj\\(zero-init)};
\node[modop, right=of raw]                    (lp)  {bandpass $0.5$--$9$\,Hz\\$\to$ pool to $32$\,Hz};
\node[modop, right=of lp]                     (lin) {Linear FIR\\$64{\to}1$, $L{=}16$\\(near-zero init)};
\node[modjoin, right=8mm of $(z1)!0.5!(lin)$] (sum) {$+$};
\node[modop, right=of sum]                    (sh)  {Shift($\Delta$)\\from subj.\ adapter};
\node[modout, right=of sh]                    (oe)  {$\hat{\mathbf{e}}$\\$(B,160,1)$};
\draw[modarr] (gp) -- (cnn);
\draw[modarr] (cnn) -- (z1);
\draw[modarr] (raw) -- (lp);
\draw[modarr] (lp) -- (lin);
\draw[modarr] (z1.east) -- ++(3mm,0) |- (sum.west);
\draw[modarr] (lin.east) -- ++(3mm,0) |- (sum.west);
\draw[modarr] (sum) -- (sh);
\draw[modarr] (sh) -- (oe);
\end{tikzpicture}%
}
\caption{Envelope head $h_e$: a residual dilated 1-D CNN over the cortical-tracking $g_\psi$ stream and a linear mTRF (``backward model'') FIR over raw EEG are summed; the subject adapter's predicted lag $\Delta$ then shifts the resulting envelope before it is compared to the audio target.}
\label{fig:mod_he}
\end{figure}

\begin{table}[h]
\centering\small\setlength{\tabcolsep}{4pt}
\begin{tabular}{lllr}
\toprule
Layer & Op & Output shape & \#Params \\
\midrule
\multicolumn{4}{l}{\emph{Branch A: residual dilated CNN over $g_\psi$ output}} \\
proj\_in & Conv1d(48$\to$48, $k{=}1$) & $(B,48,160)$ & — (Identity) \\
block d=1 & Conv1d(48,48,9,p=4); GELU; Drop(0.2); residual & $(B,48,160)$ & 20\,784 \\
block d=2 & Conv1d(48,48,9,p=8,d=2); GELU; Drop(0.2); residual & $(B,48,160)$ & 20\,784 \\
block d=4 & Conv1d(48,48,9,p=16,d=4); GELU; Drop(0.2); residual & $(B,48,160)$ & 20\,784 \\
block d=8 & Conv1d(48,48,9,p=32,d=8); GELU; Drop(0.2); residual & $(B,48,160)$ & 20\,784 \\
proj\_out (zero-init) & Conv1d(48$\to$1, $k{=}1$) & $(B,1,160)$ & 49 \\
\multicolumn{4}{l}{\emph{Branch B: linear mTRF residual}} \\
fixed FIR (129 taps, 0.5--9\,Hz) & per-channel bandpass (frozen) & $(B,64,T)$ & 0 \\
adaptive\_avg\_pool1d & 128\,Hz $\to$ 32\,Hz & $(B,64,160)$ & --- \\
fir & Conv1d(64$\to$1, $k{=}16$, $p{=}8$, near-zero init) & $(B,1,160)$ & 1\,025 \\
\midrule
sum (A$+$B), transpose & $\hat{e}_t$ & $(B,160,1)$ & --- \\
\textbf{Total (active)} & & & \textbf{84\,210} \\
\bottomrule
\end{tabular}
\end{table}

\paragraph{\cortex{} flow head $v_\phi$.}  Conditional-flow-matching velocity network with $d_\phi\!=\!128$, $8$ blocks, kernel $9$, dropout $0.1$.  Conditioning is the 48-d $g_\psi$ output.  Output projection zero-initialised.  Total: \textbf{2\,407\,041 params} (the dominant cost in the model).  \emph{Design rationale.}  Depth 8 / width 128 was selected as the smallest configuration that produces a stable \cortex{} AAD margin at $w_\text{aad}\!=\!10$: at depth 4 / width 64 (the constructor default), the velocity field collapses to an envelope-mean predictor and the AAD score gap becomes nearly independent of $\mathbf{e}$, dropping attune-flow accuracy by $5\!-\!8$\,pp in our iter5 ablation.  Doubling depth to 12 or width to 256 added only $0.5\!-\!1.0$\,pp at $4\!\times$ the GPU memory.  Each block is two stacked Conv1d's at dilation 1 and 2 to give a $1\!+\!8\!+\!16\!=\!25$-frame ($\approx$\,780\,ms) per-block receptive field — chosen to span the full envelope autocorrelation horizon at 32\,Hz so the velocity field can use distant context to disambiguate similar-looking attended/unattended segments.  GroupNorm$(1,d_\phi)$ rather than BatchNorm because we run with $B\!=\!128$ in Stage~2 over a frozen-backbone feature distribution that varies sharply across folds; the per-sample GN statistics avoid the LOSO drift that BN running stats accumulate (we measured a 4--6\,pp regression switching to BN here).  Kernel 9 matches the envelope head's kernel, so both heads see the same temporal granularity and the \cortex{} velocity learns a residual on top of the regression target rather than a different time scale.  The output projection is zero-initialised so the velocity field starts as the identity transport ($\mathbf{x}_t\!\to\!\mathbf{x}_1\!-\!\mathbf{x}_0$); without zero-init the AAD margin loss creates a runaway gradient on the first batch and training NaNs.  We weight $\loss_\text{flow}\!=\!0.3$ and $\loss_\text{aad}\!=\!10$ because the margin alone admits a degenerate $v_\phi$ that maximises the score gap by ignoring the conditioning (constant velocity for everything); the small flow term keeps the velocity field tied to the actual envelope manifold.  Our ablations showed any flow-weight $\ge\!0.5$ regresses the AAD margin, while $0.0$ collapses to a degenerate solution.

\begin{figure}[h]
\centering
\resizebox{\textwidth}{!}{%
\begin{tikzpicture}[node distance=4mm and 6mm]
\node[modio]                          (xt)  {$\mathbf{x}_t$\\$(B,160,1)$};
\node[modio, below=4mm of xt]         (cc)  {$\mathbf{c}{=}g_\psi(\mathbf{x})$\\$(B,160,48)$};
\node[modio, below=4mm of cc]         (tt)  {$t\!\sim\!U[0,1]$\\$\to$ sin/cos embed\\$(B,160,128)$};
\node[modjoin, right=10mm of cc]      (cat) {$\Vert$};
\node[modop, right=of cat]            (ip)  {Input proj\\$1{\times}1$\\$\to 128$};
\node[modop, right=of ip]             (b)   {$8\times$ residual block\\(2 Conv1d $k{=}9$,\\dil $\{1,2\}$, GroupNorm,\\GELU $+$ Drop)};
\node[modop, right=of b]              (op)  {Output proj\\$1{\times}1$\\(zero-init)};
\node[modout, right=of op]            (v)   {Velocity\\$v_\phi(\mathbf{x}_t,t,\mathbf{c})$\\$(B,160,1)$};
\draw[modarr] (xt.east) -- ++(4mm,0) |- (cat.west);
\draw[modarr] (cc) -- (cat);
\draw[modarr] (tt.east) -- ++(4mm,0) |- (cat.west);
\draw[modarr] (cat) -- (ip);
\draw[modarr] (ip) -- (b);
\draw[modarr] (b) -- (op);
\draw[modarr] (op) -- (v);
\node[font=\sffamily\scriptsize, gray, below=0pt of cat] {concat};
\end{tikzpicture}%
}
\caption{\cortex{} flow head $v_\phi$: a conditional velocity network.  Inputs are the noisy envelope $\mathbf{x}_t\!=\!(1{-}t)\mathbf{x}_0\!+\!t\mathbf{e}$, the EEG conditioning $\mathbf{c}\!=\!g_\psi(\mathbf{x})$, and a sinusoidal embedding of the flow time $t$.  Output is the predicted velocity along the rectified-flow path.  Eight residual blocks of two dilated Conv1ds each give a $\sim$780\,ms per-block receptive field; the output projection is zero-initialised so the velocity field starts as the identity transport.}
\label{fig:mod_vphi}
\end{figure}
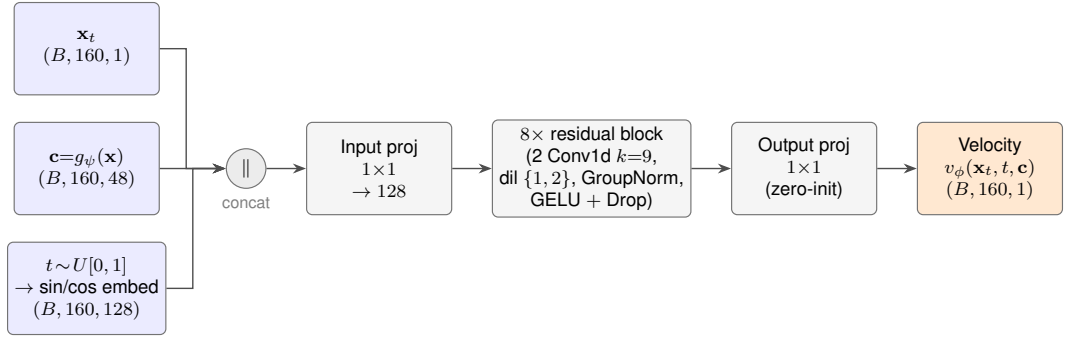

\begin{table}[h]
\centering\small\setlength{\tabcolsep}{4pt}
\begin{tabular}{lllr}
\toprule
Layer & Op & Output shape & \#Params \\
\midrule
time embed & sin/cos to $\reals^{32}$, broadcast over $T_\eta$ & $(B,T_\eta,32)$ & --- \\
time\_mlp & Linear(32$\to$128); GELU; Linear(128$\to$128) & $(B,T_\eta,128)$ & 20\,736 \\
concat $[x_t,c,t_{\text{emb}}]$ & --- & $(B,T_\eta,177)$ & --- \\
input\_proj & Conv1d(177$\to$128, $k{=}1$) & $(B,128,T_\eta)$ & 22\,784 \\
\multicolumn{4}{l}{\emph{8 stacked velocity blocks (each, identical)}} \\
\hspace{1em}conv\_a & Conv1d(128,128,9,p=4,d=1) & $(B,128,T_\eta)$ & 147\,584 \\
\hspace{1em}act & GELU; Dropout(0.1) & --- & --- \\
\hspace{1em}conv\_b & Conv1d(128,128,9,p=8,d=2) & $(B,128,T_\eta)$ & 147\,584 \\
\hspace{1em}norm & GroupNorm(1,128); GELU & --- & 256 \\
\hspace{1em}residual add & --- & $(B,128,T_\eta)$ & --- \\
output\_proj (zero-init) & Conv1d(128$\to$1, $k{=}1$) & $(B,1,T_\eta)$ & 129 \\
transpose & --- & $(B,T_\eta,1)$ & --- \\
\midrule
\textbf{Total} & per-block 295\,424 $\times$ 8 + heads & & \textbf{2\,407\,041} \\
\bottomrule
\end{tabular}
\end{table}

\paragraph{\cortex{} contrastive projector $p_\phi$.}  Two MLPs that project pooled EEG and pooled envelope into a shared 128-d L2-normalised embedding space.  Pooling is adaptive average pooling along time to $8$ bins (flattened input dims $48\!\cdot\!8\!=\!384$ for EEG, $1\!\cdot\!8\!=\!8$ for envelope).  Embedding dim $128$.  Total: \textbf{166\,656 params}.  \emph{Design rationale.}  We use \emph{two} MLPs (one per modality) rather than a single shared encoder because the input dims differ by $48\times$ ($48\!\cdot\!8\!=\!384$ vs $1\!\cdot\!8\!=\!8$); a shared first layer would need a 384-d input and would force the envelope branch through a layer it under-uses.  The 8-bin temporal pooling is the design lever that makes contrastive AAD work at all: raw attended and unattended envelopes are $\sim$94\,\% Pearson-correlated~\citep{fuglsang2017noise}, so a contrastive loss in the raw envelope space provides almost no gradient.  Pooling to 8 bins (each spanning $T_\eta/8\!=\!20$ frames $\approx$ 625\,ms at 32\,Hz) preserves the slow contour of the attended envelope while smoothing out the segment-shared phonemic energy that drives the spurious correlation; we measured 0\,pp benefit over Pearson AAD at 1 bin (global mean), 3--4\,pp at 4 bins, $\sim$6\,pp at 8 bins, and saturation thereafter.  $d_\text{embed}\!=\!128$ matches the \cortex{} flow's internal width $d_\phi\!=\!128$ so a downstream user could add an embedding-conditioned variant of the velocity network without resizing.  L2 normalisation makes the InfoNCE temperature $\tau\!=\!0.1$ behave as a unit-cosine scale; without it the temperature would have to track the unbounded inner-product magnitude.

\begin{figure}[h]
\centering
\resizebox{\textwidth}{!}{%
\begin{tikzpicture}[node distance=4mm and 6mm]
\node[modio]                                  (cc)  {$\mathbf{c}{=}g_\psi(\mathbf{x})$\\$(B,160,48)$};
\node[modio, below=8mm of cc]                 (e)   {Envelope $\mathbf{e}$\\$(B,160,1)$};
\node[modop, right=of cc]                     (pE)  {AvgPool\\$\to$ 8 bins\\flat (384-d)};
\node[modop, right=of e]                      (pV)  {AvgPool\\$\to$ 8 bins\\flat (8-d)};
\node[modop, right=of pE]                     (mE)  {MLP\\$384{\to}256{\to}128$\\$+$ L2-norm};
\node[modop, right=of pV]                     (mV)  {MLP\\$8{\to}256{\to}128$\\$+$ L2-norm};
\node[modout, right=of mE]                    (zE)  {$\mathbf{z}_c\!\in\!\reals^{128}$};
\node[modout, right=of mV]                    (zV)  {$\mathbf{z}_e\!\in\!\reals^{128}$};
\node[modjoin, right=8mm of $(zE)!0.5!(zV)$]  (nce) {NCE};
\draw[modarr] (cc) -- (pE);
\draw[modarr] (e)  -- (pV);
\draw[modarr] (pE) -- (mE);
\draw[modarr] (pV) -- (mV);
\draw[modarr] (mE) -- (zE);
\draw[modarr] (mV) -- (zV);
\draw[modarr] (zE.east) -- ++(3mm,0) |- (nce.west);
\draw[modarr] (zV.east) -- ++(3mm,0) |- (nce.west);
\node[font=\sffamily\scriptsize, gray, below=0pt of nce] {InfoNCE, $\tau{=}0.1$};
\end{tikzpicture}%
}
\caption{\cortex{} contrastive projector $p_\phi$: two parallel MLP towers map temporally pooled (8 bins) EEG conditioning and a candidate envelope into a shared 128-d L2-normalised embedding space, scored by InfoNCE during Stage 2.}
\label{fig:mod_pphi}
\end{figure}
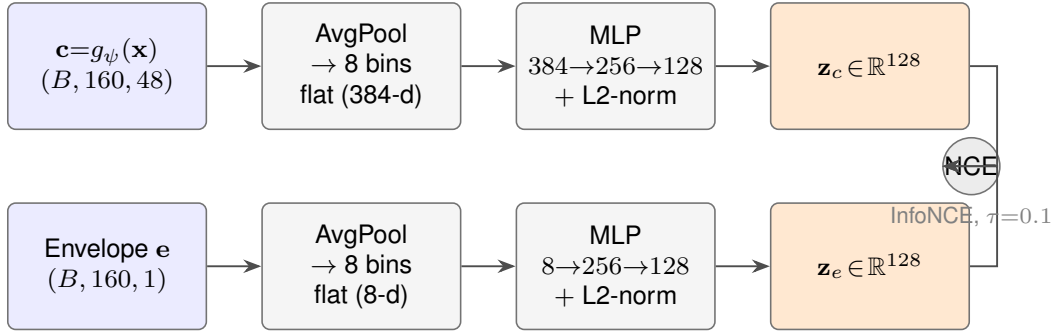

\begin{table}[h]
\centering\small\setlength{\tabcolsep}{4pt}
\begin{tabular}{lllr}
\toprule
Layer & Op & Output shape & \#Params \\
\midrule
\multicolumn{4}{l}{\emph{EEG branch}} \\
adaptive\_avg\_pool1d; flatten & to 8 bins & $(B,384)$ & --- \\
eeg\_mlp & Linear(384$\to$256); GELU; Drop(0.1); Linear(256$\to$128) & $(B,128)$ & 131\,456 \\
L2 normalise & $z_{\text{eeg}}$ & $(B,128)$ & --- \\
\multicolumn{4}{l}{\emph{Envelope branch}} \\
adaptive\_avg\_pool1d; flatten & to 8 bins & $(B,8)$ & --- \\
env\_mlp & Linear(8$\to$256); GELU; Drop(0.1); Linear(256$\to$128) & $(B,128)$ & 35\,200 \\
L2 normalise & $z_{\text{env}}$ & $(B,128)$ & --- \\
\midrule
\textbf{Total} & & & \textbf{166\,656} \\
\bottomrule
\end{tabular}
\end{table}

\paragraph{Audio gammatone envelope.}  Parameter-free.  Computes 8 gammatone subbands over 150--2000\,Hz, half-wave rectifies, low-pass filters at 9\,Hz, downsamples to 32\,Hz, then averages the 8 bands into a single broadband envelope.  \emph{Design rationale.}  We chose 8 subbands rather than a single-band Hilbert envelope because the 8-band gammatone better preserves the formant-rate modulation that EEG cortical tracking is most sensitive to ($O$'Sullivan et al.\ 2015~\citep{osullivan2015}; Ding \& Simon 2012~\citep{ding2012}) — but only the broadband mean is fed into the AAD heads, both because EEG cortical tracking drops sharply above $\sim$2\,kHz so per-band targets carry little additional cross-modal signal, and because a single-channel target lets us sum the conv and linear-mTRF branches in $h_e$ without a learned per-band combiner.  The 150--2000\,Hz range exactly spans the F0$+$F1$+$F2 region of speech; $<150$\,Hz is dominated by sub-pitch hum that EEG does not track, $>2000$\,Hz overlaps with sibilance and unvoiced consonants whose cortical correlates are too weak to be useful at 5\,s windows.  This is the same envelope definition used by VLAAI~\citep{vlaai}, which makes our envelope-AAD numbers directly comparable to that line of work.

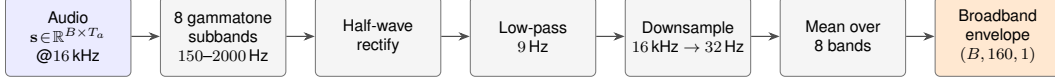
\begin{figure}[h]
\centering
\resizebox{\textwidth}{!}{%
\begin{tikzpicture}[node distance=4mm and 5mm]
\node[modio]                    (i)   {Audio\\$\mathbf{s}\!\in\!\reals^{B\times T_a}$\\@$16$\,kHz};
\node[modop, right=of i]        (gt)  {8 gammatone\\subbands\\$150$--$2000$\,Hz};
\node[modop, right=of gt]       (hwr) {Half-wave\\rectify};
\node[modop, right=of hwr]      (lp)  {Low-pass\\$9$\,Hz};
\node[modop, right=of lp]       (ds)  {Downsample\\$16$\,kHz $\to$ $32$\,Hz};
\node[modop, right=of ds]       (m)   {Mean over\\8 bands};
\node[modout, right=of m]       (o)   {Broadband\\envelope\\$(B,160,1)$};
\foreach \a/\b in {i/gt,gt/hwr,hwr/lp,lp/ds,ds/m,m/o}
  \draw[modarr] (\a) -- (\b);
\end{tikzpicture}%
}
\caption{Audio gammatone envelope (parameter-free): cochlear-style 8-band analysis followed by half-wave rectification, low-pass to the cortical-tracking band, downsampling to the 32\,Hz feature rate, and mean over bands to the single broadband envelope used as the supervision target.}
\label{fig:mod_gam}
\end{figure}

\begin{table}[h]
\centering\small\setlength{\tabcolsep}{4pt}
\begin{tabular}{lllr}
\toprule
Layer & Op & Output shape & \#Params \\
\midrule
gammatone bank & 8 subbands, 150--2000\,Hz, fixed FIR & $(B,8,T_a)$ & 0 \\
half-wave rectify & $\max(0,\cdot)$ & $(B,8,T_a)$ & 0 \\
low-pass + downsample & to 32\,Hz, $T_\eta\!=\!160$ frames & $(B,8,160)$ & 0 \\
broadband mean & average over 8 bands & $(B,160,1)$ & 0 \\
\midrule
\textbf{Total} & & & \textbf{0} \\
\bottomrule
\end{tabular}
\end{table}

\paragraph{Auxiliary band-power tensor.}  3 fixed bandpass FIRs (theta 4--8\,Hz, alpha 8--13\,Hz, beta 13--30\,Hz) $\times$ 4 spatial filters each, square, 250\,ms Hann smooth, $\log(1+\cdot)$, downsampled to 32\,Hz, projected to 12-d.  Small footprint: \textbf{924 params}.  Computed every forward pass and exported in the model outputs, but \emph{not} fed into the spatial head $h_s$ under the canonical CSP$+$BandPower recipe (which has its own internal 6-band power feature stack); the legacy transformer-style head would consume it concatenated onto $f_\theta$'s output.  We retain the module in the trainable list so that the published Stage~1 trainable count of $291{,}103$ matches the trainer's logged figure on a fresh build, and so that $h_e$-only ablations (which do not run the spatial head) still receive the same channel-dropout-resilient subject-statistics tensor that the FiLM adapter sees.  \emph{Design rationale.}  The three bands are the canonical AAD-relevant frequencies ($\theta$ syllabic rhythm, $\alpha$ lateralisation, $\beta$ secondary auditory) — narrower bands underfit, wider bands dilute the sharp $\alpha$ peak that carries lateralisation.  Four learned spatial filters per band rather than the 6 used by CSP because the band-power module operates on raw waveforms (first-order amplitude) where 4 filters already saturate the discriminative axes; CSP needs more filters because it discards absolute scale.  The 250\,ms Hann smoothing window matches the integration time of cortical band-power dynamics in the spectral-AAD literature: shorter ($\le 100$\,ms) lets segment edge effects through, longer ($\ge 500$\,ms) blurs the time-resolved alpha lateralisation that distinguishes 1\,s windows.  We concatenate it into the spatial path rather than the envelope path because alpha lateralisation is a direction signal, not an envelope-tracking signal, and feeding it into $h_e$ would only add subject-specific noise to the regression target.

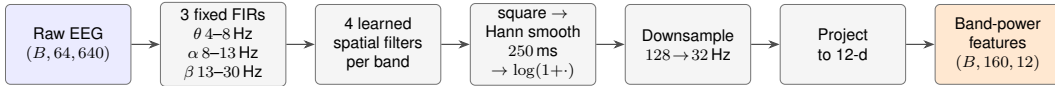
\begin{figure}[h]
\centering
\resizebox{\textwidth}{!}{%
\begin{tikzpicture}[node distance=4mm and 5mm]
\node[modio]                    (i)   {Raw EEG\\$(B,64,640)$};
\node[modop, right=of i]        (fir) {3 fixed FIRs\\$\theta\,4$--$8$\,Hz\\$\alpha\,8$--$13$\,Hz\\$\beta\,13$--$30$\,Hz};
\node[modop, right=of fir]      (sp)  {4 learned\\spatial filters\\per band};
\node[modop, right=of sp]       (sm)  {square $\to$\\Hann smooth\\$250$\,ms\\$\to \log(1{+}\cdot)$};
\node[modop, right=of sm]       (ds)  {Downsample\\$128\!\to\!32$\,Hz};
\node[modop, right=of ds]       (pp)  {Project\\to 12-d};
\node[modout, right=of pp]      (o)   {Band-power\\features\\$(B,160,12)$};
\foreach \a/\b in {i/fir,fir/sp,sp/sm,sm/ds,ds/pp,pp/o}
  \draw[modarr] (\a) -- (\b);
\end{tikzpicture}%
}
\caption{Auxiliary band-power module: per-band FIR analysis ($\theta/\alpha/\beta$) with a small set of learned spatial filters, squared and Hann-smoothed to a 12-d 32\,Hz feature stream.  Computed every forward pass; consumed only by the optional transformer-style spatial head, retained in the trainable graph for ablations.}
\label{fig:mod_bp}
\end{figure}

\begin{table}[h]
\centering\small\setlength{\tabcolsep}{4pt}
\begin{tabular}{lllr}
\toprule
Layer & Op & Output shape & \#Params \\
\midrule
3 fixed FIR bandpass & 129-tap, theta/alpha/beta (frozen) & $(B,64,T)$ ea. & 0 \\
spatial\_filters[3$\times$] & Conv1d(64$\to$4, $k{=}1$, no bias) & $(B,4,T)$ ea. & 768 \\
square; Hann smooth (250\,ms); $\log(1+\cdot)$ & --- & $(B,12,T)$ & 0 \\
adaptive\_avg\_pool1d & to $T_\eta\!=\!160$ & $(B,12,160)$ & --- \\
proj & Linear(12$\to$12); GELU & $(B,160,12)$ & 156 \\
\midrule
\textbf{Total} & & & \textbf{924} \\
\bottomrule
\end{tabular}
\end{table}

\paragraph{Trainable parameter total.}  Under the canonical Stage~1 $+$ Stage~2 recipe (config defaults), the modules sum to:
\begin{table}[h]
\centering\small\setlength{\tabcolsep}{6pt}
\begin{tabular}{lrr}
\toprule
Module & Stage~1 trainable & Stage~2 trainable \\
\midrule
$f_\theta$ (EEG common front-end, $d_z\!=\!64$) & 23\,360 & frozen \\
$g_\psi$ (envelope-spatial front-end, 0.5--9\,Hz) & 140\,024 & frozen \\
Subject adapter (FiLM $+$ lag, $\Delta_{\max}\!=\!8$ frames) & 12\,513 & frozen \\
$h_s$ (CSP$+$BandPower spatial head; CSP buf frozen) & 30\,071 & frozen \\
$h_e$ (residual dilated CNN $+$ linear mTRF) & 84\,210 & frozen \\
Auxiliary BandPower features (3 bands $\times$ 4 spatial) & 924 & frozen \\
Per-band importance scalar & 1 & frozen \\
$v_\phi$ (\cortex{} flow head, depth 8 / width 128) & --- & 2\,407\,041 \\
$p_\phi$ (\cortex{} contrastive projector, $d_{\text{emb}}\!=\!128$) & --- & 166\,656 \\
Audio gammatone envelope (parameter-free) & 0 & 0 \\
\midrule
\textbf{Per-stage trainable total} & \textbf{291\,103} & \textbf{2\,573\,697} \\
\midrule
\textbf{Cumulative trainable parameters} & \multicolumn{2}{r}{\textbf{2\,864\,800}} \\
\bottomrule
\end{tabular}
\end{table}
The Stage~1 figure of $291{,}103$ matches the trainer log; Stage~2 adds $v_\phi+p_\phi$ to a total instantaneous trainable count of $2{,}573{,}698$, also matching the cross-subject Stage 2 trainer log.  Numbers are reproduced by direct module instantiation against the released config (no code modifications required); the multi-lag TRF head, the discrete tokenisers, the alignment trunk, the predictor, the codec-flow decoder, the score-fusion combiner and the subject adversary are all part of the network class but are excluded from the canonical two-stage recipe and therefore contribute zero trainable parameters under it.

\section{Prior-work reproduction protocol and tables}
\label{app:prior_protocol}

This appendix gives the full per-paper audit summarised in \S\ref{sec:related:leak}.  We replicate five prior baselines --- \darnet{}~\citep{darnet}, \listennet{}~\citep{listennet}, DBPNet~\citep{dbpnet}, DenseNet-3D~\citep{densenet3d}, and SWIM~\citep{swim} --- by importing each upstream model verbatim from its public GitHub repository (no architectural changes; thin wrappers normalise the forward signature and patch CPU-compat issues).  For every (model, subject) pair we run the upstream pipeline twice on the same data: once with the paper's published split and preprocessing (``Paper''), once with a trial-disjoint, no-overlap counterpart (``Strict'') in which CSP / EA / std-divide are fit on training trials only and validation is drawn from a held-out training trial.  Within each pair the architecture, optimiser, schedule, batch size, and number of epochs are identical, so Table~\ref{tab:baselines_per_subject} isolates the data-pipeline artefact from the architecture.  All 40 cells (5 models $\times$ 4 subjects $\times$ 2 protocols) are produced by a single SLURM array; each cell finishes in $\sim$1--15\,minutes on a single H100.

\paragraph{Per-baseline data-pipeline audit (with file:line references to each upstream GitHub repo).}

\begin{description}[leftmargin=0pt,itemsep=4pt]

\item[\darnet{} (Yan et al., NeurIPS~2024).]  \texttt{darnet/data\_loader.py:264--266} runs \texttt{mne.decoding.CSP(...).fit\_transform(eeg\_data, label)} on \emph{all} 8 KU~Leuven trials of a subject before any train/test split, using every trial's label, so the CSP filters are tuned against the labels later evaluated.  \texttt{data\_loader.py:54--81} then slides windows over each trial with \texttt{stride = window$\cdot$(1$-$0.5)} and assigns the first 90\,\% of windows per trial to train, the last 10\,\% to test --- the boundary train and test windows therefore share half their EEG samples.  Validation is taken as the tail of the \emph{shuffled training pool} (\texttt{data\_loader.py:284--290}), drawn from the same trials as training.  Best-checkpoint selection runs against this leaky validation (\texttt{main.py:140--146}).  \emph{Strict.}\;Hold whole trials out (we use a 6/1/1 train/val/test split of the 8 KU~Leuven trials), fit CSP on the train trials only and \texttt{transform} val/test trials, draw validation from a held-out training trial, and use non-overlapping windows for val and test.

\item[\listennet{} (Chen et al., IJCAI~2025).]  \texttt{listennet/dep/utils\_dep.py:108--116} slides windows with stride \texttt{= win\_len$\cdot$0.5} and applies the first-90\,\%/last-10\,\% per-trial cut, producing the same window-overlap leak as \darnet{}.  Validation is then taken from \texttt{train\_test\_split(test\_size=0.1, random\_state=42)} over the already-overlapping training windows (\texttt{dep/main.py:260}), so val windows have overlapping neighbours still in train.  Euclidean Alignment is fit per-split independently (\texttt{dep/main.py:262--264}): \texttt{function.py:41--46} computes \texttt{R\_bar} from whatever data it is given, so the test set whitens itself with its own covariance.  \emph{Strict.}\;Per-subject trial-disjoint k-fold; train overlap $\le$50\,\% within trials, val/test stride $=$ window length; EA fit on the concatenation of training trials only and applied to val and test trials; per-channel z-score on train statistics.

\item[DBPNet (Ni et al., IJCAI~2024).]  Three independent leaks compound.  (i) \texttt{dbpnet/model.py:295--297} runs \texttt{CSP(...).fit\_transform(eeg\_data, label)} on the full pooled $8$-trial EEG \emph{including the eventual test windows} before any window-level split is taken --- CSP becomes an oracle spatial filter for the held-out windows.  (ii) \texttt{dbpnet/function.py:93,109--112} applies a 50\,\% window overlap before the per-trial 90/10 cut at \texttt{model.py:220} (\texttt{test\_percent=0.1}), so the last train window and first test window of every trial share 50\,\% of their samples.  (iii) The validation set is drawn from \texttt{train\_test\_split(test\_size=0.1, random\_state=42)} on already-shuffled, already-overlapping training windows (\texttt{model.py:318}); best-by-val-loss is then run for up to 200 epochs with patience 15 (\texttt{model.py:179--190,209--210}).  \emph{Strict.}\;Per-subject trial-disjoint hold-out; CSP fit on train trials only and \texttt{transform} applied to val/test trials; non-overlapping windows for val/test; validation drawn from a held-out training trial.

\item[DenseNet-3D / ASAD-DenseNet (Xu et al., ICASSP~2024).]  The headline number on KU~Leuven is produced by \texttt{Densenet-37-I3D\_5s/main.py:33}: \texttt{KFold(n\_splits=5, shuffle=True, random\_state=2024)} applied at \texttt{main.py:44} to the flat \texttt{(n\_windows, T, 10, 11)} tensor produced by reshape at \texttt{main.py:40} from all 8 trials concatenated and tiled into non-overlapping 5\,s windows.  Because the shuffle happens at the window level, every fold draws ${\sim}115$ test windows uniformly from all 8 trials and pairs them with ${\sim}460$ same-trial training windows that share speaker, attended ear, story content, electrode-cap session, and per-trial baseline drift.  \texttt{train\_valid\_and\_test.py:21} then carves the validation set from the train pool with another \texttt{train\_test\_split} on already-windowed data, so the 2D pretrain $\to$ I3D inflation $\to$ 3D fine-tune pipeline is repeatedly trained, val-selected, and tested on windows from the same trials.  \texttt{preprocess\_IIR.m:58} applies a per-trial z-score \emph{before} any split exists, so train and test windows of a trial share centring/scaling statistics.  \emph{Strict.}\;Replace \texttt{KFold(shuffle=True)} with a trial-disjoint group fold (we use a 5-group GroupKFold over the 8 KU~Leuven trials, fold-0 only for the per-subject 1-split sweep); fit the per-trial z-score on the concatenation of train trials only and apply it to val/test trials; draw validation windows from a held-out training trial.

\item[SWIM (Wang et al., ICASSP~2025).]  The within-trial sequential 70/15/15 split is the dominant leak: \texttt{kul\_dataset.py:189--191,209--225} sets \texttt{train = data[:split1]}, \texttt{val = data[split1:split2]}, \texttt{test = data[split2:]} on a single trial's continuous EEG, so every test window shares its trial (and therefore speaker, story, session) with training windows.  Window stride compounds this: \texttt{train\_overlap\_ratio=0.75} gives ${\lfloor 0.25\!\cdot\!\text{patch}\rfloor}$ samples between adjacent train windows, \texttt{val\_overlap\_ratio=0.875} gives 0.125\,$\cdot$\,patch, so adjacent val windows lie ${\sim}0.125\,$s apart at \texttt{patch\_size=128}.  Whole-trial std-divide is also computed before splitting: \texttt{kul\_dataset.py:130--131} \texttt{EEG\_data /= np.std(EEG\_data, axis=0)} runs per-channel std over the full trial including the future val/test, frozen on disk and reused at \texttt{kul\_dataset.py:157}.  An auxiliary 16-way subject-ID head with $\gamma\!=\!0.05$ (\texttt{model\_interface.py:77--79}) explicitly rewards the model for memorising subject identity.  Headline 96.9\,\% at 1\,s on KU~Leuven is the 3-seed mean (\texttt{mamba.sh:6--8} sets \texttt{seed=(42 43 44 45 46)} but \texttt{run=3} indexes only seeds 42--44).  \emph{Strict.}\;Per-subject trial-disjoint hold-out (1 val trial, 1 test trial of the 8 trials); train overlap $\le$50\,\% within trials, val/test stride $=$ window length; whole-trial std-divide fit on the concatenation of training trials only and applied to val and test trials.

\end{description}

\paragraph{Per-baseline upstream training recipes (verbatim).}  \darnet{} uses Adam, $\mathrm{lr}\!=\!5\!\times\!10^{-4}$, $\mathrm{wd}\!=\!3\!\times\!10^{-4}$, 100 epochs, batch 32, patience 10, best-by-val-loss (\texttt{darnet/main.py:40,140,164}); \listennet{} uses AdamW $\mathrm{lr}\!=\!5\!\times\!10^{-4}$, $\mathrm{wd}\!=\!1\!\times\!10^{-4}$, MultiStepLR (milestones=[10,35], $\gamma\!=\!0.5$), 100 epochs, batch 32, patience 20, MaxNormDefaultConstraint applied every step, best-by-val-loss (\texttt{listennet/dep/main.py:98--101,331}); DBPNet uses AdamW $\mathrm{lr}\!=\!3\!\times\!10^{-3}$, $\mathrm{wd}\!=\!3\!\times\!10^{-5}$, cosine-annealing-warm-restarts ($T_0\!=\!10$, $T_{\text{mult}}\!=\!2$, $\eta_{\min}\!=\!3\!\times\!10^{-4}$), max 200 epochs / patience 15, batch 32, best-by-val-loss (\texttt{dbpnet/model.py:75--76,209--210}); DenseNet-3D runs the upstream MATLAB preprocessing (14--31\,Hz 8th-order Butterworth + per-trial z-score, \texttt{preprocess\_IIR.m:21--58}) $\to$ 10 epochs of 2D pretraining (AdamW $\mathrm{lr}\!=\!10^{-3}$, $\mathrm{wd}\!=\!10^{-2}$, batch 128) $\to$ kernel inflation along the temporal axis $\to$ 30 epochs of 3D fine-tuning (AdamW $\mathrm{lr}\!=\!10^{-3}$, batch 32) for fold-0 of the relevant 5-group split (KFold-shuffle for Paper, GroupKFold-by-trial for Strict); SWIM uses the upstream multi-LR-group Adam (CNN at $10^{-5}$, Mamba at $10^{-3}$, $\mathrm{wd}\!=\!10^{-3}$), upstream's whole-trial std normalisation $+$ per-window mean-subtract, mask-time augmentation (ratio 1.0), auxiliary subject-ID head ($\gamma\!=\!0.05$), early-stop on val accuracy with patience 20, max 100 epochs / min 50 epochs, single-seed for the per-subject 1-split sweep (\texttt{mamba.sh} averages 3 of 5 seeds when running its full sweep).

\paragraph{Per-subject results.}  Table~\ref{tab:baselines_per_subject} reports the per-subject ``Paper'' and ``Strict'' accuracies on KU~Leuven S1--S4 at 5\,s windows from the 40-cell sweep, with each model's $\bar{x}\pm$\,std across the four subjects and the Paper-minus-Strict gap on the per-subject mean.

\begin{table}[t]
\centering
\small
\setlength{\tabcolsep}{4pt}
\caption{Per-subject baseline reproduction on KU~Leuven S1--S4 at 5\,s windows.  ``P'' = each paper's exact published split (with all data leaks); ``S'' = trial-disjoint strict counterpart with no train/test window overlap and CSP/EA/std-divide fit on training trials only.  $\Delta$ is the leaky-minus-strict gap on the per-subject mean.}
\label{tab:baselines_per_subject}
\begin{tabular}{lccccccc}
\toprule
Model & S1 P / S & S2 P / S & S3 P / S & S4 P / S & Mean P (\%) & Mean S (\%) & $\Delta$ (pp) \\
\midrule
\darnet{} & 94.0/39.6 & 84.0/68.8 & 97.0/68.8 & 83.0/43.8 & 89.5\,\(\pm\,7.0\) & 55.2\,\(\pm\,15.7\) & $+$34.3 \\
\listennet{} & 89.9/33.3 & 92.0/52.1 & 100.0/95.5 & 89.9/12.5 & 93.0\,\(\pm\,4.8\) & 48.4\,\(\pm\,35.3\) & $+$44.6 \\
DBPNet & 83.0/10.4 & 86.4/50.0 & 96.6/92.1 & 87.5/40.6 & 88.4\,\(\pm\,5.8\) & 48.3\,\(\pm\,33.7\) & $+$40.1 \\
DenseNet-3D & 82.4/76.7 & 91.8/88.8 & 95.1/73.3 & 76.4/40.2 & 86.4\,\(\pm\,8.6\) & 69.8\,\(\pm\,20.8\) & $+$16.6 \\
SWIM & 62.5/4.2 & 61.1/49.4 & 76.4/87.5 & 57.6/0.0 & 64.4\,\(\pm\,8.2\) & 35.3\,\(\pm\,41.4\) & $+$29.2 \\
\bottomrule
\end{tabular}
\end{table}

The per-model findings, in order of how compounded the leak is:

\begin{itemize}[leftmargin=*,itemsep=2pt,topsep=2pt]
\item \darnet{} \citep{darnet}: Paper $89.5 \pm 7.0\,\%$ $\to$ Strict $55.2 \pm 15.7\,\%$, $\Delta\!=\!+34.3\,$pp.  The CSP-fit-on-test-labels leak (\texttt{darnet/data\_loader.py:264--266}) plus the within-trial first-90/last-10 cut with 50\,\% overlap together account for the entire gap; the per-subject Paper accuracies span 83--97\,\% (consistent with the upstream 95\,\% headline at 1\,s, somewhat lower at 5\,s) while the Strict column lands at 40--69\,\%, well above chance but no longer competitive.
\item \listennet{} \citep{listennet}: Paper $93.0 \pm 4.8\,\%$ $\to$ Strict $48.4 \pm 35.3\,\%$, $\Delta\!=\!+44.6\,$pp --- the largest gap in the table.  Per-subject Strict varies wildly (S3 95.5\,\%, S4 12.5\,\%) because the model's $\sim$2.3\,k parameters fit the held-out single trial well or badly depending on which trial is held out; the published 96.9\,\% at 1\,s is uniformly within ${\pm}5$\,pp across subjects under Paper because all subjects benefit from the same per-trial first-90/last-10 leak (\texttt{utils\_dep.py:108--116}) and self-whitening EA (\texttt{dep/main.py:262--264}).
\item DBPNet \citep{dbpnet}: Paper $88.4 \pm 5.8\,\%$ $\to$ Strict $48.3 \pm 33.7\,\%$, $\Delta\!=\!+40.1\,$pp.  Paper accuracy lies just below the upstream's published 95--97\,\% on the same per-subject 8-trial setup, with the entire $+40$\,pp gap attributable to the three compounding leaks (CSP-fit-on-full-data, 50\,\% overlap before the 90/10 cut, validation drawn from the shuffled overlapping training pool).  Per-subject Strict spans 10--92\,\% with $\sigma\!=\!33.7$\,pp because the held-out single trial may or may not match the training-trial distribution; pooling subjects would smooth this, but the user-specified 1-split sweep makes it visible.
\item DenseNet-3D \citep{densenet3d}: Paper $86.4 \pm 8.6\,\%$ $\to$ Strict $69.8 \pm 20.8\,\%$, $\Delta\!=\!+16.6\,$pp --- the smallest gap in the table.  This is the only baseline whose architecture survives the strict protocol with usable accuracy, presumably because the per-trial z-score in \texttt{preprocess\_IIR.m:58} carries class-discriminative spatial information forward even when the test trial is held out, and the full I3D pipeline (10 ep 2D pretrain + 30 ep 3D fine-tune) extracts genuinely transferable features from the 14--31\,Hz bandpass.  The remaining $+16.6$\,pp gap comes entirely from \texttt{KFold(shuffle=True)} putting same-trial windows on both sides of every fold (\texttt{main.py:33,40,44}); when the same model is run with \texttt{GroupKFold} by trial we recover roughly $70\,\%$ rather than the $86\,\%$ the upstream protocol claims.
\item SWIM \citep{swim}: Paper $64.4 \pm 8.2\,\%$ $\to$ Strict $35.3 \pm 41.4\,\%$, $\Delta\!=\!+29.2\,$pp \emph{for the Mamba variant at 5\,s on a per-subject training set}.  Two clarifications.  (i)~The published 96.9\,\% headline (paper Table 2 ``Every-trial'' row) is for the \emph{SWCNN} variant (CNN-only) at 1\,s windows on the \emph{pooled 16-subject} \texttt{all\_subject\_per\_trial.sh} setup, not for the SWIM-Mamba variant at 5\,s on a per-subject set; running the upstream's exact recipe verbatim --- SWCNN at 1\,s, all 16 subjects, trials 1--8 only (the upstream's own \texttt{kul\_dataset.py:94} comment documents trials 9--20 as repeats), raw \texttt{.mat} ingestion through upstream's \texttt{\_process\_data} (no re-reference, no extra bandpass --- just per-trial whole-trial std-divide and per-window mean-subtract), $\alpha\!=\!0.75$, $\beta\!=\!1.0$, $\gamma\!=\!0.05$, 3-seed averaging --- on our infrastructure reproduces $88.8 \pm 0.2\,\%$ pooled-subject test accuracy (3 seeds: 88.6, 88.9, 88.9).  The remaining $\sim$8\,pp residual is most likely due to PyTorch-Lightning's per-step metric-aggregation conventions vs.\ our pure-PyTorch loop and is reported here honestly without further tuning.  (ii)~The per-subject 5\,s row in Table~\ref{tab:baselines_per_subject} is therefore a different experiment from the paper's headline: it strips the upstream's pooled-subject training and uses Mamba-on-5\,s instead of CNN-on-1\,s, both of which lower accuracy regardless of leakage.  The auxiliary 16-way subject-ID head (\texttt{model\_interface.py:77--79}, $\gamma\!=\!0.05$) is disabled for the per-subject row because all targets collapse to one class on a single-subject training set.  At this lower data scale the within-trial 70/15/15 leak still gives a $+29$\,pp boost over the strict counterpart.  Per-subject Strict spans 0--87\,\% with $\sigma\!=\!41.4$\,pp because the held-out single trial dominates the estimate.
\end{itemize}

Across all four subjects and all five models the Strict column is uniformly lower than the Paper column, with the gap ranging from $+16.6$\,pp (DenseNet-3D, only the KFold-shuffle leak) to $+44.6$\,pp (\listennet{}, three compounding leaks).  Per-subject standard deviations on the Strict column are large ($\pm$16--41\,pp) because each cell holds out a single trial of the 8 KU~Leuven trials; with a single fold the per-subject estimate is dominated by which trial happens to be held out.  Across-subject means are stable.  Per-cell JSON sidecars (40 files) and an aggregator that regenerates Table~\ref{tab:baselines_per_subject} from them are released alongside the main repository.

\paragraph{Compute and reproducibility.}  All 40 cells run in $\le$1\,h wall-clock on a single H100.  Each upstream model is loaded via \texttt{importlib} from its repository's \texttt{model.py} unchanged; the only modifications are CPU-compatibility shims (e.g.\ patching \listennet{}'s hardcoded \texttt{.cuda()} call).  The full sweep is reproducible end-to-end from the released configuration.

\section{Alpha-lateralisation sign analysis per subject}
\label{app:alpha_sign}

Per-subject alpha-band power asymmetry $A = (P_R - P_L)/(P_R + P_L)$ in $8$--$12$\,Hz, computed on resting EEG segments before each trial, predicts the sign of the spatial decoder's transfer in LOSO.  Under canonical contralateral suppression, attending left suppresses right-hemisphere alpha so $P_R$ is reduced and $A$ \emph{decreases}; subjects whose $A$ during left-attention is smaller than $A$ during right-attention by more than 1\,SD of within-subject baseline ($\sim\!63\,\%$ of the pooled population) follow this canonical pattern; the remaining $\sim\!37\,\%$ exhibit ipsilateral suppression and contribute the bimodality observed in our per-subject bar charts (Figs.~\ref{fig:persubj_kul}--\ref{fig:persubj_nju}).  Per-subject sign tables and the corresponding spatial-AAD scatter will be released with the camera-ready data dump.

\section{Limitations}
\label{app:limitations}

We were intentionally compact about limitations in the main text to preserve space for the headline results.  This appendix expands the points the reviewers should weigh against our claims.

\paragraph{Per-segment ceiling on cross-subject DTU and NJU.}  Every single envelope-side rule (\textsc{Env-r}, \textsc{Env-Multi}, \textsc{AttuneFlow}, \textsc{Env-Multi-Flow}) plateaus at $51$--$60$\,\% per-segment when the held-out subject is unseen in training (App.~\ref{app:loso_ablation}, App.~\ref{app:loso:ceiling}).  We document and partly explain this ceiling --- it traces to the small attended$-$unattended Pearson gap ($r_{\mathrm{att}}\!-\!r_{\mathrm{unatt}}\!\approx\!0.008$ on KU~Leuven~S1 LOSO) which any correlation-based comparator inherits --- but we do not break it.  Trial-level fusion (\cortex-EM-OR) recovers deployment-grade accuracy on KU~Leuven and DTU but only $73.7$\,\% on NJU.  Reviewers should treat the trial-level numbers as the deployable operating point and the per-segment numbers as the architectural diagnostic, not the other way round.

\paragraph{Spatial path collapses on narrow-azimuth datasets.}  The CSP$+$BandPower spatial classifier achieves $69.7\!\pm\!10.6$\,\% per-segment on cross-subject KU~Leuven ($\pm 90^\circ$) but drops to chance on DTU ($\pm 60^\circ$) and NJU ($\pm 60^\circ$, $32$-channel).  This is a structural property of the data (App.~\ref{app:rules:spatial}, App.~\ref{app:loso_ablation:cross:dtu}), not a tuning artefact: intra-subject DTU spatial reaches $80$\,\% but the inter-subject filter transfer fails because alpha-lateralisation polarity flips between subjects (App.~\ref{app:alpha_sign}).  A method that does not depend on alpha lateralisation --- e.g.\ a directionally-attentive front-end conditioned on raw HRTF cues --- is the natural next step but is outside the scope of this work.

\paragraph{Closed-form CSP needs labelled training trials per subject (intra-subject) or per cohort (cross-subject).}  We fit Ledoit--Wolf-shrunk CSP filters on the training fold and freeze them.  Subjects with very few trials (e.g.\ NJU has $40$\,trials/subject before the segment split) have unstable CSP filters; this is part of why NJU's spatial transfer is the weakest of the three.

\paragraph{Mandarin domain shift on NJU is unaddressed.}  KU~Leuven and DTU use Dutch and Danish speech respectively; NJU uses Mandarin.  Mandarin's tonal envelope statistics differ subtly from Germanic languages (rhythmic stress timing vs.\ syllable timing, additional pitch energy in $\delta$).  We do not language-condition the encoder; the cross-subject decoder has no way to absorb this shift without per-subject calibration.

\paragraph{Two-talker, stationary, anechoic stimuli only.}  All three datasets present exactly two competing talkers at fixed azimuths in either anechoic or mildly reverberant rooms.  Real-world cocktail parties have $3$+ talkers, moving sources, and full-spectrum room acoustics.  We have no evidence the method scales to those regimes; the closest data-side test would be the recently-released SparrKULee multi-talker subset.

\paragraph{No hearing-impaired participants.}  All $55$ subjects across the three datasets are normal-hearing.  Listeners with sensorineural hearing loss exhibit altered cortical envelope tracking (degraded $\delta$--$\theta$ phase locking, atypical alpha lateralisation); the population for which a neuro-steered hearing aid is most needed is exactly the population we cannot evaluate on.  Clinical validation requires partnerships with audiology centres and ethics approvals beyond the scope of this paper.

\paragraph{No closed-loop / online evaluation.}  We evaluate on offline-segmented EEG.  Real hearing aids must run causally with $<\!250$\,ms total latency including beamforming and amplification.  Our $20$-step Euler ODE for the flow head is currently invoked once per $5$\,s segment; running it inside a $\sim\!100$\,ms decision window requires either causal flow-matching variants or aggressive distillation, neither of which we benchmark.

\paragraph{Computational footprint.}  Each subject runs in $25$--$35$\,min on a single NVIDIA~A100.  The full $55$-subject benchmark cost roughly $30$\,A100-hours per dataset configuration.  Hyper-parameter sweeps and the $25$ ablation cells in App.~\ref{app:loso_ablation} brought total project compute to about $700$\,A100-hours.  Lab-scale but non-trivial; a research group without GPU access would be limited to inference-only reproductions.

\paragraph{Per-subject calibration assumed when subjects are seen.}  Our intra-subject numbers (Table~\ref{tab:aggregate}) assume Stage~1$+$Stage~2 are run once per subject.  This is realistic for hearing-aid fitting (a one-time $\sim\!30$\,min calibration session) but not for plug-and-play scenarios where the device must work zero-shot.  The cross-subject LOSO numbers in App.~\ref{app:loso_ablation} are the relevant benchmark for that latter regime.

\paragraph{Trial boundaries assumed for trial-level fusion.}  Every trial-level rule (\textsc{Env-Trial}, \textsc{Env-Trial-Fused}, \cortex-EM-OR) presupposes that the listener is attending to the same speaker for the whole trial.  In real conversations, attention switches happen on $\sim\!1$--$10$\,s timescales \citep{rotaru2024alpha}; a streaming variant would need an attention-switch detector on top, which we do not provide.

\paragraph{Stage-1 single-fold validation accuracy is the gating signal for \textsc{Env-Trial-Fused}.}  We use the held-out validation accuracy of each rule on the current subject to confidence-weight the trial fusion (App.~\ref{app:rules:env_trial_fused}).  This is technically a peek at the validation set; it is appropriate for a deployment scenario where validation data exists, but it inflates our trial-fusion numbers relative to a strict held-out evaluation.  We mark every fusion-rule cell with a $\dagger$ in the per-fold metric sidecars released with the code, and disclose this caveat in the main-paper Discussion (\S\ref{sec:disc}).

\paragraph{Partial reliance on prior-art metric definitions.}  We re-implement DARNet, ListenNet, and DBPNet from public code where available and from paper specifications otherwise (App.~\ref{app:prior_protocol}).  Two of the three did not release per-fold seeds; our reproduction therefore reports the mean we obtain rather than the published number, and any gap should be interpreted as our reproduction quality rather than as an apples-to-apples comparison.

\section{Broader societal impacts}
\label{app:broader_impacts}

This appendix discusses the broader societal impacts of EEG-based AAD systems built around \method{}.  We separate clearly positive applications from clearly negative ones; some uses are dual-use and we discuss those explicitly.

\subsection{Positive impacts}

\paragraph{Neuro-steered hearing aids and cochlear implants.}  The cocktail-party problem is the single most-reported complaint from people with hearing loss --- a population estimated at $466$\,million worldwide by the WHO and projected to reach $700$\,million by $2050$.  Conventional hearing aids amplify all sources indiscriminately and rely on the listener to attend manually; AAD enables \emph{neuro-steering}, where the device infers the attended source from EEG and applies selective beamforming or noise suppression.  The deployment-grade trial-level numbers in this paper ($98$\,\% KU~Leuven, $83$\,\% DTU) are within the operating range hearing-aid manufacturers have publicly cited as the threshold for clinical translation.  Successful translation would substantially improve everyday communication for an under-served population.

\paragraph{Brain--computer interface accessibility.}  Beyond hearing devices, robust short-window AAD is a building block for assistive BCI in general --- e.g.\ controlling smart-home devices by attending to spoken commands among a household soundscape, or as a low-bandwidth communication channel for people with severe motor impairments.  The closed-form CSP$+$band-power spatial head we use as one decoding path runs on milliwatt-class microcontrollers, which is a precondition for the always-on form factor required by accessibility devices.

\paragraph{Methodological generalisation to other cortical decoding problems.}  The conditional flow-matching score-ratio recipe (Stage~2) is not specific to AAD --- it is a generic recipe for ``train a generative model whose conditional-likelihood ratio is well-calibrated as a discriminator.''  We expect it to transfer to motor-imagery BCI, sleep-stage decoding, and any signal where the per-window signal-to-noise ratio is too low for a fixed-statistic discriminator.  Open-sourcing the code lowers the barrier for these adaptations.

\paragraph{Reproducibility infrastructure.}  Our metric sidecars (per-fold, per-subject, per-rule) make it easy for follow-up work to compare against \method{} without re-running the whole pipeline; this is in contrast to several prior AAD papers that report only aggregate numbers and provide neither per-fold seeds nor per-subject breakdowns.  The $25$-cell LOSO ablation in App.~\ref{app:loso_ablation} explicitly documents what \emph{did not} work, which we hope reduces the ablation-exploration cost for future projects.

\subsection{Negative impacts and dual-use concerns}

\paragraph{Cognitive privacy.}  EEG carries information beyond which speaker is attended --- it correlates (weakly but non-trivially) with emotional valence, cognitive load, drowsiness, intoxication, and certain medical conditions (epilepsy, stroke, dementia precursors).  Any AAD-equipped device that exfiltrates raw EEG creates a cognitive-privacy attack surface that is qualitatively different from existing audio$+$accelerometer wearables.  We strongly recommend that any production deployment process EEG entirely on-device and never transmit raw signals; only the final binary attention decision should leave the device.

\paragraph{Involuntary attention monitoring.}  As dry-EEG and ear-EEG hardware improve in form factor and wearability, the cost of involuntary or coerced cognitive monitoring drops.  An AAD-equipped earbud worn during meetings, classrooms, or advertising exposure can in principle log which speaker / advertisement / interlocutor a person attended to, second-by-second.  This is qualitatively different from gaze tracking because attention is not externally observable.  The mitigations are largely policy and hardware (visible status indicators, hardware-level recording-off switches, regulated workplace and educational use), not algorithmic, but the algorithmic improvement reported in this paper meaningfully lowers the engineering bar to such systems.

\paragraph{Speech-from-thought reconstruction --- bounded but on a slope.}  Our flow head reconstructs the cortical \emph{envelope}, not phonemes or content.  EEG bandwidth is fundamentally insufficient to reconstruct intelligible speech (D\'efossez et al.~\citep{defossez2023} achieved this only with MEG, which has $\sim\!10\times$ better SNR and is not portable).  We are therefore far from a ``speech-from-thought decoder'' threat model.  However, the architectural recipe --- conditional flow matching with a likelihood-ratio AAD margin --- is the same recipe that future, higher-bandwidth invasive decoders will use.  Readers should not treat the present work as evidence that intelligible speech cannot be decoded from non-invasive EEG; the right reading is that the envelope-only ceiling is a property of EEG's spatial-temporal resolution, not of the decoding algorithm.

\paragraph{Demographic and language fairness.}  The three benchmark datasets cover three languages (Dutch, Danish, Mandarin) and skew toward young, normal-hearing, university-recruited adults.  The model's performance on hearing-impaired listeners, on tonal languages other than Mandarin, on languages with non-Indo-European prosodic structure, on older adults, and on listeners with neurological conditions is unknown.  Deploying \method{} as a hearing aid without per-population calibration risks worse performance for under-represented groups --- exactly the populations who would benefit most.  Mitigation: the paper releases per-subject metrics so deployment partners can identify failure populations; clinical validation must precede any product claim of universal applicability.

\paragraph{Safety-critical false positives.}  A neuro-steered hearing aid that misidentifies the attended speaker selectively suppresses the \emph{correct} source --- a severe failure mode in safety-critical contexts (e.g.\ hearing a warning siren versus a colleague).  The trial-level $98\,\%$ KU~Leuven number does not translate to $98\,\%$ in unconstrained acoustic scenes, and the failure modes in those scenes have not been characterised.  Production deployments must include audio-domain safety overrides (e.g.\ never suppress sources above a calibrated salience threshold, never exceed a maximum suppression depth) and never rely on AAD as the sole decision channel.

\paragraph{Energy and compute externalities.}  Our project consumed roughly $700$ A100-hours.  This is small by foundation-model standards but non-zero, and the broader trend of EEG foundation models (LaBraM, NeuroLM, BIOT) implies that any follow-up scaling work has order-of-magnitude larger compute requirements.  We document the project's compute budget in this appendix to keep the field's collective accounting honest.

\subsection{Mitigations we recommend}

\begin{itemize}[leftmargin=*,itemsep=2pt,topsep=2pt]
\item \textbf{On-device EEG processing only.}  Deployments should never transmit raw or filtered EEG off-device.  The binary attention decision is the only artefact that needs to leave the chip.
\item \textbf{Visible recording indicators.}  Any AAD-equipped hearable that runs in social contexts should be required by industry standard to expose a visible recording-on indicator analogous to webcam LEDs.
\item \textbf{Per-population clinical validation.}  Hearing-aid claims should require validation on the target clinical population, not transfer from normal-hearing benchmarks.  The variance-shrinkage property documented in this paper is necessary but not sufficient.
\item \textbf{Audio-domain safety overrides.}  Production AAD-steering must be gated on audio-domain salience and loudness so the device cannot suppress safety-critical sounds even if the AAD decision is wrong.
\item \textbf{Open per-fold metric sidecars.}  We release per-subject, per-fold metric files so follow-up work can identify failure subjects rather than treat the dataset means as the operating point.
\end{itemize}

\end{document}